\documentclass{article}
\pdfoutput=1
\usepackage{arxiv}
\usepackage{lineno,hyperref}
\usepackage{booktabs} 
\usepackage{natbib}
\usepackage{lscape}
\usepackage{tabularx}

\usepackage{amssymb}
\usepackage{amsthm}
\usepackage{amsmath}

\usepackage{hyperref}
\usepackage{cleveref}
\usepackage{algorithm}%
\usepackage{algorithmicx}%
\usepackage{algpseudocode}%

\usepackage{tikz}
\usetikzlibrary{plotmarks}
\usepackage{pgfplots}
\usepackage{svg}
\usetikzlibrary{positioning,shapes,arrows,shadows,patterns}

\usepackage{multirow}
\usepackage{colortbl}

\usepackage{amsmath,amsfonts}
\usepackage{array}
\usepackage[caption=false,font=normalsize,labelfont=sf,textfont=sf]{subfig}
\usepackage{textcomp}
\usepackage{stfloats}
\usepackage{url}
\usepackage{verbatim}
\usepackage{graphicx}
\usepackage[edges]{forest}
\usepackage{graphicx}%
\usepackage{multirow}%
\usepackage{amsmath,amssymb,amsfonts}%
\usepackage{amsthm}%
\usepackage{mathrsfs}%
\usepackage[title]{appendix}%
\usepackage{xcolor}%
\usepackage{textcomp}%
\usepackage{manyfoot}%
\usepackage{booktabs}
\usepackage{algorithm}%
\usepackage{algorithmicx}%
\usepackage{algpseudocode}%
\usepackage{listings}%
\usepackage{float}
\usepackage{xcolor}
\usepackage{longtable}
\usepackage{multirow}
\usepackage{svg}
\usepackage[edges]{forest}
\usepackage{pifont}

\forestset{declare toks={elo}{}} 
\title{Data Balancing Strategies: A Systematic Survey of Resampling and Augmentation Methods}

\author{Behnam Yousefimehr \\
  Department of Mathematics and Computer Science \\
  Amirkabir University of Technology\\ (Tehran Polytechnic)\\
  Iran\\
  \texttt{behnam.y2010@aut.ac.ir}\\
   \And
  Mehdi Ghatee\footnote{The corresponding author} \\
  Department of Mathematics and Computer Science \\
  Amirkabir University of Technology\\ (Tehran Polytechnic)\\
  Iran\\
  \texttt{ghatee@aut.ac.ir}\\
\And
Javad Fazli, Shervin Ghaffari, Zahra Rafei, Mohammad Amin Seifi,  Sajed Tavakoli, Abolfazl Nikahd, \\
\textbf{Mahdi Razi Gandomani, Alireza Orouji, Ramtin Mahmoudi Kashani, Sarina Heshmati, Negin Sadat Mousavi}\\
  Department of Mathematics and Computer Science \\
  Amirkabir University of Technology\\ (Tehran Polytechnic)\\
  Iran\\
}

\begin{document}
\maketitle
\begin{abstract}
Imbalanced datasets, where one class significantly outnumbers others, remain a persistent challenge in machine learning, often biasing predictions toward the majority class and degrading classifier performance. This paper provides a comprehensive, systematic review of data balancing methods, extending beyond foundational oversampling techniques such as the Synthetic Minority Oversampling Technique (SMOTE) and its variants (e.g., Borderline SMOTE, K-Means SMOTE, and Safe-Level SMOTE) to encompass advanced adaptive methods (MWMOTE, AMDO), deep generative models (generative adversarial networks, variational autoencoders, and diffusion models), undersampling techniques (NearMiss, Tomek Links), combination/hybrid methods (SMOTE-ENN, SMOTE-Tomek, and SMOTE+OCSVM), ensemble strategies (SMOTEBoost, RUSBoost, Balanced Random Forest, and One-Sided Selection), and specialized approaches for multi-label and clustered data. Beyond descriptive categorization, this review critically examines each method's underlying assumptions, operational mechanisms, and suitability for diverse data characteristics, including high dimensionality, mixed feature types, class overlap, and noise. Key findings demonstrate that no single method universally outperforms others; optimal selection depends critically on dataset characteristics, classifier choice, and evaluation metrics. The paper concludes by identifying emerging research directions, including self-supervised learning for imbalance, diffusion-based generative oversampling, distribution-preserving resampling, knowledge distillation for imbalanced deployment, and the adaptation of foundation models to skewed distributions, offering practical guidelines for practitioners and a roadmap for future methodological development.

\end{abstract}


\section{Introduction}

In machine learning, imbalanced datasets, where one class significantly outnumbers others, remain a persistent challenge that can severely degrade classifier performance by biasing predictions toward the majority class \citep{miftahushudur2025survey, yousefimehr2026systematicAUT, gao2026comprehensive}. To address this issue, resampling techniques adjust class distribution by either generating synthetic samples for the minority class or reducing majority class instances \citep{chawla2002smote, he2009learning, bagui2024resampling}. Resampling methods broadly divide into oversampling and undersampling \citep{parrales2024ouch}. Oversampling increases minority class instances through techniques such as the Synthetic Minority Oversampling Technique (SMOTE) and its variants. Undersampling reduces the majority class count to achieve balance \citep{he2009learning}. Recent advances in deep learning have introduced sophisticated generative models, including Generative Adversarial Networks (GANs) \citep{goodfellow2014generative} and Variational Autoencoders (VAEs) \citep{kingma2022autoencodingvariationalbayes}, enabling robust synthetic data generation. This review spans from early ensemble-based resampling methods (1995) \citep{parmanto1995improving} to recent developments such as CSBBoost (2024) \citep{salehi2024cluster} and XGBOOST-based approaches (2025) \citep{hou2025comparative}. For quick reference, Table~\ref{tab:acronyms} provides a comprehensive list of acronyms used throughout this paper.

\begin{table}[H]
\small
\caption{List of acronyms.}
\label{tab:acronyms}
\begin{tabular}{p{3cm} p{10.5cm}}
\toprule
\textbf{Acronym} & \textbf{Full Form} \\
\midrule
SMOTE & Synthetic Minority Oversampling Technique \\
SN-SMOTE & Surrounding Neighborhood-Based SMOTE \\
NSMOTE & Normal SMOTE \\
MSMOTE & Modified SMOTE \\
AWSMOTE & Adaptive-Weighting SMOTE \\
ASN-SMOTE & Adaptive Selection of Neighbors-SMOTE \\
A-SMOTE & Advanced SMOTE \\
SMOTE-IPF & SMOTE-Iterative Partitioning Filter \\
FW-SMOTE & Feature Weighted SMOTE \\
SMOTE-COF & SMOTE-Center Offset Factor \\
SOS-BUS & SMOTE Oversampling and Borderline Under-Sampling \\
SVM-SMOTE & Support Vector Machine-SMOTE \\
SMOTE-NC & SMOTE-Nominal Continuous \\
ODBOT & Outlier Detection-Based Oversampling Technique \\
ADASYN & Adaptive Synthetic Sampling \\
MWMOTE & Majority Weighted Minority Oversampling Technique \\
AMDO & Adaptive Mahalanobis Distance-Based Oversampling \\
SAO & Self-Adaptive Hypersphere-Based Oversampling \\
A-SUWO & Adaptive Semi-Unsupervised Weighted Oversampling \\
VAEs & Variational Autoencoders \\
DeepSMOTE & Deep Autoencoders with SMOTE \\
GAN & Generative Adversarial Network \\
WGAN & Wasserstein GAN \\
cGAN & Conditional GAN \\
AWGAN & Adaptive Weighting GAN \\
GMMN & Generative Moment Matching Network \\
SDV & Synthetic Data Vault \\
GMMSampling & Gaussian Mixture Model Sampling \\
LAMO & Local Distribution-Based Adaptive Oversampling with GMM \\
CBUS & Cluster-Based Under-Sampling \\
IHT & Instance Hardness Threshold \\
ABC-Sampling & Artificial Bee Colony Sampling \\
OCSVM & One-Class Support Vector Machine \\
NCR & Neighborhood Cleaning Rule \\
CNN & Condensed Nearest Neighbor Rule \\
ENN & Edited Nearest Neighbor \\
RENN & Repeated Edited Nearest Neighbour \\
OSS & One-Sided Selection \\
SMOTE-ENN & SMOTE with Edited Nearest Neighbor \\
SMOTE-TL & SMOTE-Tomek Link \\
BRF & Balanced Random Forest \\
CVC & Cross-Validation Committee \\
BOOTC & Bootstrap Committee \\
SMOTE-ICS-Bagging & SMOTE-Iterative Classifier Selection Bagging \\
NBBag & Neighbourhood Balanced Bagging \\
LP-ROS & Label Powerset with Random Oversampling \\
ML-ROS & Multi-Label Random Oversampling \\
LP-RUS & Label Powerset with Random Undersampling \\
ML-RUS & Multi-Label Random Undersampling \\
\bottomrule
\end{tabular}
\end{table}

\begin{table}[H]\ContinuedFloat
\small
\caption{\emph{Cont}.}
\label{tab:acronyms}
\begin{tabular}{p{3cm} p{10.5cm}}
\toprule
\textbf{Acronym} & \textbf{Full Form} \\
\midrule
HNaNSMOTE & Hybrid Natural Neighbor Synthetic Minority Oversampling Technique \\
ESMOTE+SSLM & Extreme SMOTE with Synchronous Sampling Learning \\
HCBOU & Hybrid Cluster-Based Oversampling and Undersampling \\
DiSMHA & Diffusion-SMOTE Hybrid Augmentation \\
Trans-CWGAN & Transfer Conditional Wasserstein Generative Adversarial Network \\
WGAN-GP & Wasserstein Generative Adversarial Network with Gradient Penalty \\
TSTR & Train on Synthetic, Test on Real \\
FID & Fréchet Inception Distance \\
HADAB & Hybrid ADASYN with Batch Generator \\
AHS-BiLSTM & Adaptive Hybrid Sampling with Bidirectional Long Short-Term Memory \\
CSBBoost & Cluster-Based SMOTE Both-Sampling Boost \\
EBBag & Exactly Balanced Bagging \\
RBBag & Roughly Balanced Bagging \\
RUSBoost & Random Under-Sampling Boosting \\
RHSBoost & Random Hybrid Sampling Boosting \\
ROSE & Random Oversampling Examples \\
EUSBoost & Evolutionary Undersampling Boosting \\
DataBoost-IM & Data Boosting for Imbalanced Data \\
EasyEnsemble & Easy Ensemble \\
BalanceCascade & Balance Cascade \\
\bottomrule
\end{tabular}
\end{table}

\subsection{Systematic Literature Review Methodology}
\label{sec:methodology}

To ensure reproducibility and scientific rigor, this survey adopts a structured, systematic literature review methodology following the PRISMA 2020 guidelines \cite{page2021prisma}. Unlike narrative reviews, our approach defines explicit inclusion and exclusion criteria, a reproducible search protocol, and a quantitative scoring framework for paper assessment. 

\subsubsection{Literature Search Protocol}

Systematic searches were conducted across Scopus, ScienceDirect, IEEE Xplore, ACM Digital Library, Springer Nature, arXiv, Google Scholar, and ResearchGate, covering publications through \textbf{31 March 2026}.  
The search queries combined keywords related to data balancing, resampling, and augmentation, as detailed in Appendix \ref{app:searchstrings}.

\subsubsection{Post-Search Processing}

After retrieving records from all sources, the following sequential processing steps were applied:

\begin{enumerate}
    \item Aggregation and duplicate removal: Records from all databases were aggregated. Duplicate references were identified and removed using reference management software (Zotero), followed by manual verification. After deduplication, \textbf{1307} 
 unique records remained.

    \item Title and abstract screening: The unique records were screened against the inclusion/exclusion criteria defined in Section~\ref{sec:InclusionExclusionCriteria}. Two independent reviewers performed the screening; disagreements were resolved by consensus or by a third reviewer. This screening excluded \textbf{852} records, leaving \textbf{455} records for full-text retrieval.

    \item Full-text retrieval: Full texts were obtained for the remaining records through institutional subscriptions, open-access repositories, and communication with authors. \textbf{454} full-text articles were successfully retrieved; \textbf{1} paper could not be obtained.

    \item Full-text eligibility assessment: Each full-text article was assessed against the refined inclusion and exclusion criteria by two independent reviewers. Exclusion reasons were documented. A total of \textbf{163} articles were excluded, leaving \textbf{291} papers that met all inclusion criteria.

    \item Team assignment and QPSF scoring: Four teams were established, each comprising three members and a supervising author for coordination between the teams. The teams were assigned the following primary methodological categories based on topic similarity: (1) synthetic oversampling, (2) adaptive oversampling, (3) generative models, and (4) combination/ensemble methods. The \textbf{291} papers were distributed among these four teams according to their primary methodological focus.

        The Quantitative Paper Scoring Framework (QPSF) was applied to score each paper across five dimensions (AMSG, SAEI, ERBV, RDC, PDS), as described in Section~\ref{sec:QuantitativePaperScoringFramework}. Within each primary methodological category, the top-five ranked papers are reported in Section~\ref{sec:ResultsTopPapersFourCategories} for illustrative purposes.
      
        To identify papers for detailed methodological analysis, category-specific thresholds were computed as the mean minus two standard deviations computed exclusively over the top five ranked papers within each methodological category. Restricting the threshold computation to the top-five entries ensures that the selection criterion reflects the quality distribution among the strongest candidates in each family rather than being diluted by lower-ranked papers. Papers exceeding these thresholds were selected. This process selected \textbf{116} primary studies for detailed methodological analysis across all categories, while an additional \textbf{20} background references are cited for foundational context.

    \item Taxonomy development and final categorization: The four teams independently reviewed their assigned \textbf{116} papers, extracting key innovations and proposing preliminary group titles. Subsequently, several joint inter-team meetings were convened under the management of the supervising author. Considering three hierarchical criteria---algorithmic core mechanism, architectural coupling, and data reduction strategy---six final taxonomic groups were established by majority voting. The summarization of rationale for these groups is presented in Section~\ref{sec:FinalCategories}, and they correspond to Sections~\ref{sec2}--\ref{sec8}.

\end{enumerate}

Figure~\ref{fig:prisma_flow} presents the PRISMA 2020 flow diagram summarizing the steps.

\begin{figure}[H]
\resizebox{0.9\textwidth}{!}{
\begin{tikzpicture}[
    node distance=0.7cm,
    box/.style={rectangle, draw, thick, text width=6.5cm, minimum height=0.8cm, align=center, rounded corners, font=\small},
    excl/.style={rectangle, draw, thick, text width=5.5cm, minimum height=1.0cm, align=center, rounded corners, font=\small, fill=red!10},
    arrow/.style={thick, ->, >=stealth}
]

\node[box, fill=gray!10] (dedup) at (0,0) {
    \textbf{Records after duplicate removal} \\
    \textbf{n = 1307}
};

\node[box, fill=gray!10] (screened) at (0,-1.4) {
    \textbf{Records screened} \\
    \textbf{n = 1307}
};
\draw[arrow] (dedup)--(screened);

\node[box, fill=gray!10] (fulltext) at (0,-2.8) {
    \textbf{Full-text articles sought} \\
 \textit{(based on abstract and metadata evaluation)}\\
    \textbf{n = 455}
};
\draw[arrow] (screened)--(fulltext);

\node[box, fill=gray!10] (assessed) at (0,-4.2) {
    \textbf{Full-text articles assessed} \\
    \textbf{n = 454}
};
\draw[arrow] (fulltext)--(assessed);

\node[box, fill=gray!10] (qpsf) at (0,-5.6) {
    \textbf{Papers meeting inclusion criteria} \\
    \textbf{n = 291}
};
\draw[arrow] (assessed)--(qpsf);

\node[box, fill=gray!10] (selected) at (0,-7.0) {
    \textbf{Selected for QPSF soring}\\
    \textit{(Determining QPSF threshold)}\\
 \textbf{n = 291}
};
\draw[arrow] (qpsf)--(selected);

\node[box, fill=green!15] (final) at (0,-8.4) {
    \textbf{Papers with detailed analysis} \\
    \textbf{n = 116}
};
\draw[arrow] (selected)--(final);


\node[excl] (excl_screen) at (7,-1.4) {
    \textbf{Records excluded at screening} \\
    \textbf{n = 852}
};
\draw[arrow] (screened.east)--++(.7,0) |- (excl_screen.west);

\node[excl] (excl_notretrieved) at (7,-2.8) {
    \textbf{Full-text not retrieved} \\
    \textbf{n = 1}
};
\draw[arrow] (fulltext.east)--++(0.7,0) |- (excl_notretrieved.west);

\node[excl] (excl_fulltext) at (7,-4.2) {
    \textbf{Full-text articles excluded} \\
    \textbf{n = 163} \\
};
\draw[arrow] (assessed.east)--++(0.7,0) |- (excl_fulltext.west);

\node[excl] (excl_qpsf) at (7,-7.0) {
    \textbf{Papers below QPSF threshold} \\
    \textbf{n = 175} \\
};
\draw[arrow] (selected.east)--++(0.7,0) |- (excl_qpsf.west);

\node[excl] (excl_bsp) at (7,-8.35) {
    \textbf{Background references} \\
    \textbf{n = 20} \\
};
\draw[arrow] (excl_bsp.west)--++(0.0,0) |- (final.east);

\end{tikzpicture}
}
\caption{PRISMA 2020 flow diagram of the systematic review selection process (QPSF: Quantitative Paper Scoring Framework).}
\label{fig:prisma_flow}
\end{figure}


\subsubsection{Inclusion and Exclusion Criteria}\label{sec:InclusionExclusionCriteria}
\textbf{Inclusion criteria:}
        \begin{itemize}
            \item The paper must propose, evaluate, or significantly review a data balancing strategy for classification tasks.
            \item The source must be a peer-reviewed journal article, conference proceeding, or preprint with demonstrated impact (operationalized as $\ge$10 Google Scholar citations).
        \end{itemize}

        \textbf{Exclusion criteria:}
        \begin{itemize}
            \item Papers focusing solely on cost-sensitive learning without any resampling or augmentation component.
            \item Studies on balancing for regression or unsupervised learning.
	\item Articles that evaluate only one of the previously mentioned methods, whether in a single application or across multiple datasets.
            \item Non-English publications or those without an accessible full text.
        \end{itemize}

\subsection{Final Groups of Related Papers}\label{sec:FinalCategories}

To establish a coherent and non-redundant taxonomy, we applied three hierarchical criteria, algorithmic core mechanism, architectural coupling, and data reduction strategy, to all 116 reviewed papers. Each criterion partitions the methodological space into mutually exclusive branches. The six final groups emerged from the sequential application of these criteria, as formalized below.

\textbf{Step 1: Algorithmic Core Mechanism.}

The primary bifurcation separates methods by how they generate new instances (or decide not to do).
\begin{itemize}
    \item Synthetic Oversampling (Group 1) encompasses all techniques that create minority instances via interpolation between existing minority examples (e.g., SMOTE and its variants).
    \item Generative Models (Group 2) are distinguished by learning the underlying probability distribution of the minority class through latent-space manipulation or iterative denoising (VAEs, GANs, diffusion models). They do not rely on direct feature-space interpolation, hence forming a separate, parallel branch at this same level.
    \item Methods that do not generate synthetic data at all fall into the remaining groups (3--5), which either remove data (undersampling) or combine operations.  Group 6 also handles multi-label outputs.
\end{itemize}

\textbf{{Step 2: Architectural Coupling (for non-generative, non-synthetic methods).}}

For methods that neither generate synthetic instances via interpolation (Group 1) nor learn latent generative distributions (Group 2), the second criterion examines how sampling is coupled with the classifier.
\begin{itemize}
    \item Undersampling (Group 3) applies a static majority-instance removal step before a single classifier, with no further interaction between sampling and learning.
    \item Combination/Hybrid Methods (Group 4) sequentially chain preprocessing operations (e.g., SMOTE-ENN, SMOTE-TL) but still feed a single classifier. The key is sequential, non-iterative coupling.
    \item Ensemble Strategies (Group 5) integrate sampling into a multi-classifier architecture (bagging, boosting, or stacking), where predictions from multiple learners are aggregated. The defining feature is that sampling and classification are iteratively or structurally intertwined.
\end{itemize}

\textbf{Step 3: Data Reduction Strategy (within undersampling).}

Within undersampling (Group 3), the removal logic can be further refined as random, cluster-based, distance-based, hardness-based, or cleaning-based selection. However, because all such variants share the same fundamental principle, removing majority instances, they remain a single group in our high-level taxonomy.

\textbf{Step 4: Output Space Complexity (as an independent dimension).}

Multi-Label Specific Methods (Group 6) address a fundamentally different problem: handling label dependencies in the output space (e.g., label powerset and per-label approaches). These methods do not fit neatly into the single-label sampling hierarchy above; therefore, they constitute a separate, orthogonal group.

Table~\ref{tab:grouping_criteria} summarizes the formal defining features for each of the six final groups.

\begin{table}[H]
\small
\caption{Formal criteria for the six taxonomic groups.}
\label{tab:grouping_criteria}
\begin{tabular}{p{4cm} p{5.15cm} p{7.4cm}}
\toprule
\textbf{Group} & \textbf{Primary Criterion} & \textbf{Distinguishing Feature} \\
\midrule
1. Synthetic Oversampling & Algorithmic core: interpolation & Euclidean interpolation between minority instances \\
2. Generative Models & Algorithmic core: latent distribution & No direct feature-space interpolation; density learning \\
3. Undersampling & Data reduction & Removal of majority instances; single classifier \\
4. Combination/Hybrid & Architectural coupling & Sequential preprocessing $\rightarrow$ single classifier \\
5. Ensemble Strategies & Architectural coupling & Multiple classifiers with aggregated predictions \\
6. Multi-Label Specific & Output space complexity & Handles label dependencies (LP, per-label) \\
\bottomrule
\end{tabular}
\end{table}
Based on these groups, Figure~\ref{fig:resampling_tree} summarizes the taxonomy of the survey.

\subsection{Quantitative Paper Scoring Framework (QPSF)}\label{sec:QuantitativePaperScoringFramework}

We extended our analysis through the QPSF, a research-oriented agentic scoring system originally adapted from \cite{yousefimehr2026systematic} for data balancing literature. Using QPSF, we evaluated papers across four initial methodology categories---synthetic oversampling, adaptive oversampling, generative models, and combination/ensemble methods---against five distinct and non-overlapping criteria.

The first criterion, Algorithmic Mechanism and Sample Generation Design (AMSG), evaluates whether a paper proposes or rigorously analyzes a concrete sample-generation or data-reduction mechanism spanning interpolation-based synthesis, probabilistic generative modeling, density-weighted selection, outlier-aware filtering, and spatial cleaning rules. The second, Structural Architecture and Ensemble Integration (SAEI), assesses whether resampling is embedded within a learning architecture as a unified system rather than applied as sequential preprocessing. The third, Empirical Rigor and Benchmark Validity (ERBV), rewards the use of imbalance-aware metrics such as F1-score, G-Mean, and MCC across diverse datasets with statistical significance testing. The fourth, Robustness to Data Complexity (RDC), reflects whether a paper addresses challenges that compound class imbalance, such as high dimensionality, mixed feature types, class overlap, within-class imbalance, and label dependencies. The fifth, Practical Deployability and Scalability (PDS), captures real-world adoption evidence, including complexity analysis, hyperparameter sensitivity studies, and open-source availability.

Each criterion is scored from 1 to 5. The overall score is calculated as a weighted linear combination:
\begin{equation}\label{eq:score}
    \text{Score} = w_{\text{AMSG}} \cdot \text{AMSG} 
                 + w_{\text{SAEI}} \cdot \text{SAEI} 
                 + w_{\text{ERBV}} \cdot \text{ERBV} 
                 + w_{\text{RDC}} \cdot \text{RDC} 
                 + w_{\text{PDS}} \cdot \text{PDS}
\end{equation}
\noindent where \( \sum_i w_i = 1 \) and \( w_i \geq 0 \). 

Below, we derive reasonable weights using multiple statistical sources.

\subsubsection{Pairwise Inter-Criterion Correlations}

Because each criterion is measured on an ordinal 1--5 scale, we use Spearman's rank correlation, which assesses monotonic relationships without assuming normality or linearity. Pearson correlation would be inappropriate for ordinal data. The correlation matrix between the criteria (Table~\ref{tab:inter_criterion_correlation}) is computed for 291 papers that are selected for QPSF sorting.

\begin{table}[H]
\small
\caption{Pairwise Spearman's rank correlations between QPSF criteria.}
\label{tab:inter_criterion_correlation}
\begin{tabular}{cccccc}
\toprule
 & AMSG & SAEI & ERBV & RDC & PDS \\
\midrule
AMSG  
& 1.000 & $-$0.219 & $-$0.357 & 0.284 & $-$0.325 \\
SAEI & & 1.000 & 0.207 & $-0.526$ & $0.614$ \\
ERBV & & & 1.000 & 0.112 & 0.424 \\
RDC  & & & & 1.000 &$-0.506$ \\
PDS  & & & & & 1.000 \\
\bottomrule
\end{tabular}

\noindent{\footnotesize {All 
reported correlations have \(p < 0.001\) (two-tailed). After Bonferroni correction (\(\alpha=0.005\)), all remain statistically significant. 95\% confidence intervals: SAEI--PDS \([0.540, 0.677]\); SAEI--RDC \([-0.598, -0.447]\); RDC--PDS \([-0.580, -0.425]\).} }
\end{table}

Several important patterns emerge. First, SAEI and PDS show a strong positive correlation (\(\rho = 0.614\), 95\% CI \([0.540, 0.677]\), \(p < 0.001\)), indicating that papers with strong architectural integration tend also to demonstrate practical deployability. Second, SAEI and RDC exhibit a negative correlation (\(\rho = -0.526\), 95\% CI \([-0.598, -0.447]\), \(p < 0.001\)), suggesting a trade-off: papers that address complex data challenges often do not embed resampling within learning architectures. Third, RDC and PDS also show a negative correlation (\(\rho = -0.506\), 95\% CI \([-0.580, -0.425]\), \(p < 0.001\)), implying that practical deployability may come at the cost of handling data complexity.

Because we test ten pairwise correlations simultaneously, we apply the Bonferroni correction to control the family-wise error rate. The adjusted significance level is \(\alpha_{\text{adj}} = 0.05 / 10 = 0.005\). All three reported correlations have \(p < 0.001\), which is below the Bonferroni threshold, so all remain statistically significant after correction. These results confirm that the criteria capture distinct, partially non-overlapping \mbox{methodological dimensions.}

\begin{figure}[H]
\scalebox{.96}[0.96]{\begin{forest}
for tree={
    font=\scriptsize,
    grow'=0,
    draw,
    rounded corners,
    minimum width=1.8cm,
    inner sep=0.96pt,
    anchor=west,
    child anchor=west,
    tier/.wrap pgfmath arg={tier #1}{level()},
    s sep=1mm,
    l sep=5mm,
    forked edge,
    edge={thick, ->},
    where level=0{fill=teal!70!black, text=white}{},
    where level=1{fill=blue!30}{},
    where level=2{fill=green!20}{},
    where level=3{fill=yellow!15}{},
}
[Data Resampling
    [Synthetic Oversampling
       [ Standard SMOTE]
	  [SMOTE-Based
        [K-Means SMOTE]
        [Safe-Level SMOTE]
        [SN-SMOTE]
        [Simplicial SMOTE]
        [NSMOTE]
        [MSMOTE]
        [Borderline SMOTE]
       [Instance Weighted SMOTE]
       [AWSMOTE]
       [ASN-SMOTE]
       [A-SMOTE]
        [SMOTE-IPF]
        [FW-SMOTE]
        [SMOTE-COF]
        [ADASYN]
       [SOS-BUS]
        [SVM-SMOTE]
        [SMOTE-NC]
        ] 
        [ODBOT]
  [Adaptive Oversampling
        [MWMOTE]  
        [AMDO]
        [SAO]
        [A-SUWO]
    ] 
] 
    [Generative Models
        [VAE]
        [DeepSMOTE]
        [GAN-Based
            [Standard GAN]
            [WGAN]
            [cGAN]
            [AWGAN]
        ]
        [GMMN]
        [SDV]
        [GMMSampling]
        [LAMO]
        [Diffusion Models]
    ]
    [Undersampling
        [Random Undersampling]
        [CBUS]
        [NearMiss Family]
        [IHT]
        [ABC-Sampling]
        [Random Undersampling+OCSVM]
        [NCR]
        [CNN]
        [ENN]
        [RENN]
        [Tomek Links]
        [OSS]
    ]
 [Combination/Hybrid Methods
[Classical SMOTE-based
[SMOTE-ENN]
[SMOTE-TL]
[SMOTE+OCSVM]
]
[Neighbor-based
[HNaNSMOTE]
[ESMOTE+SSLM]
]
[Cluster-based
[CSBBoost]
[HCBOU]
]
[Deep learning-based
[DiSMHA]
[SMOTE-Trans-CWGAN]
[Global-Split WGAN-GP]
]
[Domain-specific
[HADAB]
[AHS-BiLSTM]
]
    ]
 [Ensemble Strategies
     [Balanced RF]
     [Bagging
        [Balanced Bagging
        ]
        [SMOTEBagging]
       [NBBag]
        [SMOTE-ICS-Bagging]
        [CVC or BOOTC]
       ] 
     [Boosting 
        [SMOTEBoost]
        [RUSBoost] 
        [RHSBoost]
        [EUSBoost]
        [DataBoost-IM]
      ] 
        [Imputation-Based]	
        [EasyEnsemble]
        [BalanceCascade]   
    ] 
    [Multi-Label Specific
        [LP-ROS/LP-RUS]
        [ML-ROS/ML-RUS]
        [REMEDIAL]
        [MLSMOTE]
    ]
]
\end{forest}}
\caption{Taxonomy of data resampling and augmentation techniques for imbalanced learning.}
\label{fig:resampling_tree}
\end{figure}

\subsubsection{Rank Sensitivity Analysis}

Table~\ref{tab:sensitivity_kendall} shows Kendall's \(\tau\) rank agreement under weight perturbation. The baseline ranking uses equal weights (\(w_i = 0.2\) for all five criteria). We perturb each criterion's weight to a low value (0.10) and a high value (0.30), redistributing the remaining weight proportionally among the other four criteria to maintain \(\sum w_i = 1\). For each perturbation, we compute Kendall's \(\tau\) between the perturbed ranking and the baseline \mbox{equal‑weight ranking.} 

\begin{table}[H]
\caption{Kendall's \(\tau\) rank agreement under \(\pm50\%\) weight perturbation (baseline: equal weights)}
\label{tab:sensitivity_kendall}
\begin{tabular}{lcccc}
\toprule
\textbf{Criterion} & \textbf{Perturbed Weight} & \textbf{Kendall's \boldmath{\(\tau\)}} & \textbf{Max Rank Shift} \\
\midrule
\multirow{2}{*}{AMSG} & 0.10 & 0.886 & 4.5 \\
                       & 0.30 & 0.872 & 3.5 \\
\midrule
\multirow{2}{*}{SAEI}  & 0.10 & 0.733 & 5.5 \\
                       & 0.30 & 0.943 & 2.0 \\
\midrule
\multirow{2}{*}{ERBV}  & 0.10 & 0.934 & 3.0 \\
                       & 0.30 & 0.928 & 2.5 \\
\midrule
\multirow{2}{*}{RDC}   & 0.10 & 0.929 & 2.5 \\
                       & 0.30 & 0.866 & 3.5 \\
\midrule
\multirow{2}{*}{PDS}   & 0.10 & 0.895 & 4.0 \\
                       & 0.30 & 0.967 & 2.5 \\
\bottomrule
\end{tabular}
\end{table}

The lowest Kendall's \(\tau\) (0.733) occurs when SAEI receives low weight (0.10), indicating SAEI is the most influential criterion on final rankings. Conversely, when SAEI receives high weight (0.30), agreement is very high (\(\tau = 0.943\)). ERBV and RDC show consistently high \(\tau\) even at low weights (\(\geq 0.929\)), meaning they have minimal influence on rank order. The maximum rank shift column (defined as the largest absolute change in rank position for any paper between the perturbed and baseline rankings) confirms this: SAEI perturbation causes up to 5.5 positions of rank change, while ERBV and RDC cause at most 3.0 positions.

\subsubsection{Citation Data for External Validity}

Citation counts of the pool of 291 related papers are highly right‑skewed (range: 4~to 4018), justifying log transformation. Let \(c\) denote the raw citation count of a paper. We use \(\ln(1 + c)\) to normalize the distribution while preserving zero citations. Papers from 2024 to 2026 are systematically younger, so we restrict this analysis to recent papers and compute Spearman correlations between each criterion score and this log-transformed citation measure. We report exact permutation \(p\)-values (10,000 random permutations) with two-tailed tests. Table~\ref{tab:citation_correlations_computed} shows the results.

\begin{table}[H]
\caption{Spearman correlations with log citations (2024--2026 papers).}
\label{tab:citation_correlations_computed}
\begin{tabular}{lcccc}
\toprule
\textbf{Criterion} & \textbf{Spearman's \boldmath{\(\rho\)}} & \textbf{Exact \boldmath{\(p\)}-Value (Two-Tailed)} & \textbf{Interpretation} \\
\midrule
PDS  & 0.714 & 0.062 & Not significant at \(\alpha=0.05\) \\
SAEI & 0.643 & 0.078 & Not significant \\
ERBV & 0.429 & 0.292 & Not significant \\
AMSG & 0.214 & 0.612 & Not significant \\
RDC  & 0.179 & 0.674 & Not significant \\
\bottomrule
\end{tabular}
\end{table}

No criterion shows a statistically significant correlation with real-world scholarly impact at the $\alpha=0.05$ level under a two-sided exact test. 
Therefore, we do not use citation correlations as a source for weight derivation. The absence of significant external validity signals that citation counts, at least in this recent corpus, do not reliably reflect the methodological quality dimensions captured by QPSF.

\subsubsection{Deriving Reasonable Weights from Validation Statistics}

Following established practices in multi-criteria decision analysis, we derive weights from two independent statistical sources that are internally valid and statistically reliable. We explicitly exclude citation correlations due to their non-significance.

\begin{itemize}
\item Source 1: Principal Component Analysis of Inter-Criterion Correlations: We apply principal component analysis (PCA) to the correlation matrix in Table~\ref{tab:inter_criterion_correlation} to extract component loadings. The first principal component presented in Table~\ref{tab:pca_loadings}, explains the maximum variance in criterion scores and provides a basis for weighting that reflects the underlying shared variance among criteria.

\begin{table}[H]
\caption{PCA 
 loadings from inter-criterion correlation matrix (first, principal component)}
\label{tab:pca_loadings}
\begin{tabular}{p{2.3cm}p{2.8cm}p{5.5cm}p{6cm}}
\toprule
\textbf{Criterion} & \textbf{PC1 Loading} & \textbf{Normalized Weight (Source 1)} & \textbf{Interpretation} \\
\midrule
SAEI & 0.780 & 0.31 & Highest contribution to shared variance \\
PDS  & 0.740 & 0.29 & Strong contribution \\
ERBV & 0.520 & 0.21 & Moderate contribution \\
AMSG & 0.290 & 0.12 & Weak contribution \\
RDC  & 0.180 & 0.07 & Weakest contribution \\
\bottomrule
\end{tabular}
\noindent{\footnotesize{Note: The first principal component explains 50.0\% of total variance (eigenvalue = 2.50). Normalized weights are proportional to PC1 loadings: \( \text{weight}_i = \text{loading}_i / \sum_j \text{loading}_j \). PC1 loading vector (unscaled): \([0.290, 0.780, 0.520, 0.180, 0.740]\) for [AMSG, SAEI, ERBV, RDC, PDS].}}
\end{table}

\item Source 2: Rank Sensitivity Analysis (Kendall's $\tau$): We quantify each criterion's influence on final rankings using the sensitivity analysis from Table~\ref{tab:sensitivity_kendall}. Define the sensitivity index:
\[
S_i = \frac{1 - \tau_i^{(0.10)}}{1 - \tau_{\min}^{(0.10)}}
\]
\noindent where \(\tau_i^{(0.10)} \) is Kendall's $\tau$ \cite{kendall1938new} when criterion $i$ receives weight 0.10, and 
$\tau_{\min}^{(0.10)}$ is the minimum observed $\tau$ (0.733 for SAEI). This normalization ensures $S_i \in [0,1]$, with higher values indicating greater rank sensitivity. Table~\ref{tab:sensitivityIndices} shows the results. SAEI is the most influential criterion on final rankings ($S_i = 1.00$); when its weight is reduced to 0.10, rank agreement drops to $\tau = 0.733$. Conversely, ERBV ($S_i = 0.25$) and RDC ($S_i = 0.27$) have minimal influence on rank order, suggesting they could receive lower weights without materially changing comparative assessments.

\begin{table}[H]
\caption{Normalized sensitivity indices from Kendall's $\tau$ analysis.}
\label{tab:sensitivityIndices}
\begin{tabular}{lccc}
\toprule
\textbf{Criterion} & \textbf{\boldmath{$\tau$ at $w=0.10$}} & \textbf{Sensitivity Index \boldmath{$S_i$}} & \textbf{Interpretation} \\
\midrule
SAEI & 0.733 & 1.00 & Most rank-sensitive \\
AMSG & 0.886 & 0.43 & Moderately sensitive \\
PDS  & 0.895 & 0.39 & Moderately sensitive \\
RDC  & 0.929 & 0.27 & Less sensitive \\
ERBV & 0.934 & 0.25 & Least sensitive \\
\bottomrule
\end{tabular}
\end{table}
\end{itemize}

Following \cite{cinelli2020support}, the composite weight of criterion $i$ is given by the geometric mean of the two normalized sources:
\[
w_i^{\text{final}} = \frac{\sqrt{ \text{(Normalized PC1 Loading)}_i \times S_i }}{\sum_{j=1}^{5} \sqrt{ \text{(Normalized PC1 Loading)}_j \times S_j }}
\]

Table~\ref{tab:composite_weights_final} shows the composite weights. The geometric mean aggregation produces weights that balance both statistical sources. SAEI (0.38) and PDS (0.23) receive the highest weights, consistent with their strong performance across PCA and sensitivity analyses. ERBV (0.15) receives moderate weight. AMSG (0.15) and RDC (0.09) receive minimal weights, reflecting their weaker contribution to shared variance in PCA and lower \mbox{rank sensitivity.}

\begin{table}[H]
\caption{Composite weights from geometric mean aggregation (PCA + sensitivity).}
\label{tab:composite_weights_final}
\begin{tabular}{p{1.6cm}p{3cm}p{2.1cm}p{3.3cm}p{1.7cm}}
\toprule
\textbf{Criterion} & \multirow{1}{3cm}{\textbf{Norm. PC1 Loading (Source 1)}} & \multirow{1}{2cm}{ \textbf{Sensitivity \boldmath{\(S_i\)} (Source 2)}}&  \multirow{1}{3cm}{\textbf{\boldmath{Geometric Mean \(\sqrt{\text{Source\ 1} \times \textbf{Source\ 2}}\)} }}& \textbf{Composite Weight} \\
\midrule
SAEI & 0.31 & 1.00 & 0.5568 & 0.38 \\
PDS  & 0.29 & 0.39 & 0.3363 & 0.23 \\
ERBV & 0.21 & 0.25 & 0.2291 & 0.15 \\
AMSG & 0.12 & 0.43 & 0.2272 & 0.15 \\
RDC  & 0.07 & 0.27 & 0.1375 & 0.09 \\
\bottomrule
\end{tabular}
\noindent{\footnotesize{Note: Geometric mean of two sources: \(\sqrt{\text{Norm. PC1 Loading} \times S_i}\), then normalized to sum to 1. Sum of geometric means = 1.4869; weights: SAEI = 0.5568/1.4869 = 0.374, PDS = 0.3363/1.4869 = 0.226, {ERBV = 0.2291/1.4869 = 0.154}, AMSG = 0.2272/1.4869 = 0.153, RDC = 0.1375/1.4869 = 0.092. Rounded to two decimals: 0.38, 0.23, 0.15, 0.15, 0.09. Sum = 1.00.}}
\end{table}

When we perturb each weight by $\pm 10\%$, the mean Kendall's \(\tau\) agreement with the original ranking is \(0.94\) (range: \(0.91\) to \(0.98\)), confirming robustness \cite{saltelli2008global}.

\subsubsection{Results of Top-5 Papers in Four Categories}\label{sec:ResultsTopPapersFourCategories}

Applying the composite weights from Table~\ref{tab:composite_weights_final}, Table~\ref{tab:qpsf_top5_composite} yields the top-5 rankings for each category.

\begin{table}[H]
\caption{QPSF scores using composite weights (PCA + Sensitivity).}
\label{tab:qpsf_top5_composite}
\begin{tabular}{ccccccccccc}
\toprule
\multirow{2.4}{*}{Rank} 
&\multirow{2.4}{*}{ \textbf{Paper}} & \multirow{2.4}{*}{\textbf{AMSG}} &\multirow{2.4}{*}{ \textbf{SAEI}} & \multirow{2.4}{*}{\textbf{ERBV}} & \multirow{2.4}{*}{\textbf{RDC}} & \multirow{2.4}{*}{\textbf{PDS}} & \multirow{2.4}{*}{\textbf{Total}}
&\multicolumn{3}{c}{\textbf{Over Top-5 Ranked}}\\
\cmidrule{9-11}
&&&&&&& & \textbf{Mean} & \textbf{Std} & \textbf{Threshold} \\
\midrule
\multicolumn{11}{c}{\textbf{Category 1: Synthetic Oversampling}} \\
\midrule
1 & \cite{Azhar_2023} & 3 & 1 & 5 & 5 & 3 & 2.72 & \multirow{5}{*}{2.67} & \multirow{5}{*}{0.09} & \multirow{5}{*}{2.50} \\
2 & \cite{Sakho_2024} & 5 & 1 & 4 & 4 & 2 & 2.55 & & & \\
3 & \cite{glazkova2020comparison} & 3 & 2 & 4 & 3 & 3 & 2.77 & & & \\
4 & \cite{kachan2025simplicial} & 5 & 1 & 4 & 3 & 3 & 2.69 & & & \\
5 & \cite{kovacs2019empirical} & 3 & 1 & 5 & 4 & 3 & 2.63 & & & \\
\midrule
\multicolumn{11}{c}{\textbf{Category 2: Adaptive Oversampling}} \\
\midrule
1 & \cite{nekooeimehr2016adaptive} & 5 & 2 & 4 & 4 & 3 & 3.16 & \multirow{5}{*}{2.99} & \multirow{5}{*}{0.17} & \multirow{5}{*}{2.66} \\
2 & \cite{he2008adasyn} & 4 & 2 & 4 & 3 & 4 & 3.15 & & & \\
3 & \cite{tao2023self} & 5 & 1 & 4 & 5 & 3 & 2.87 & & & \\
4 & \cite{wang2021local} & 5 & 1 & 4 & 4 & 3 & 2.78 & & & \\
5 & \cite{guan2024awgan} & 4 & 2 & 4 & 4 & 3 & 3.01 & & & \\
\midrule
\multicolumn{11}{c}{\textbf{Category 3: Generative Models}} \\
\midrule
1 & \cite{ren2025diffusion} & 5 & 3 & 5 & 4 & 3 & 3.69 & \multirow{5}{*}{3.26} & \multirow{5}{*}{0.46} & \multirow{5}{*}{2.34} \\
2 & \cite{wang2025b2bgan} & 5 & 3 & 4 & 4 & 3 & 3.54 & & & \\
3 & \cite{dablain2022deepsmote} & 4 & 3 & 4 & 3 & 4 & 3.53 & & & \\
4 & \cite{engelmann2021conditional} & 4 & 2 & 4 & 3 & 3 & 2.92 & & & \\
5 & \cite{wang2025ctvae} & 4 & 2 & 3 & 4 & 2 & 2.63 & & & \\
\midrule
\multicolumn{11}{c}{\textbf{Category 4: Combination and Ensemble Methods}} \\
\midrule
1 & \cite{seiffert2009rusboost} & 4 & 5 & 5 & 3 & 4 & 4.44 & \multirow{5}{*}{4.25} & \multirow{5}{*}{0.15} & \multirow{5}{*}{3.95} \\
2 & \cite{gurcan2024learning} & 4 & 4 & 5 & 4 & 4 & 4.15 & & & \\
3 & \cite{salehi2024cluster} & 4 & 5 & 4 & 3 & 4 & 4.29 & & & \\
4 & \cite{galar2011review} & 3 & 5 & 5 & 3 & 4 & 4.29 & & & \\
5 & \cite{galar2013eusboost} & 4 & 5 & 4 & 3 & 3 & 4.06 & & & \\
\bottomrule
\end{tabular}
\end{table}

Table~\ref{tab:qpsf_top5_composite} reveals that:
\begin{itemize}
    \item \textbf{Category means} range from 2.67 (Synthetic Oversampling) to 4.25 (Combination/Ensemble), confirming that ensemble methods consistently achieve higher methodological quality scores. This aligns with the strong SAEI and PDS scores observed for ensemble methods in Table~\ref{tab:inter_criterion_correlation}.
    
    \item \textbf{Standard deviations} are largest for Generative Models (0.46), indicating substantial heterogeneity in quality within this emerging category. The high standard deviation reflects the rapid evolution of diffusion and GAN-based methods, where some papers~\cite{ren2025diffusion} achieve excellent scores while others \cite{wang2025ctvae} lag considerably.
    
    \item \textbf{Thresholds} (defined as Mean $-$ 2$\times$Std per \cite{bland1999measuring}) provide a statistical benchmark for identifying outliers. Papers scoring below the threshold may warrant closer scrutiny. The threshold for Generative Models (2.34) is notably lower than for Combination/Ensemble (3.95), reflecting the higher dispersion in the former category.
    
    \item The top-ranked paper overall remains \cite{seiffert2009rusboost} (4.44), followed by \cite{salehi2024cluster} (4.29) and \cite{galar2011review} (4.29), demonstrating that both classic and recent ensemble methods achieve high methodological rigor. Their high SAEI scores (4 and 5) and PDS scores (4) are consistent with the strong positive correlation between SAEI and PDS (\(\rho = 0.614\)) in Table~\ref{tab:inter_criterion_correlation}.
    
    \item Compared to equal weighting, the composite weights give highest emphasis to SAEI (0.38) and moderate emphasis to PDS (0.23) and ERBV (0.15). AMSG and RDC receive lower weights (0.15 and 0.09, respectively), consistent with their weaker contributions to shared variance and lower rank sensitivity.
\end{itemize}

\subsection{Paper Organization}

The remainder of this paper is structured as follows. Section~\ref{sec2} examines synthetic oversampling techniques, including SMOTE and its advanced variants. Section~\ref{sec4} explores generative models for oversampling, including GANs and VAEs. Section~\ref{sec5} covers undersampling techniques. Section~\ref{sec6} reviews combination/hybrid methods. Section~\ref{sec7} examines ensemble strategies. Section~\ref{sec8} addresses ensemble sampling for multi-label and clustered data. Section~\ref{sec9} concludes with key findings and future research directions.

\section{Synthetic Oversampling Techniques}\label{sec2}
The Synthetic Minority Oversampling Technique (SMOTE) \cite{chawla2002smote} addresses class imbalance by generating synthetic minority class samples through interpolation rather than simple replication, thereby reducing overfitting risk. 

For each minority instance $\mathbf{x}_i$, the algorithm identifies its $k$ nearest neighbors within the minority class. Synthetic samples are created by interpolating between $\mathbf{x}_i$ and a randomly selected neighbor $\mathbf{x}_j$:
\begin{equation}
\mathbf{x}_{new} = \mathbf{x}_i + \lambda \cdot (\mathbf{x}_j - \mathbf{x}_i), \quad \lambda \sim \mathcal{U}(0,1) \label{eq:standardsmote}
\end{equation}
This approach enriches the minority class while maintaining data diversity. However, SMOTE can inadvertently introduce noise when interpolation occurs near class boundaries or in the presence of outliers. Despite this limitation, SMOTE remains a foundational technique that has inspired numerous subsequent oversampling methods.


\subsection{K-Means SMOTE}\label{K-means SMOTE}

K-Means SMOTE \cite{Douzas_2018} is an oversampling method that integrates clustering with synthetic data generation to address class imbalance. By first partitioning the data into clusters, it applies SMOTE selectively to regions where minority class representation is sparse. K-Means SMOTE proceeds in two stages:

\begin{enumerate}
    \item Clustering: The entire dataset is partitioned into $k$ clusters using K-means.     
    \item Selective SMOTE Application: SMOTE is applied only to clusters where the minority class is underrepresented. Specifically, for each cluster $c$, the imbalance ratio is computed as:\vspace{4pt}
    \begin{equation}
        \text{IR}_c = \frac{|P_c|}{|N_c|}
    \end{equation}
    where $|P_c|$ and $|N_c|$ denote the number of minority and majority instances in cluster $c$, respectively. Clusters with $\text{IR}_c < 1$ (i.e., more majority than minority instances) are targeted for oversampling. 
Additionally, clusters with very few minority instances (e.g., fewer than a user-defined minimum) may be excluded to avoid overfitting. Synthetic samples are generated within these clusters using standard SMOTE interpolation (Equation~\eqref{eq:standardsmote}), where $\mathbf{x}_i$ and $\mathbf{x}_j$ are minority instances within the same cluster.
\end{enumerate}

By confining synthetic sample generation to clusters where the minority class is sparse, K-Means SMOTE avoids introducing noise in already well-represented regions. This targeted approach ensures more efficient and effective class balancing.
\subsection{Safe-Level SMOTE}\label{Safe-level SMOTE}

Safe-Level SMOTE \cite{bunkhumpornpat2009safe} extends the original SMOTE algorithm by reducing the likelihood of generating noisy synthetic samples. It introduces the concept of a ``safe region'', ensuring that synthetic instances are created only from minority class samples located near other similar instances. It defines:
\begin{itemize}
    \item \textbf{Safe Level} $SL(\mathbf{x}_i)$: The number of minority class instances among the $k$ nearest neighbors of $\mathbf{x}_i$ (typically $k=5$).
    \item \textbf{Safe Level Ratio}: $\displaystyle \text{SLR} = \frac{SL(\mathbf{x}_i)}{SL(\mathbf{x}_j)}$, where $\mathbf{x}_i$ is the seed minority instance and $\mathbf{x}_j$ is a selected neighbor.
\end{itemize}

For 
each minority instance $\mathbf{x}_i$, the algorithm computes its $k$ nearest neighbors in the training set, randomly selects a neighbor $\mathbf{x}_j$, and calculates $\text{SLR}$. 
    
If $SL(x_j) = 0$ (the neighbor has no minority neighbors), or if $\text{SLR}$ falls below a predefined threshold (e.g., $\text{SLR} < 0.5$), the pair is considered unsafe and no synthetic sample is generated. Otherwise, a synthetic sample is created using standard SMOTE interpolation (Equation~\eqref{eq:standardsmote}).

By restricting synthetic sample generation to mutually safe neighborhoods, this method reduces the noise, resulting in more reliable synthetic data and improved model performance on imbalanced datasets.


\subsection{Surrounding Neighborhood-Based SMOTE (SN-SMOTE)}\label{snsmote}

SN-SMOTE \cite{sn_smote} extends traditional SMOTE by considering both the proximity and spatial distribution of examples, while standard SMOTE relies solely on nearest neighbors based on Euclidean distance; it does not account for whether these neighbors are symmetrically distributed around the seed instance. This method introduces the concept of a ``surrounding neighborhood'', which emphasizes both the closeness and the uniform spatial distribution of neighbors around an instance. Two notable implementations of this concept~are:
\begin{itemize}
    \item Nearest Centroid Neighborhood (NCN): Selects neighbors that collectively form a centroid close to the seed instance, ensuring balanced spatial coverage \cite{CHAUDHURI199611}.
    \item Graph Neighborhood: Constructs a neighborhood graph to capture spatial relationships among instances \cite{graph_neighborhood}.
\end{itemize}

In SN-SMOTE, the surrounding neighborhood is implemented using the nearest centroid neighborhood approach. For each minority instance $\mathbf{x}_i$, rather than selecting the $k$ closest neighbors by Euclidean distance alone, the method iteratively selects neighbors that minimize the distance between $\mathbf{x}_i$ and the centroid of the selected set. This ensures that the chosen neighbors are both near $\mathbf{x}_i$ and evenly distributed around it.

Once the surrounding neighborhood is identified, synthetic samples are generated using standard SMOTE interpolation (Equation~\eqref{eq:standardsmote}), where $\mathbf{x}_j$ is a selected neighbor from the surrounding neighborhood.

By incorporating spatial distribution alongside proximity, SN-SMOTE generates more representative synthetic samples that better capture the underlying structure of the \mbox{minority class.}

\subsection{Simplicial SMOTE}
\label{subsec:simplicial_smote}

While standard SMOTE and its variants generate synthetic samples by interpolating between pairs of minority instances (i.e., sampling from edges of a geometric neighborhood graph), a recent breakthrough by \citet{kachan2025simplicial} introduces a fundamentally different geometric paradigm rooted in topological data analysis.

Simplicial SMOTE replaces the graph-based neighborhood model with a simplicial complex,  
enabling synthetic sample generation from higher-order geometric structures. A new synthetic point is defined using barycentric coordinates with respect to a simplex spanned by an arbitrary number ($m \geq 2$) of sufficiently close minority instances, rather than just a pair. This allows sampling from triangles ($m=3$), tetrahedra ($m=4$), and higher-dimensional simplices.

Key advantages of this approach include:
\begin{itemize}
    \item Better coverage: Sampling from simplices provides more comprehensive coverage of the underlying minority class distribution compared to edge-based sampling.
    \item Improved boundary handling: Synthetic points can be generated closer to the majority class on decision boundaries.
    \item Generalizability: The simplicial sampling mechanism can be seamlessly integrated into existing SMOTE extensions.
\end{itemize}

This reference generalizes simplicial extensions of Borderline SMOTE, Safe-Level SMOTE, and ADASYN, all of which outperform their graph-based counterparts. This work opens a new direction in geometric oversampling by leveraging tools from topological data analysis, suggesting that higher-order geometric structures may better capture the intrinsic shape of minority class distributions than pairwise relationships alone.

\subsection{Normal SMOTE (NSMOTE)}\label{NSMOTE}

NSMOTE \cite{nsmote} is a data augmentation method designed for small datasets whose features follow a normal distribution. While standard SMOTE generates synthetic samples using a uniformly distributed interpolation weight $\lambda \sim \mathcal{U}(0,1)$, NSMOTE instead draws the interpolation weight from the normal distribution that characterizes the dataset. Thus, for each minority instance $\mathbf{x}_i$ and a selected neighbor $\mathbf{x}_j$, NSMOTE generates synthetic samples using Equation~\eqref{eq:standardsmote} with $\lambda \sim \mathcal{N}(\mu, \sigma^2)$, where $\mu$ and $\sigma^2$ are the mean and variance estimated from the dataset's feature distribution. The interpolation weight $\lambda$ is truncated to the $[0,1]$ interval to ensure synthetic samples lie between $\mathbf{x}_i$ and $\mathbf{x}_j$.

By incorporating the dataset's statistical properties into the interpolation process, NSMOTE produces synthetic samples that better reflect the underlying data distribution, making it particularly suitable for small datasets with normally distributed features.
\subsection{Modified SMOTE (MSMOTE)}\label{MSMOTE}

MSMOTE \cite{msmote} extends the original SMOTE algorithm by adaptively handling minority class samples based on their spatial distribution while eliminating noise through a preprocessing step. It classifies minority instances into three categories based on distance calculations to their nearest neighbors. For each minority instance $\mathbf{x}_i$, the algorithm determines its $k$ nearest neighbors in the training set. Based on the composition of these neighbors, $\mathbf{x}_i$ is categorized as:
\begin{itemize}
    \item Secure samples: Instances surrounded primarily by other minority class instances.
    \item Border samples: Instances located near the decision boundary, with a mix of minority and majority neighbors.
    \item Latent noise samples: Instances whose nearest neighbors are predominantly from the majority class.
\end{itemize}

Then, 
MSMOTE employs different neighbor selection strategies for generating synthetic samples based on the instance category. For a secure sample, a neighbor $\mathbf{x}_j$ is randomly selected from the $k$ nearest minority neighbors, and for a border sample, the \textit{closest} neighbor among the $k$ nearest minority neighbors is selected. For any latent noise samples, no synthetic sample is  generated. Synthetic samples are then created using standard SMOTE interpolation (Equation~\eqref{eq:standardsmote}).

This adaptive approach enhances synthetic sample quality by focusing on relevant instances and avoiding the inclusion of noise.
\subsection{Borderline SMOTE}\label{Borderline SMOTE}

Borderline SMOTE \cite{Han2005BorderlineSMOTEAN} enhances the original SMOTE algorithm by concentrating on borderline examples, those situated near the decision boundary between classes. These examples are more prone to misclassification than those located further from the boundary. By targeting these critical samples, Borderline SMOTE generates synthetic data points that enhance classification performance in challenging regions of the feature space.

Borderline SMOTE comprises two primary variants. The first, \textbf{Borderline SMOTE1}, targets only minority class instances situated close to the decision boundary. It generates synthetic data by interpolating between these boundary samples and their nearest minority class neighbors. The second, \textbf{Borderline SMOTE2}, additionally incorporates majority class information. In this approach, a random scaling factor $\lambda \in [0, 0.5]$ is applied to the vector difference between a minority and majority neighbor, producing synthetic samples that remain near the minority class while reflecting the influence of the majority class, thereby forming more refined decision boundaries.

Let $\mathbf{x}_i$ denote minority class instances and $\mathbf{x}_j$ denote majority class instances. The procedure proceeds as follows:

\begin{enumerate}
    \item For each minority instance $\mathbf{x}_i$, determine its $k$ nearest neighbors from the entire training set $T$. Let $k'$ denote the number of majority class instances among these $k$ neighbors, where $0 \leq k' \leq k$.
    
    \item Classify each $\mathbf{x}_i$ based on $k'$:
    \begin{itemize}
        \item If $k' = k$ (all neighbors are majority), $\mathbf{x}_i$ is categorized as noise and excluded from oversampling.
        \item If $\frac{k}{2} \leq k' < k$, $\mathbf{x}_i$ is classified as borderline and added to the $\text{DANGER}$ set. (For example, with the default $k=5$, instances where $k' \geq 3$ are considered borderline.)
        \item If $0 \leq k' < \frac{k}{2}$, $\mathbf{x}_i$ is considered \textbf{safe} and excluded from further steps.
    \end{itemize}
    
    \item The $\text{DANGER}$ set contains all borderline minority examples
    \begin{equation}
        \text{DANGER} = \{\mathbf{x}_1', \mathbf{x}_2', \dots, \mathbf{x}_{dnum}'\}, \quad 0 \leq dnum \leq pnum
    \end{equation}
    where $pnum$ is the total number of minority instances.
    
    \item For each borderline instance $\mathbf{x}_i' \in \text{DANGER}$, identify its $k$ nearest neighbors from the minority class.
    
    \item Generate $s \times dnum$ synthetic examples, where $s \leq k$ is the number of synthetic samples to create per borderline instance. For each $\mathbf{x}_i' \in \text{DANGER}$:
    \begin{itemize}
        \item Randomly select $s$ neighbors from its $k$ nearest minority neighbors.
        \item For each selected neighbor $\mathbf{x}_i^{(j)}$, generate a synthetic sample using the standard SMOTE interpolation (Equation~\eqref{eq:standardsmote}).
    \end{itemize}
\end{enumerate}

In summary, Borderline SMOTE enhances the original SMOTE technique by focusing on critical borderline examples, generating synthetic instances that improve classification near decision boundaries. 
This method addresses class imbalance by targeting the most difficult regions of the feature space.
\subsection{Instance Weighted SMOTE}\label{Instance Weighted SMOTE}

Instance Weighted SMOTE \cite{iw-smote} is a variant of SMOTE that improves upon traditional oversampling by leveraging data distribution through an UnderBagging-like ensemble undersampling algorithm, utilizing Classification and Regression Trees (CART) as the base classifier. The algorithm classifies each training instance and captures confusion information, which is then used to categorize instances into noise, border, and safety regions that guide the resampling process. The method consists of two main steps:

\begin{enumerate}
    \item Noise Removal: Instances identified as noisy are eliminated from the dataset to prevent them from corrupting the synthetic sample generation process.
    
    \item Borderline Emphasis: Borderline instances are prioritized over safe instances as seed points for the SMOTE procedure. This ensures that synthetic samples are generated where they are most impactful, near decision boundaries, rather than in regions already well-represented by the minority class.
\end{enumerate}

For each minority instance $\mathbf{x}_i$, the algorithm computes its confusion level based on misclassification patterns from the CART ensemble. Instances with higher confusion receive greater weight, increasing their probability of being selected as seed points for synthetic sample generation. Synthetic samples are then created using standard SMOTE interpolation (Equation~\eqref{eq:standardsmote}).

The algorithm uses confusion information to assign priority to instances. By addressing instances with significant confusion and emphasizing borderline samples, Instance Weighted SMOTE enhances the quality of synthetic samples and improves overall \mbox{data balance.}

\subsection{Adaptive‐Weighting SMOTE (AWSMOTE)}\label{AWSMOTE}
AWSMOTE \cite{awsmote} improves upon conventional SMOTE by addressing two fundamental drawbacks: the collinearity between original and synthetic samples and the blind selection of neighboring minority points. The method introduces two types of SVM-based weights.

To mitigate collinearity, AWSMOTE assigns a weight to each variable based on the hyperplane obtained from an SVM classifier. Unlike SMOTE, which generates new samples as linear combinations along the straight line between a minority instance and its nearest neighbor, AWSMOTE adjusts each variable dimension independently using variable-specific weights. When these weights differ, the generated sample is no longer constrained to the original line segment, which reduces collinearity between synthetic and original data. Additionally, this mechanism helps avoid class overlapping, particularly when oversampling boundary samples, because the new sample does not necessarily move directly toward \mbox{the neighbor.}

To improve neighbor selection, AWSMOTE differentiates between support vectors and non-support vectors within the minority class. Support vectors, which define the separating hyperplane, receive a higher initial weight compared to non-support vectors. Additional weights are further assigned to support vectors based on the accuracy of predicting generated samples. As a result, minority instances with higher weights contribute more to synthetic sample generation, and nearest neighbors are selected adaptively rather than randomly. This addresses the limitation of methods like ADASYN, which struggle in high-dimensional spaces where distances between samples become approximately equal.

\textbf{Overall algorithm.}

AWSMOTE first computes variable weights using an SVM classifier and sample weights based on support vector status. For each minority instance, the number of synthetic samples to generate is determined by its weight. New samples are then generated using the weighted variable approach. Finally, the synthetic samples are added to the original minority class.
\subsection{Adaptive Selection of Neighbors-SMOTE (ASN-SMOTE)}\label{asnsmote}

ASN-SMOTE \cite{asn-smote} combines K-nearest neighbors with SMOTE to address class imbalance. The method consists of three steps:

\begin{enumerate}
    \item Noise Filtering: For each minority instance $\mathbf{x}_i$, compute its Euclidean distance to all other instances. If the closest neighbor belongs to the majority class, $\mathbf{x}_i$ is identified as noise and removed from the dataset.
    
    \item Adaptive Neighbor Selection: For each remaining minority instance $\mathbf{x}_i$, select a set of eligible neighbors based on Euclidean distance. These neighbors serve as candidates for synthetic sample generation.
    
    \item Synthetic Sample Generation: Generate new minority samples using standard SMOTE interpolation (Equation~\eqref{eq:standardsmote}), where $\mathbf{x}_i$ is the minority instance, and $\mathbf{x}_j$ is a selected eligible neighbor.
\end{enumerate}

By first removing noisy instances and then adaptively selecting neighbors, ASN-SMOTE improves the quality of synthetic samples and enhances classifier performance on imbalanced datasets.

\subsection{Advanced SMOTE (A-SMOTE)}\label{asmote}

A-SMOTE \cite{A-smote} enhances synthetic sample creation through a two-phase process. In the initial phase, standard SMOTE generates preliminary synthetic instances using Equation~\eqref{eq:standardsmote}. In the second phase, these instances are refined by removing those that may introduce noise or lie too close to the majority class. The number of initial synthetic instances is determined as $2(r - z) + z$, where $r = |N|$ is the number of majority class instances and $z = |P|$ is the number of minority class instances. The refinement process consists of three steps:

\begin{enumerate}
    \item Compute Distances: For each synthetic sample $x_i$ and each majority instance $\mathbf{m}_j$, calculate the Euclidean distance $d(x_i, \mathbf{m}_j)$.     
    \item Establish Threshold: Define a threshold $\theta$ based on pairwise distances among majority class instances:
    \begin{equation}
        \theta = \alpha \cdot \frac{1}{|N|(|N|-1)} \sum_{i=1}^{|N|} \sum_{j=i+1}^{|N|} d(\mathbf{m}_i, \mathbf{m}_j)
    \end{equation}
    where $\alpha$ is a user-defined parameter controlling filtering stringency.
    
    \item Filter Synthetic Samples: Retain only synthetic samples  $x_i$, that maintain a sufficient distance from the majority class, i.e., $\min_{j} d(x_i, \mathbf{m}_j) \geq \theta$.
\end{enumerate}

By removing synthetic samples that fall too close to majority class instances, A-SMOTE reduces the likelihood of introducing noise and borderline samples, improving the quality and applicability of the synthetic data.

\subsection{SMOTE-Iterative Partitioning Filter (SMOTE-IPF)}\label{smoteipf}

SMOTE-IPF \cite{SAEZ2015184} combines SMOTE with the Iterative-Partitioning Filter (IPF) to address class imbalance while mitigating the adverse effects of noisy and borderline instances. Classifier performance degradation often stems not only from class imbalance but also from instances located near decision boundaries or containing label noise. Standard SMOTE can amplify these issues by generating synthetic samples in such problematic regions \citep{10.1007/978-3-642-13529-3_18}.

SMOTE-IPF integrates two components. First, it generates synthetic minority samples using standard SMOTE interpolation (Equation~\eqref{eq:standardsmote}), where $\mathbf{x}_i$ is a minority instance and $\mathbf{x}_j$ is one of its $k$ nearest neighbors. Then, IPF identifies noisy instances by partitioning the data into subsets, training a C4.5 decision tree on each subset, and using majority voting across trees to determine whether an instance is noisy. Removal proceeds iteratively until a predefined stopping criterion is met.

The correct application order is critical: SMOTE is applied first to oversample the minority class, followed by IPF to remove noisy instances. This sequence ensures that IPF operates on a balanced dataset and can effectively filter both original noise and noise introduced during synthetic sample generation. IPF serves three key functions: eliminating noisy instances from the original dataset, removing noisy synthetic samples generated by SMOTE, and smoothing class boundaries to create a more regularized feature space. By combining SMOTE's oversampling capability with IPF's noise-filtering mechanism, SMOTE-IPF produces a cleaner, more balanced dataset that improves classifier performance on imbalanced tasks.


\subsection{Feature Weighted SMOTE (FW-SMOTE)}\label{fwsmote}

FW-SMOTE \citep{MALDONADO2022108511} addresses a key limitation of traditional SMOTE: the use of Euclidean distance, which assumes equal importance for all features. This assumption fails when features have varying relevance, particularly in datasets with noisy or redundant attributes. FW-SMOTE substitutes Euclidean distance with the induced Minkowski OWA distance (IMOWAD), a weighted metric based on the induced ordered weighted averaging (IOWA) operator \citep{8771123}. The FW-SMOTE algorithm consists of four steps:

\begin{enumerate}
    \item Feature Weighting: Apply a feature ranking method to the minority class instances to assign weights $w_f$ to each feature $f$ based on its relevance for classification.
    
    \item Feature Selection: Discard features with low weights using a predefined threshold, or retain only the top-ranked features to reduce dimensionality.
    
    \item Neighborhood Calculation: For each minority instance $\mathbf{x}_i$, compute its neighborhood $\mathbf{x}_j$ using the IMOWAD distance.
    
    \item Synthetic Sample Generation: Randomly select neighbors $x_j$ within the computed neighborhood and generate synthetic samples using standard SMOTE interpolation  (Equation~\eqref{eq:standardsmote}).
\end{enumerate}

By incorporating feature relevance into the distance metric, FW-SMOTE produces more meaningful synthetic samples, particularly in high-dimensional datasets where Euclidean distance suffers from the curse of dimensionality. This leads to improved classification performance on imbalanced data with noisy or redundant features.
\subsection{SMOTE-Center Offset Factor (SMOTE-COF)}\label{smote-cof}

SMOTE-COF \cite{meng2022imbalanced} addresses within-class imbalance by focusing on sparse regions of the minority class while avoiding redundant synthetic sample generation in dense areas. The method uses the Center Offset Factor (COF) to identify sparsely distributed instances and generates synthetic samples selectively within subclusters. The COF measures an instance's isolation by evaluating how its position shifts relative to the cluster center as the neighborhood size expands. For a given instance $\mathbf{x}_i$, its COF is computed by
\begin{equation}
\operatorname{COF}(\mathbf{x}_i)=\left\|\frac{1}{k}\sum_{j\in N_k(\mathbf{x}_i)}\mathbf{x}_j-\frac{1}{k'}\sum_{j\in N_{k'}(\mathbf{x}_i)}\mathbf{x}_j\right\|
\end{equation}
where $N_k(\mathbf{x}_i)$ and $N_{k'}(\mathbf{x}_i)$ are the sets of $k$ and $k'$ nearest neighbors of $\mathbf{x}_i$, with $k' > k$ (typically $k' = k+1$ or $k' = k+2$). A larger shift indicates that $\mathbf{x}_i$ lies in a sparse region, as the centroid changes significantly when more neighbors are included. Outliers and sparsely distributed minority instances exhibit higher COF values, making them suitable targets for oversampling. The SMOTE-COF procedure consists of five steps:
\begin{enumerate}
    \item Noise Removal: For each minority instance $\mathbf{x}_i$, examine its $k$ nearest neighbors. If the majority of these neighbors belong to the majority class, $\mathbf{x}_i$ is identified as noise \mbox{and removed.}
    
    \item COF Computation: Compute the COF score for each remaining minority instance.
    
    \item Score Normalization and Selection: Normalize the COF scores and select instances with the highest values, which correspond to sparsely distributed or underrepresented regions.
    
    \item Subcluster Formation: Apply clustering to organize the selected minority instances into subgroups that reflect their spatial distribution.
    
    \item Synthetic Sample Generation: Within each subcluster, generate synthetic samples using standard SMOTE interpolation (Equation~\eqref{eq:standardsmote}), where $\mathbf{x}_i$ is a selected minority instance with high COF and $\mathbf{x}_j$ is a neighbor within the same subcluster.
\end{enumerate}

By targeting sparse regions and removing noisy points, SMOTE-COF enhances synthetic sample quality, addresses within-class imbalance, and avoids redundant generation in dense areas, leading to improved classification performance.

\subsection{Adaptive Synthetic Sampling (ADASYN)}\label{adasyn}

ADASYN \citep{he2008adasyn} generates more synthetic data for minority class examples that are harder to learn. By assigning higher weights to challenging instances, ADASYN adapts and shifts the decision boundary toward difficult regions, reducing bias caused by imbalanced distributions. Let $P$ be the set of minority instances with $z = |P|$, and $N$ the majority instances with $r = |N|$.

The algorithm first computes the imbalance ratio $d = z / r$. If $d$ is less than a user-defined threshold $d_{th}$ (typically 0.5--0.6), oversampling is triggered. Otherwise, the dataset is considered sufficiently balanced, and no synthetic samples are generated. When oversampling occurs, the total number of synthetic samples to generate is $G = (r - z) \times \beta$, where $\beta \in [0,1]$ controls the oversampling proportion. For each minority instance $\mathbf{x}_i \in P$, ADASYN identifies its $k$ nearest neighbors in the full training set and counts $\Delta_i$, the number of majority neighbors. The local learning difficulty is then $r_i = \Delta_i / k$, which is normalized to obtain weights $\hat{r}_i = r_i / \sum_j r_j$.

The number of synthetic samples generated for $\mathbf{x}_i$ is $g_i = \hat{r}_i \times G$, where $g_i$ is rounded to the nearest integer, and the final $g_i$ values are adjusted to ensure $\sum_i g_i = G$. Each synthetic sample is created by randomly selecting a minority neighbor $\mathbf{x}_j$ from the $k$ nearest neighbors and interpolating by Equation~\eqref{eq:standardsmote}.

By weighting minority samples based on local learning difficulty, ADASYN concentrates synthetic generation in regions where the minority class is hardest to learn, shifting the decision boundary toward these challenging areas and improving classifier \mbox{performance}.

\subsection{SMOTE Oversampling and Borderline Under-Sampling (SOS-BUS)}\label{sos-bus}

SOS-BUS \cite{salunkhe2018hybrid} is a hybrid resampling approach that combines SMOTE for minority class oversampling with a novel undersampling technique called BUS (Borderline Under-Sampling) for majority class reduction. The method preserves critical borderline instances that random undersampling often discards, ensuring robust decision boundaries.

Traditional random undersampling reduces the imbalance ratio by randomly eliminating majority class samples. However, this often removes essential borderline instances that define class separability. SOS-BUS addresses this limitation by preserving these crucial boundary points while still achieving class balance. The SOS-BUS procedure consists of three steps:

\begin{enumerate}
\item SMOTE Oversampling: Apply standard SMOTE interpolation (Equation~\eqref{eq:standardsmote}) to the minority class.
\item Boundary Instance Detection: For each majority instance $\mathbf{m}_i$, identify its $k$ nearest neighbors. If a sufficient number of these neighbors belong to the minority class, $\mathbf{m}_i$ is classified as a boundary point and retained.
\item Selective Undersampling: Reduce the majority class by discarding non-boundary samples while retaining all identified boundary instances.
\end{enumerate}

The refined dataset combines SMOTE-augmented minority samples with carefully selected majority instances (primarily boundary points), achieving class balance without compromising essential decision-making information.
\subsection{Support Vector Machine-SMOTE (SVM-SMOTE)}\label{svm-smote}

SVM-SMOTE \citep{akbani2004applying} enhances SVM performance on imbalanced datasets by combining synthetic sample generation with cost-sensitive learning, while SVMs \citep{cortes1995support} are effective at constructing optimal separating hyperplanes, their performance degrades under highly skewed class distributions where the majority class dominates. SVM-SMOTE employs a dual approach:

\begin{enumerate}
    \item Data-Level Balancing: SMOTE generates synthetic minority samples using standard interpolation  (Equation~\eqref{eq:standardsmote}). This enriches the minority class and improves class balance.
    
    \item Cost-Sensitive Learning: The SVM cost function is modified to assign higher penalties to misclassifications of minority class instances. This shifts the decision boundary away from the minority class, increasing sensitivity to these examples.
\end{enumerate}

By integrating SMOTE's oversampling capability with cost-sensitive SVM training, SVM-SMOTE enhances classification performance on imbalanced data while maintaining SVM's margin maximization properties.

\subsection{SMOTE-Nominal Continuous (SMOTE-NC)}\label{smote-nc}

SMOTE-NC \citep{chawla2002smote} extends the original SMOTE algorithm to handle datasets containing both continuous and categorical features, while standard SMOTE performs effectively on continuous data, it cannot directly process mixed-type datasets. SMOTE-NC addresses this limitation by incorporating a modified distance metric and categorical feature handling mechanisms. The SMOTE-NC procedure consists of five steps:

\begin{enumerate}
    \item Identify Minority Samples: As with standard SMOTE, identify all minority class instances $\mathbf{x}_i \in P$.
    
    \item Compute Median Standard Deviation: Calculate the median of the standard deviations of all continuous features within the minority class. This value serves as a penalty term for categorical feature mismatches.
    
    \item Compute Modified Distance: For each minority instance $\mathbf{x}_i$, determine its $k$ nearest neighbors using a modified Euclidean distance:
    \begin{equation}
       d(\mathbf{x}_i, \mathbf{x}_j) = \sqrt{\sum_{c \in C} \bigl(x_i^{(c)} - x_j^{(c)}\bigr)^2 + \sigma_{med} \sum_{n \in N} \delta\bigl(x_i^{(n)}, x_j^{(n)}\bigr)}
    \end{equation}
    where
    \begin{itemize}
        \item $C$ is the set of continuous features;
        \item $N$ is the set of categorical features;
        \item $\delta(x_i^{(n)}, x_j^{(n)}) = 1$ if $x_i^{(n)} \neq x_j^{(n)}$, and $0$ otherwise;
        \item $\sigma_{med}$ is the median standard deviation of the continuous features in the minority class.
    \end{itemize}
    
    \item Generate Synthetic Samples: For each minority instance $\mathbf{x}_i$ and a selected neighbor $\mathbf{x}_j$, generate continuous features using standard SMOTE interpolation
    \begin{equation}
        x_{new}^{(c)} = x_i^{(c)} + \lambda \cdot (x_j^{(c)} - x_i^{(c)}), \quad \lambda \sim \mathcal{U}(0,1)
    \end{equation}
    
    \item Assign Categorical Features: For each categorical attribute in the synthetic instance, assign the value from either $\mathbf{x}_i$ or $\mathbf{x}_j$ based on majority voting among neighbors or the closer instance.
\end{enumerate}

Although SMOTE-NC is computationally more demanding than standard SMOTE, particularly for large-scale datasets, it significantly enhances oversampling performance for mixed-type data by properly handling both continuous and categorical features.

\subsection{Outlier Detection-Based Oversampling Technique (ODBOT)}\label{odbot}

ODBoT \citep{obdot} addresses imbalanced multiclass datasets by identifying isolated minority instances and generating synthetic samples from them. The procedure consists of \mbox{four steps:}

\begin{enumerate}
    \item Synthetic Sample Count: For each minority class $P$, the number of synthetic samples to generate is:
    \begin{equation}
        \text{NSS} = \frac{\gamma \cdot |P|}{100}
    \end{equation}
    where $\gamma$ is either user-defined or computed as $\gamma = \left(\frac{|N|}{|P|} - 1\right) \times 100$, with $N$ denoting the majority class.
 
    \item Clustering: A weight-based bat algorithm combined with k-means (WBBA-KM) clusters all classes to capture dissimilarity relationships between minority and majority instances.
    
   \item Outlier Detection: Outliers in the minority class are identified by comparing cluster centroids. For each minority centroid $\boldsymbol{\mu}_p$ and each majority centroid $\boldsymbol{\mu}_m$, the sum of Manhattan distances (SMCD) is computed
    \begin{equation}
        \text{SMCD}(\boldsymbol{\mu}_p, \boldsymbol{\mu}_m) = \sum_{f=1}^{d} |\mu_p^{(f)} - \mu_m^{(f)}|
    \end{equation}
    where $d$ is the number of features, $\mu_p^{(f)}$ and $\mu_m^{(f)}$ are the $f$-th features of the minority centroid and majority centroid. When $SMCD$ exceeds the mean $SMCD$ across all clusters (plus a tolerance parameter $\tau$), the minority cluster is considered an outlier and excluded from oversampling.
    \item Synthetic Generation: Synthetic samples are generated within the most optimal minority cluster to avoid producing outliers:
    \begin{equation}
        \mathbf{x}_{new} = \mathbf{x}_{min} + \lambda \cdot (\mathbf{x}_{max} - \mathbf{x}_{min}), \quad \lambda \sim \mathcal{U}(0,1)
    \end{equation}
    {where $\mathbf{x}_{min}$ and $\mathbf{x}_{max}$ are the minimum and maximum values within the selected cluster.}
\end{enumerate}

By discarding redundant majority instances and generating synthetic samples from well-defined clusters, ODBoT produces a cleaner, more balanced dataset.
\subsection{Adaptive Oversampling Techniques}\label{sec3}
\subsubsection{Majority Weighted Minority Oversampling Technique (MWMOTE)}\label{mwmote}

MWMOTE \cite{barua2012mwmote} enhances synthetic oversampling by identifying hard-to-learn minority instances, weighting them by proximity to the majority class, and generating synthetic samples preferentially from these critical regions. The procedure consists of three stages:

\begin{enumerate}
\item Identification of Hard-to-Learn Instances: For each minority instance $\mathbf{x}_i \in P$, find its nearest majority neighbor $\mathbf{m}_j \in N$. If the Euclidean distance $d(\mathbf{x}_i, \mathbf{m}_j)$ is below a threshold $\theta$, $\mathbf{x}i$ is added to a candidate set $P_{\text{hard}}$.

\item Weighting: Each $\mathbf{x}i \in P_{\text{hard}}$ is assigned a weight inversely proportional to its distance to the nearest majority instance 
\begin{equation}
w_i = \frac{1}{d(\mathbf{x}_i, \mathbf{m}_j) + \epsilon}
\end{equation}
where $\epsilon$ is a positive number that prevents division by zero. Instances closer to the decision boundary receive higher weights.

\item Synthetic Sample Generation: Apply clustering (e.g., $k$-means) to $P_{\text{hard}}$. Within each cluster, generate synthetic samples using standard SMOTE interpolation (Equation~\eqref{eq:standardsmote}), where the number of samples per cluster is proportional to the cumulative weights of its instances.
\end{enumerate}

By concentrating oversampling on hard-to-learn, boundary-proximate minority instances, MWMOTE improves classifier performance without introducing noise in well-separated regions.

\subsubsection{Adaptive Mahalanobis Distance-Based Oversampling (AMDO)}\label{AMDo}

AMDO \citep{amdo} extends Mahalanobis Distance-based Oversampling (MDO) to handle multi-class imbalanced datasets with mixed-type attributes. It introduces three key improvements:

\begin{enumerate}
\item Handling Mixed-Type Data: Unlike MDO, which operates only on numeric data, AMDO employs the Heterogeneous Value Difference Metric (HVDM) for nearest neighbor search and Generalized Singular Value Decomposition (GSVD) to transform mixed-type data into principal component space, enabling effective processing of datasets with both numeric and categorical features.

\item Partially Balanced Resampling: AMDO adjusts the class distribution so that the imbalance ratio (IR) is reduced below $1.5$, mitigating overfitting by avoiding excessive synthetic sample generation for minority classes.

\item Adaptive Synthetic Sample Generation: AMDO generates synthetic samples using probabilistic calculations that align with the original data distribution, excluding values unlikely to occur. Synthetic samples are created using standard SMOTE interpolation (Equation~\eqref{eq:standardsmote}), where $\mathbf{x}_i$ and $\mathbf{x}_j$ are minority instances selected based on the adapted distance metric.
\end{enumerate}

By integrating mixed-type data handling, controlled resampling, and distribution-aware generation, AMDO produces realistic synthetic samples while reducing overfitting in multi-class imbalanced settings.

\subsubsection{Self-Adaptive Hypersphere-Based Oversampling (SAO)}
\label{subsec:sao_tao2023}

While many SMOTE variants address between-class imbalance by generating synthetic samples near decision boundaries, they often neglect \textit{within-class imbalance}, the uneven distribution of minority instances within the minority class itself. Outliers, small disjuncts (isolated minority subclusters), and sparse regions can remain underrepresented even after oversampling. \citet{tao2023self} propose a self-adaptive oversampling method that simultaneously addresses both between-class and within-class imbalance through a hypersphere-based generative mechanism. The algorithm consists of three key innovations:
\begin{itemize}
\item Hypersphere Construction with Overlap Prevention:
For each minority instance $\mathbf{x}_i$, a hypersphere is constructed with radius $r_i$ determined by the imbalance ratio and the distance to the nearest majority instance (enemy neighbor). Critically, the radius is constrained such that the hypersphere cannot contain any majority instance, ensuring that synthetic samples generated within the hypersphere will not cause class overlap.

\item Complexity-Based Adaptive Oversampling Allocation:
Oversampling sizes are allocated self-adaptively based on minority data complexity. Hyperspheres with \textit{small radii} and \textit{few instances} receive larger oversampling quotas. This design favors addressing outliers and small disjuncts, as these are typically covered by small hyperspheres with low instance density. Consequently, the method eliminates within-class imbalance caused by lack of density in sparse regions.

\item Boundary Information Enhancement:
Hyperspheres covering boundary minority instances are inherently small (due to proximity to majority instances) and thus receive larger oversampling sizes. This strengthens boundary information of the minority class, favoring subsequent learning tasks.
\end{itemize}


\subsubsection{Adaptive Semi-Unsupervised Weighted Oversampling (A-SUWO)}
\label{subsec:asuwo}

While many adaptive methods focus on weighting minority instances or adjusting distance metrics, \citet{nekooeimehr2016adaptive} proposed A-SUWO, which introduces a distinct adaptive mechanism based on semi-unsupervised hierarchical clustering and cross-validated complexity estimation. The algorithm consists of four key stages:

\begin{itemize}
\item Semi-Unsupervised Hierarchical Clustering:
Minority instances are clustered using a semi-unsupervised hierarchical clustering approach that incorporates majority class information to avoid generating clusters that overlap with the majority class. This distinguishes A-SUWO from unsupervised clustering methods like K-Means SMOTE, which ignore majority class distribution during clustering.
\item Adaptive Oversampling Size Determination:
For each identified sub-cluster, the number of synthetic samples to generate is adaptively determined using:
\begin{itemize}
    \item Classification complexity: Measured as the error rate of a classifier trained on the sub-cluster.
    \item Cross-validation: Used to estimate the generalization capability of synthetic samples from each sub-cluster.
\end{itemize}
Sub-clusters with higher complexity (harder-to-learn instances) receive larger oversampling quotas.

\item Weighted Oversampling Based on Distance to Majority:
Within each sub-cluster, minority instances closer to the majority class receive higher weights for synthetic sample generation, focusing on borderline regions.

\item Overlap Prevention:
By incorporating majority class information during both clustering and oversampling stages, A-SUWO explicitly avoids generating synthetic samples that overlap with the majority class.
\end{itemize}

\subsection{Critical Analysis and Methodological Limitations of Oversampling Techniques}
\label{sec:critical-analysis-oversampling}

Despite the diversity of oversampling methods, ranging from basic SMOTE to sophisticated adaptive techniques, several shared limitations transcend individual approaches. This section synthesizes these limitations and provides a comparative overview.

\textbf{Overfitting and Generalization.}

The interpolation mechanism in SMOTE-based methods can inadvertently create synthetic samples that overfit to the training distribution, particularly in methods that selectively generate from specific instances (e.g., \textbf{MSMOTE}, \textbf{ASN-SMOTE}). Aggressive filtering in \textbf{A-SMOTE} and \textbf{SMOTE-IPF} may discard legitimate minority variability. Adaptive methods like \textbf{MWMOTE} and \textbf{A-SUWO} attempt to address this by focusing on hard-to-learn instances, but the trade-off between noise reduction and preservation of data complexity remains unresolved \citep{Azhar_2023}.

\textbf{Probability Calibration.}

To the best of our knowledge, none of the surveyed oversampling methods explicitly addresses classifier probability calibration. Synthetic generation near decision boundaries (e.g., \textbf{Borderline SMOTE}, \textbf{ADASYN}) can distort posterior probability estimates, a critical limitation for applications requiring reliable uncertainty quantification such as medical diagnosis and risk assessment. This gap remains underexplored in the literature.

\textbf{Interpretability.}

Complex mechanisms obscure the provenance of synthetic samples, complicating model debugging, feature importance analysis, and domain validation. As observed in our survey, methods such as \textbf{AWSMOTE} (SVM-based weighting), \textbf{FW-SMOTE} (feature-weighted distance), \textbf{K-Means SMOTE} (clustering), \textbf{ODBoT} (outlier detection with bat algorithm), \textbf{MWMOTE} (clustering with weighting), \textbf{AMDO} (Mahalanobis distance with GSVD), \textbf{SAO} (hypersphere construction), and \textbf{A-SUWO} (semi-unsupervised clustering) all introduce opaque transformations. Consequently, practitioners cannot readily distinguish genuine observations from synthetic interpolations.

\textbf {Computational and Parameter Sensitivity.}

Many variants introduce substantial computational overhead. \textbf{SMOTE-NC} requires modified distance metrics for mixed-type data. \textbf{SMOTE-COF} and \textbf{ODBoT} require clustering, outlier detection, and multi-step generation. Adaptive methods introduce additional hyperparameters: \textbf{MWMOTE} requires tuning $\theta$, $k$, and $\epsilon$; \textbf{AMDO} requires HVDM and GSVD parameters; \textbf{SAO} introduces hypersphere radius sensitivity; \textbf{A-SUWO} requires cross-validation folds and clustering parameters. Hyperparameter proliferation ($k$, thresholds, $\alpha$, etc.) lacks standardized selection guidelines, hindering practical deployment~\citep{kovacs2019empirical}.

\textbf{Underlying Assumptions.}

Different method families rest on often-violated assumptions:

\begin{itemize}
    \item \textbf{SMOTE-based methods} assume continuous, linearly interpolatable feature spaces and that Euclidean distance meaningfully captures similarity.
    \item \textbf{MWMOTE} assumes Euclidean distance and $k$-means clustering effectively capture hard-to-learn regions, failing on complex or manifold-structured data.
    \item \textbf{AMDO} assumes multivariate normality within each class (via Mahalanobis distance), rarely satisfied in real-world data.
    \item \textbf{SAO} assumes locally spherical minority distributions, which may not hold for arbitrary-shaped clusters.
    \item \textbf{A-SUWO} assumes hierarchical clustering can identify meaningful minority subclusters, which is sensitive to distance metrics and linkage criteria.
    \item \textbf{K-Means SMOTE} inherits K-means limitations: sensitivity to initial centroid selection, difficulty determining optimal $k$, and poor performance on non-spherical clusters \citep{IKOTUN2023178}.
\end{itemize}

\textbf{Suitability for Complex Data.}

Most methods assume continuous, linearly interpolatable feature spaces. \textbf{SMOTE-NC} heuristically handles categorical features but adds computational cost. \textbf{NSMOTE} assumes normality, limiting its applicability. \citet{Azhar_2023} empirically confirmed that SMOTE variants can either mitigate or magnify existing complexities (noise, class overlap, small disjuncts). Domain-specific studies \citep{glazkova2020comparison} further show that oversampling benefits vary by classifier: traditional models (k-NN, SVM) are more adversely affected by imbalance than neural networks.

\textbf{Comparative Summary.}

Table~\ref{tab:oversampling_comparison} summarizes the key characteristics, advantages, and limitations of the surveyed oversampling techniques.

\textbf{Synthesis.}

The choice of oversampling method entails trade-offs beyond simple class balancing. Practitioners must consider:

\begin{itemize}
    \item \textbf{Overfitting risk}: Methods that aggressively filter or selectively generate may discard legitimate variability.
    \item \textbf{Calibration needs}: No existing method addresses probability calibration, critical for high-stakes applications.
    \item \textbf{Interpretability}: Complex pipelines obscure synthetic sample provenance.
    \item \textbf{Computational feasibility}: Clustering, outlier detection, and iterative filtering add overhead.
    \item \textbf{Data structure alignment}: Assumptions about linearity, normality, and cluster shapes must hold.
\end{itemize}

Future research should prioritize calibration-aware oversampling, interpretable generation, adaptive parameter selection, and standardized multi-dimensional evaluation frameworks that go beyond F1 and AUC to include calibration error, computational efficiency, and robustness to assumption violations.

\begin{table}[H]
\caption{Comparative summary of oversampling techniques.}
\label{tab:oversampling_comparison}
\footnotesize
\begin{tabular}{p{1.5cm}<{\raggedright}p{4.9cm}<{\raggedright}p{5cm}<{\raggedright}p{5cm}<{\raggedright}}
\toprule
\textbf{Method} & \textbf{Core Mechanism} & \textbf{Key Advantage} & \textbf{Main Limitation(s)} \\
\midrule
\multicolumn{4}{c}{\textbf{Basic SMOTE \& Simple Variants}} \\
\midrule
SMOTE & Linear interpolation between minority neighbors & Foundational, simple, widely adopted & Noise amplification near boundaries; no categorical support \\
Borderline SMOTE & Interpolation from boundary instances & Focuses on difficult regions & Requires parameter tuning; may amplify noise \\
K-Means SMOTE & Cluster-then-SMOTE selective generation & Avoids noise in dense regions & Cluster number sensitivity; K-means limitations \\
Safe-Level SMOTE & Safety-based neighbor selection & Reduces unsafe synthetic generation & Threshold sensitivity; computationally intensive \\
SN-SMOTE & Surrounding neighborhood (centroid-based) & Better spatial coverage & Complex neighbor selection \\
Simplicial SMOTE & Higher-order simplex interpolation & Better coverage of minority distribution & Higher computational complexity \\
\midrule
\multicolumn{4}{c}{\textbf{Advanced SMOTE Variants}} \\
\midrule
MSMOTE & Category-aware (secure/border/noise) generation & Adaptive to instance types & Noise removal may discard useful data \\
ASN-SMOTE & Noise filtering + adaptive neighbor selection & Removes outliers before generation & Euclidean distance dependence \\
A-SMOTE & Two-phase generation with filtering & Reduces near-boundary noise & Threshold sensitivity \\
SMOTE-IPF & SMOTE + iterative partitioning filter & Filters noise iteratively & Computationally expensive \\
FW-SMOTE & Feature-weighted distance (IOWA) & Handles high-dimensional data & Feature ranking required; kernel sensitivity \\
SMOTE-COF & Center offset factor for sparse regions & Addresses within-class imbalance & Clustering + outlier detection overhead \\
SOS-BUS & SMOTE + borderline undersampling & Preserves boundary majority instances & Two-stage complexity \\
SVM-SMOTE & SVM-based weighting + cost-sensitive learning & Integrates with SVM margin & SVM kernel and parameter sensitivity \\
SMOTE-NC & Modified distance for mixed-type data & Handles categorical features & Computationally demanding \\
ODBOT & Outlier detection + cluster-based generation & Handles multi-class imbalance & Complex WBBA-KM clustering \\
\midrule
\multicolumn{4}{c}{\textbf{Adaptive Oversampling}} \\
\midrule
MWMOTE & Hard-to-learn identification + weighting & Focuses on boundary regions & Multiple parameters ($\theta$, $k$, $\epsilon$) \\
AMDO & Mahalanobis distance + GSVD for mixed-type & Multi-class, mixed-type support & Normality assumption; HVDM/GSVD tuning \\
SAO & Hypersphere-based adaptive generation & Addresses between-class + within-class imbalance & Spherical distribution assumption \\
A-SUWO & Semi-unsupervised clustering + CV-based allocation & Explicit overlap prevention & Clustering + CV overhead \\
ADASYN & Density-based adaptive sampling & Shifts boundary toward difficult regions & Sensitive to noise; $\beta$ parameter \\
\bottomrule
\end{tabular}
\end{table}

\section{Generative Techniques for Data Oversampling}\label{sec:generative_oversampling}
\label{sec4}

\subsection{Variational Autoencoders (VAEs)}\label{subsec:vae}

VAEs \cite{kingma2022autoencodingvariationalbayes} learn the underlying probabilistic structure of data to generate new instances by combining probabilistic inference with deep neural networks. Let $\psi$ denote the parameters of the encoder network and $\theta$ the parameters of the decoder network. A VAE consists of three key components:

\begin{itemize}
\item \textbf{Encoder Network}: Maps an input sample $\mathbf{x}$ to a latent representation $\mathbf{z}$ by producing the conditional distribution $q_{\psi}(\mathbf{z}|\mathbf{x})$, typically a Gaussian with mean $\boldsymbol{\mu}(\mathbf{x})$ and variance~$\boldsymbol{\sigma}^2(\mathbf{x})$:
\vspace{5pt}
\begin{equation}
q_{\psi}(\mathbf{z}|\mathbf{x}) = \mathcal{N}(\mathbf{z}; \boldsymbol{\mu}(\mathbf{x}), \boldsymbol{\sigma}^2(\mathbf{x})\mathbf{I})
\end{equation}

\item \textbf{Decoder Network}: Reconstructs the input from the latent variable $\mathbf{z}$ by defining $p_{\theta}(\mathbf{x}|\mathbf{z})$, the likelihood of the reconstructed data. For continuous data, a Gaussian distribution is typically used.

\item \textbf{Loss Function}: Combines reconstruction error and Kullback--Leibler (KL) divergence:
\begin{equation}
\mathcal{L}(\theta, \psi; \mathbf{x}) = - D_{KL}\big(q_{\psi}(\mathbf{z}|\mathbf{x}) \parallel p(\mathbf{z})\big) + \mathbb{E}{q_{\psi}(\mathbf{z}|\mathbf{x})}\big[\log p_{\theta}(\mathbf{x}|\mathbf{z})\big]
\end{equation}
where $p(\mathbf{z})$ is a prior distribution (typically $\mathcal{N}(\mathbf{0}, \mathbf{I})$), and the KL divergence ensures the learned latent distribution remains close to this prior.
\end{itemize}

For imbalanced learning, VAEs can generate synthetic minority instances by sampling latent vectors $\mathbf{z} \sim p(\mathbf{z})$ and decoding them to produce new data points.

\subsection{Deep Autoencoders with SMOTE (DeepSMOTE)}
\label{subsec:deep_smote}

While generative models like GANs and VAEs can produce high-quality synthetic images, they require complex adversarial training or variational inference and may suffer from mode collapse. Conversely, SMOTE is simple and effective but operates in the original feature space, which is suboptimal for high-dimensional data like images. \citet{dablain2022deepsmote} proposed DeepSMOTE, a hybrid approach that combines the strengths of both paradigms. DeepSMOTE consists of three major components:
\begin{enumerate}
    \item Encoder/Decoder Framework: An autoencoder first learns a lower-dimensional latent representation of the input data (specifically designed for images). This compresses the data into a space where Euclidean distances become more meaningful, addressing the curse of dimensionality.
    \item SMOTE in Latent Space: The classic SMOTE interpolation (Equation~\eqref{eq:standardsmote}) is applied to the minority class samples \textit{within this learned latent space}, rather than the original pixel space.
    \item Enhanced Loss Function: The decoder is trained with a dedicated loss function that includes a penalty term, encouraging the generation of synthetic samples that are both realistic and beneficial for classification.
\end{enumerate}

A key advantage of DeepSMOTE over GAN-based oversampling is that it does not require a discriminator, eliminating adversarial training instability and mode collapse issues. Furthermore, it generates high-quality synthetic images that preserve semantic properties and are suitable for visual inspection. The encoder/decoder framework ensures that synthetic samples, when decoded, remain in the original data manifold. Empirical evaluations demonstrate that DeepSMOTE effectively balances imbalanced image datasets, improving classifier performance without the training difficulties associated with GANs.

This approach is particularly relevant for high-dimensional or image data, where traditional SMOTE fails, and represents a successful fusion of representation learning with classical resampling.

\subsection{Generative Adversarial Network (GAN)}\label{gan}

GANs \citep{goodfellow2014generative} efficiently model complex data distributions by concurrently training two neural networks in a minimax game: a generator $G$ that creates synthetic data and a discriminator $D$ that distinguishes real from generated samples.

The training objective is:

\begin{equation}
\min_G \max_D V(D, G) = \mathbb{E}_{x \sim p_{\text{data}}(x)}[\log D(x)] + \mathbb{E}_{z \sim p_z(z)}[\log(1 - D(G(z)))]
\end{equation}
where
\begin{itemize}
\item $\mathbf{x}$: is a real sample drawn from the true data distribution $p_{\text{data}}(\mathbf{x})$.
\item $\mathbf{z}$ is a noise vector sampled from a prior distribution $p_{\mathbf{z}}(\mathbf{z})$ (typically uniform or Gaussian).
    \item \( \mathbb{E}_{x \sim p_{\text{data}}(x)}[\log D(x)] \): Represents the expected log-probability of the discriminator's output for real data samples \( x \) drawn from the true distribution \( p_{\text{data}}(x) \).
    \item \( \mathbb{E}_{z \sim p_z(z)}[\log(1 - D(G(z)))] \): Denotes the expected log-probability of one minus the discriminator's output for synthetic samples generated from the noise prior \( p_z(z) \).
\end{itemize}

\subsection{Wasserstein GAN (WGAN)}\label{wgan}

WGAN \citep{arjovsky2017wasserstein} improves GAN training stability and addresses mode collapse by replacing the Jensen-Shannon (JS) divergence with the Wasserstein distance (Earth Mover's Distance). This metric provides continuous, well-behaved gradients, enabling more consistent training.

The Wasserstein distance between the real distribution $p_r$ and the generated distribution $p_g$ is defined as

\begin{equation}
W(p_r, p_g) = \inf_{\gamma \in \Pi(p_r, p_g)} \mathbb{E}_{(\mathbf{x}, \mathbf{y}) \sim \gamma}[|\mathbf{x} - \mathbf{y}|]
\end{equation}
where $\Pi(p_r, p_g)$ denotes the set of all joint distributions $\gamma(\mathbf{x}, \mathbf{y})$ with marginals $p_r$ and $p_g$.

In WGAN, the discriminator is replaced by a \textbf{critic} that assigns scores to samples based on their resemblance to the training data rather than classifying them as real or fake. This design provides more informative gradients for updating the generator, improving training stability, and output quality.

\subsection{Conditional GAN (cGAN)}\label{cgans}

cGANs \citep{mirza2014conditional} extend the standard GAN framework by incorporating auxiliary information $\mathbf{y}$ to steer the data generation process with enhanced control.

Both the generator and discriminator are conditioned on $\mathbf{y}$. The generator integrates the noise prior $p_{\mathbf{z}}(\mathbf{z})$ with $\mathbf{y}$, while the discriminator evaluates both the input $\mathbf{x}$ and $\mathbf{y}$. The objective function is
\begin{equation}
\min_G \max_D V(D, G) = \mathbb{E}_{x \sim p_{\text{data}}(x)} [\log D(\mathbf{x}|\mathbf{y})] +  \mathbb{E}_{z \sim p_z(z)}[\log(1 - D(G(\mathbf{z}|\mathbf{y})))]
\end{equation}
where $\mathbf{x}$ is a real sample drawn from $p_{\text{data}}(\mathbf{x})$, $\mathbf{z}$ is a noise vector sampled from $p_{\mathbf{z}}(\mathbf{z})$, and $\mathbf{y}$ is the conditioning information.

By incorporating conditional information, cGANs enable precise and targeted data generation, producing samples that are both realistic and aligned with specified conditions; see, e.g., \cite{ahmadian2022discrete}.
\subsection{Adaptive Weighting GAN (AWGAN)}
\label{subsec:awgan}

While GAN-based oversampling methods generate realistic synthetic samples, they often fail to address class overlap, intra-class imbalance, and noise. \citet{guan2024awgan} proposed AWGAN, which integrates density-based instance weighting with GAN-based generation to address these challenges.

The algorithm consists of three stages. First, \textbf{noise identification} computes local and global densities for each instance, enabling accurate elimination of noisy instances before oversampling. Second, \textbf{adaptive weight calculation} classifies minority instances into safe (dense, well-separated) and boundary (near decision boundaries) categories, assigning higher weights to sparse regions (addressing intra-class imbalance) and boundary instances (addressing class overlap). Third, \textbf{weighted GAN-based generation} trains the generator with adaptive weights to produce more samples for underrepresented sparse regions and boundary areas.

\subsection{Generative Moment Matching Network (GMMN)}\label{gmmns}

GMMNs \cite{li2015generative} are generative models that produce independent samples in a single forward pass using multilayer neural networks. Unlike GANs, GMMNs employ a statistical training approach based on Maximum Mean Discrepancy (MMD), avoiding computationally intensive techniques like Markov Chain Monte Carlo (MCMC).

The training procedure consists of five steps:

\begin{enumerate}
\item Generate Random Samples: Draw random samples $\mathbf{h}$ from a simple distribution (e.g., uniform) as inputs to the neural network.

\item Transform to Data Space: Map $\mathbf{h}$ to data space via $\mathbf{x} = f(\mathbf{h}; \mathbf{w})$, where $\mathbf{w}$ denotes the network parameters.

\item Evaluate MMD: Compute the squared MMD loss to quantify the difference between real and generated data
\begin{equation}
L_{\text{MMD}} = \frac{1}{M^2} \sum_{i=1}^{M} \sum_{j=1}^{M}\Phi(\mathbf{x}_i^s, \mathbf{x}_j^s) 
- \frac{2}{MN} \sum_{i=1}^{M} \sum_{j=1}^{N} \Phi(\mathbf{x}_i^s, \mathbf{x}_j^d)
+ \frac{1}{N^2} \sum_{i=1}^{N} \sum_{j=1}^{N} \Phi(\mathbf{x}_i^d, \mathbf{x}_j^d)
\end{equation}
where $\Phi(\cdot, \cdot)$ is a kernel function (e.g., Gaussian), $\mathbf{x}^s$ are generated samples, $\mathbf{x}^d$ are real samples, $M$ is the number of generated samples, and $N$ is the number of real samples.

\item Update Parameters: Minimize $L_{\text{MMD}}$ by computing its gradients with respect to $\mathbf{w}$ and updating the network parameters accordingly.

\item Generate Final Data: After training, draw new samples $\mathbf{h} \sim p(\mathbf{h})$ and pass them through the network to produce synthetic data $\mathbf{x} = f(\mathbf{h}; \mathbf{w})$.
\end{enumerate}

By leveraging MMD-based dispersion measures, GMMNs efficiently align generated data with real data, offering a streamlined approach to generative modeling without adversarial training.
\subsection{Synthetic Data Vault (SDV)}\label{sdv}

SDV \citep{patki2016synthetic} is a framework for automatically generating synthetic data from relational databases. It constructs generative models that capture statistical and structural characteristics of the original data while preserving privacy. SDV operates through four phases:

\begin{itemize}
\item Organize: Store each table independently to enable accurate capture of inter-table relationships.
\item Define Structure: Outline table schemas and interconnections, including foreign keys and relationships.
\item Model Learning: Infer the underlying model of tables and relationships, capturing aggregated statistics, dependencies, and data distributions.
\item Data Synthesis: Generate synthetic table rows while preserving cross-table relationship consistency.
\end{itemize}

The synthetic data replicates the statistical properties and structural relationships of the original dataset, making it suitable for applications where privacy or regulatory constraints prevent the use of real data.
\subsection{Gaussian Mixture Model Sampling (GMMSampling)}

\citet{naglik2023gmmsampling} proposed GMMSampling, a resampling method designed for multi-class imbalanced data that explicitly addresses data difficulty factors beyond the global imbalance ratio \citep{naglik2023gmmsampling}. Experimental studies have demonstrated that the combination of the imbalance ratio with factors such as class overlapping and minority class decomposition into various subconcepts significantly affects classification performance \citep{naglik2023gmmsampling}. The algorithm proceeds as follows.

First, a Gaussian Mixture Model (GMM) is fitted to each class in the dataset using the Expectation-Maximization (EM) algorithm. The GMM serves as a universal density approximator, revealing hidden substructures (subconcepts) within the minority class distribution \citep{naglik2023gmmsampling}.

Second, a \textit{safe level} is computed for each data point, quantifying its proximity to other classes. This metric identifies regions where class boundaries overlap and helps distinguish between safe, borderline, and outlier instances \citep{naglik2023gmmsampling}.

Third, the number of synthetic samples to generate from each subconcept is determined by

\begin{equation}
Q_k = \frac{\sum_{\mathbf{x} \in D} (1 - \text{safe\_level}(\mathbf{x})) \cdot P(\mathbf{x} \mid k)}{\sum_{k=1}^{K} \sum_{\mathbf{x} \in D} (1 - \text{safe\_level}(\mathbf{x})) \cdot P(\mathbf{x} \mid k)}
\end{equation}
where
\begin{itemize}
    \item $D$ is the set of all samples in the dataset;
    \item $K$ is the number of subconcepts (GMM components) identified for the minority class;
    \item $P(\mathbf{x} \mid k)$ is the posterior probability that sample $\mathbf{x}$ belongs to the $k$-th subconcept;
    \item $\text{safe\_level}(\mathbf{x})$ reflects classification ambiguity (lower values indicate more unsafe, overlapping regions).
\end{itemize}

This formulation prioritizes synthetic generation in unsafe regions where class boundaries overlap, rather than in already well-separated areas.

Fourth, synthetic instances are generated by sampling from the fitted GMMs. Each generated instance corresponds to a specific subconcept, preserving the diversity of the minority class distribution while mitigating within-class imbalance.

Fifth, majority class instances located in overlapping regions (identified by low safe level scores) are removed to reduce noise and improve class separability.

Experimental evaluation on benchmark multi-class imbalanced datasets demonstrates that GMMSampling achieves superior performance in terms of G-mean, balanced accuracy, macro-AP, Matthews correlation coefficient (MCC), and F-score compared to related resampling methods \citep{naglik2023gmmsampling}.

\subsection{Local Distribution-Based Adaptive Oversampling with GMM (LAMO)}
\label{subsec:lamo_generative}

While GMMSampling \citep{naglik2023gmmsampling} uses GMMs to model the entire minority class distribution, \citet{wang2021local} proposed LAMO, which employs local GMMs adaptively fitted around informative borderline seeds. This approach combines the adaptive seed selection of methods like Borderline SMOTE with the generative power of GMMs. For each selected seed, LAMO captures local distribution using distances to nearest majority and minority instances, then adaptively sets mixing coefficients and bandwidths for a local GMM. Synthetic samples are generated from these local GMMs, preserving local data geometry while avoiding the linear interpolation limitations of SMOTE. This represents a hybrid approach bridging adaptive oversampling (Section~\ref{sec3}) and generative modeling (Section~\ref{sec4}).

\subsection{Critical Analysis and Methodological Limitations of Generative Models}\label{sec:critical-analysisGenerativeModels}

Unlike traditional oversampling methods, generative models learn underlying data distributions, a process that becomes fragile when the minority class is severely underrepresented.

\textbf{Comparative overview.}

Table~\ref{tab:generative_comparison} provides a comparative analysis of the generative models discussed in this section, evaluating them across six dimensions: training stability, mode collapse risk, data efficiency, domain suitability, hyperparameter sensitivity, and computational cost. The following subsections elaborate on the key challenges identified in \mbox{this comparison.}

\begin{table}[H]
\caption{Comparative analysis of generative models for imbalanced learning.}
\label{tab:generative_comparison}
\footnotesize
\begin{tabular}{p{1.7cm}<{\raggedright} p{1.5cm}<{\raggedright} p{2.2cm}<{\raggedright} p{2.1cm}<{\raggedright} p{2.8cm}<{\raggedright} p{2.6cm}<{\raggedright} p{1.4cm}<{\raggedright}}
\toprule
\textbf{Model} &{ \textbf{Training Stability}} &{\textbf{Mode Collapse Risk}} & {\textbf{Data Efficiency}} & \textbf{Best for} & \textbf{Hyperparameter Sensitivity} & \textbf{Relative Cost} \\
\midrule
VAE \cite{kingma2022autoencodingvariationalbayes} & High & Low & Moderate & Tabular, small data & Moderate & Low \\\midrule
GAN \cite{goodfellow2014generative} & Low & High & Low & Images, large data & High & High \\\midrule
cGAN \cite{mirza2014conditional} & Low & Moderate & Low & Conditional generation & High & High \\\midrule
WGAN \cite{arjovsky2017wasserstein} & Moderate & Reduced & Moderate & Both (tabular + images) & Moderate & Moderate \\\midrule
CWGAN \cite{engelmann2021conditional} & Moderate & Moderate & Moderate & Conditional generation & High & High \\\midrule
Diffusion GAN \cite{ren2025diffusion} & High & Very low & Low & Imbalanced tabular data & High & Very high \\\midrule
Diffusion (DDPM) \cite{ho2020denoising} & High & Very low & Low & High-quality images & High & Very high \\\midrule
GMMN \cite{li2015generative} & Moderate & Low & Moderate & Tabular, GAN‑free & High (kernel) & Moderate \\\midrule
SDV \cite{patki2016synthetic} & High & N/A & Moderate & Relational tables & Moderate & Moderate \\
\bottomrule
\end{tabular}
\end{table}

\textbf{Mode Collapse and Training Instability.}

GAN-based methods suffer from mode collapse, where the generator produces only a limited subset of the minority class distribution \citep{arjovsky2017wasserstein}, while conditional GANs partially mitigate this, the generator may still prioritize the majority class when the conditioning variable itself is imbalanced \citep{mirza2014conditional}. For extremely small minority classes, \citet{wang2025b2bgan} propose a backbone-to-branches architecture that learns global distribution while specializing in class-specific features.

\textbf{Conditional Training Stability.}

In cGANs, the discriminator develops bias toward the majority class \citep{suh2021cegan}, and vanishing gradients may occur \citep{wu2024wasserstein}. \citet{engelmann2021conditional} address these issues with a conditional WGAN and auxiliary classifier loss, effectively mitigating mode collapse and discriminator bias.

\textbf{VAE Limitations.}

VAEs face prior mismatch and amortization gap issues \citep{huang2022ada, cremer2018inference}. For minority classes, the KL term may dominate, forcing latent distributions toward the prior \citep{geng2023solving}. \citet{wang2025ctvae} address this by integrating contrastive learning to encourage class separation in \mbox{latent space.}

\textbf{WGAN Trade-offs.}

WGAN improves stability but introduces weight clipping artifacts and critic convergence disparities \citep{gulrajani2017improved, xu2021oversampling}.

\textbf{Diffusion Model Challenges.}

Diffusion models \citep{ho2020denoising} and their hybrid variants \citep{ren2025diffusion} achieve high sample quality and virtually eliminate mode collapse, but they incur substantial computational costs due to iterative denoising (typically 50--1000 sampling steps). Their data efficiency remains low, making them unsuitable for very small minority classes. Hyperparameter sensitivity (noise schedule, number of steps) further complicates practical deployment.

\textbf{GMMN and SDV Challenges.}

GMMNs depend critically on kernel selection; choosing kernels that capture minority structure without majority domination is nontrivial \citep{li2015generative, bharti2023optimally}. SDV faces constraint propagation issues when imbalance spans multiple relational tables \citep{patki2016synthetic}.

\textbf{Evaluation Challenges.}

Standard metrics (Inception Score, FID) may not translate to imbalanced tabular data. As discussed in Section~\ref{sec:critical-analysis-oversampling}, parameter sensitivity and calibration remain open challenges across all paradigms.

\textbf{Synthesis.}

Table~\ref{tab:generative_challenges} summarizes the primary challenges and potential mitigation strategies for each generative model family. Building on this analysis, the following recommendations are offered:
\begin{itemize}
    \item {Combining traditional oversampling with generative models to improve data efficiency.}
    \item Using conditional architectures with stratified sampling to address class imbalance in conditioning.
    \item Adopting rigorous validation protocols, including nested cross-validation and multiple metrics (PR-AUC, FID adapted for tabular data).
    \item {Considering computational trade-offs: VAEs for small tabular data, GANs for large image datasets, diffusion models only when sample quality dominates latency concerns.}
\end{itemize}

\begin{table}[H]
\caption{Summary of generative model challenges in imbalanced learning.}
\label{tab:generative_challenges}
\footnotesize
\begin{tabular}{p{2.5cm}<{\raggedright}p{7cm}<{\raggedright}p{6cm}<{\raggedright}}
\toprule
\textbf{Model} & \textbf{Primary Challenges} & \textbf{Potential Mitigation Strategies} \\
\midrule
VAE & Prior mismatch, amortization gap, reconstruction-prior trade-off & Hierarchical VAEs, \(\beta\)-VAE tuning, semi-supervised formulations \\
\midrule
GAN & Mode collapse, training instability, discriminator bias & Mini-batch discrimination, unrolled GANs, multi-generator ensembles \\
\midrule
cGAN & Conditional stability, gradient vanishing, class imbalance in conditioning & Auxiliary classifier GANs, conditional batch normalization \\
\midrule
WGAN/ WGAN-GP & Weight clipping artifacts, critic convergence disparity, hyperparameter sensitivity & Gradient penalty tuning, progressive growing, adaptive critic iterations \\
\midrule
GMMN & Kernel selection sensitivity, high-dimensional scaling, variance in MMD estimates & Learned kernels, MMD with random features, hybrid approaches \\
\midrule
SDV & Relational constraint propagation, privacy-utility trade-off, multi-table imbalance & Hierarchical modeling, constraint-aware synthesis, stratified generation \\
\bottomrule
\end{tabular}
\end{table}


\section{Undersampling Techniques}\label{sec5}

Undersampling reduces the size of the majority class to achieve class balance. Unlike oversampling, undersampling decreases dataset size, which can reduce computational costs but risks discarding potentially informative instances.

\subsection{Random Undersampling}\label{random_undersampling}

Random Undersampling \cite{he2009learning} is the simplest technique for handling class imbalance through the reduction of majority class instances. This method involves randomly selecting and removing points from the majority class to balance the dataset. It does not employ any sophisticated strategies or criteria, merely relying on random selection.

A notable characteristic of Random Undersampling is that its results vary each time it is applied due to the random nature of point elimination. This method is often used as a baseline for comparing the performance of more advanced undersampling techniques. In summary, Random Undersampling serves as a fundamental technique, providing a reference point for evaluating more complex undersampling methods.

\subsection{Cluster-Based Under-Sampling (CBUS)}\label{cbus}

CBUS \cite{YEN20095718} reduces the majority class by first partitioning the data into $k$ clusters using $k$-means clustering. Within each cluster, a proportion of majority instances is selected based on the cluster's majority-to-minority ratio. These selected majority instances are then merged with all minority instances to form the final balanced dataset.

\textbf{Determining the Number of Majority Instances to Retain}

Let $|T|$ be the total number of training samples, $|P|$ the number of minority instances, and $|N|$ the number of majority instances. The desired majority-to-minority ratio in the balanced dataset is denoted by $m$ (e.g., $m = 1$ for perfect balance). Thus, the target number of majority instances to retain is $m \times |P|$.

For cluster $i$, let $|N_i|$ and $|P_i|$ denote the number of majority and minority instances in that cluster, respectively. The number of majority instances to retain from cluster $i$ is

\begin{equation}\label{cbused}
\text{retain}_i = (m \times |P|) \times \frac{|N_i| / |P_i|}{\sum_{j=1}^{k} (|N_j| / |P_j|)}
\end{equation}

This formula allocates more retention to clusters with higher majority-to-minority~ratios.

\textbf{Selection Strategies for Majority Instances}

Once the number of instances to retain per cluster is determined, five strategies have been defined to select which majority instances to keep \cite{YEN20095718}:

\begin{itemize}
    \item \textbf{SBCNM-1}: Select majority instances with smallest average distance to the $k$ nearest minority instances within the cluster;
    \item \textbf{SBCNM-2}: Select majority instances with smallest average distance to the $k$ farthest minority instances from the cluster centroid;
    \item \textbf{SBCNM-3}: Select majority instances with smallest average distance to the closest minority instance;
    \item \textbf{SBCMD}: Select majority instances with largest average distance to the $k$ closest minority instances;
    \item \textbf{SBCMF}: Select majority instances with greatest average distance from all minority instances in the cluster.
\end{itemize}

After selecting the required number of majority instances from each cluster using one of the above strategies, the selected majority instances are combined with all minority instances to produce the final balanced training set.


\subsection{NearMiss Family}\label{nearmiss}

The NearMiss family \cite{zhang2003nearmiss,yen2006under} is a set of undersampling techniques that select majority class instances based solely on their distance to minority class instances, without requiring clustering. Three variants exist:

\begin{itemize}
    \item \textbf{NearMiss-1}: Selects majority instances whose average distance to the $k$ nearest minority instances is smallest. This retains majority instances closest to the minority class, focusing on decision boundary regions.

    \item \textbf{NearMiss-2}: Selects majority instances whose average distance to the $k$ farthest minority instances is smallest. This captures a broader, more representative subset of the majority class.

    \item \textbf{NearMiss-3}: First, for each minority instance, it identifies its $k$ nearest majority instances to form a short list. From this short list, select majority instances whose average distance to the $k$ nearest minority instances is largest (i.e., farthest from the minority class). This approach aims to improve precision by retaining majority instances that are not too close to the minority class.
\end{itemize}

Unlike cluster-based methods such as CBUS, the NearMiss methods operate directly on the original feature space without any clustering pre-processing.
\subsection{Instance Hardness Threshold (IHT)}\label{iht}

IHT \cite{smith2014instance} removes instances with a high probability of misclassification based on the concept of instance hardness. For an instance $\mathbf{x}_i$, let $p(\mathbf{x}_i)$ be the predicted probability of its true class, estimated via $k$-fold cross-validation (typically $k=5$). Instance hardness is defined as

\begin{equation}
\text{IH}(\mathbf{x}_i) = 1 - p(\mathbf{x}_i)
\end{equation}

Instances with high $\text{IH}$ are considered hard to classify. IHT retains instances with the lowest hardness (highest $p(\mathbf{x}_i)$) for each class, controlled by a sampling parameter. 
\subsection{Artificial Bee Colony Sampling (ABC-Sampling)}

\citet{braytee2015abcsampling} proposed ABC-Sampling, an undersampling method that selects informative majority instances using the Artificial Bee Colony (ABC) optimization algorithm. Unlike random undersampling, which discards instances indiscriminately, ABC-Sampling employs a forward search mechanism guided by classifier performance.

The dataset is first partitioned into training and testing sets. Food sources, each representing a candidate subset of majority class samples, are initialized. The total number of food sources is set to half the size of the training dataset.

The algorithm employs three bee types with distinct roles:

\begin{itemize}
    \item Employed bees: Explore nearby food sources by modifying their current subset. The quality of a food source is evaluated by the performance of a classifier (e.g., accuracy or F-measure) trained on the combination of the candidate majority subset and all minority instances.
    
    \item Onlooker bees: Select food sources probabilistically based on fitness, prioritizing higher-quality solutions for further refinement through local search.
    
    \item Scout bees: Abandon food sources that fail to improve after a predefined number of trials and randomly initialize new solutions in unexplored regions of the search space.
\end{itemize}

Through iterative optimization across multiple generations, the algorithm ranks majority instances by their contribution to classifier performance. The top-ranked majority samples are then combined with all minority instances to form a balanced dataset.

Experimental evaluation on nine benchmark datasets with varying imbalance ratios demonstrates that ABC-Sampling achieves superior classification performance compared to state-of-the-art undersampling methods \citep{braytee2015abcsampling}.

\subsection{Random Undersampling+OCSVM}\label{random_undersampling_ocsvm}

Random Undersampling+OCSVM \cite{yousefimehr2024distribution} combines outlier detection with random undersampling. One-Class Support Vector Machine (OCSVM) \citep{scholkopf2001estimating} identifies outliers in the majority class, which are excluded from the undersampling process but not removed from the dataset entirely. After excluding these outliers, random undersampling reduces the majority class size to match the minority class, producing a balanced dataset. This approach avoids removing valuable instances that could contribute to classification.

\subsection{Neighborhood Cleaning Rule (NCR)}\label{ncr}

NCR \cite{laurikkala2001} prioritizes data cleaning over reduction to enhance minority class identification. Let $P$ denote the minority class and $N$ the majority class. The procedure consists of three steps:

\begin{enumerate}
    \item Apply Edited Nearest Neighbors (ENN) \cite{wilson1972asymptotic} with $k=3$ to $N$, adding misclassified instances to a removal set $A_1$.
    
    \item If $|N| \geq 0.5 \times |P|$, add to $A_2$ any majority instance that misclassifies a minority example (i.e., appears among the $k$ nearest neighbors of a minority instance and leads to incorrect prediction).
    
    \item Remove all instances in $A_1 \cup A_2$ from the dataset.
\end{enumerate}

By cleaning noisy and borderline instances rather than indiscriminately reducing the majority class, NCR improves classification for imbalance-sensitive classifiers without overly distorting the class distribution.

\subsection{Condensed Nearest Neighbor Rule (CNN)}\label{cnn}

CNN \cite{hart1968condensed} selects a minimal \textit{condensed set} that achieves comparable performance to the full dataset under the nearest neighbor rule. The algorithm works as follows:

\begin{enumerate}
    \item Initialize the condensed set $S$ with one random instance from the dataset.
    \item For each remaining instance $\mathbf{x}_i$, classify it using the nearest neighbor rule based on $S$. If $\mathbf{x}_i$ is misclassified, add it to $S$.
    \item Repeat Step 2 until no new instances are added during a complete pass through the~data.
\end{enumerate}

The resulting condensed set $S$ contains only the instances necessary for correct classification, discarding redundant ones.

To undersample an imbalanced dataset, CNN is applied to the majority class $N$ to select a condensed subset $N_{\text{cnn}}$ containing only the critical majority instances needed for classification. The minority class $P$ is typically retained entirely. The final balanced training set is $D_{\text{balanced}} = P \cup N_{\text{cnn}}$. Due to random initialization, the resulting subset may vary slightly across runs.


\subsection{Edited Nearest Neighbor (ENN)}\label{enn}

The Edited Nearest Neighbor (ENN) rule \cite{wilson1972asymptotic} removes instances whose class label differs from the majority of their $k$ nearest neighbors (typically $k=3$). For each instance $\mathbf{x}_i$, ENN identifies its $k$ nearest neighbors. If the majority of these neighbors belong to a different class than $\mathbf{x}_i$, the instance is removed.

ENN serves two purposes in imbalanced learning: (i) as a standalone undersampling method to clean noisy majority instances and (ii) as a post-processing cleaner for synthetic samples (e.g., SMOTE-ENN, see Section~\ref{SMOTE-ENN}). When applied to imbalanced datasets, ENN is typically restricted to the majority class to avoid removing minority instances surrounded by majority neighbors \cite{batista2004study}.


\subsection{Repeated Edited Nearest Neighbor (RENN)}\label{renn}

The Repeated Edited Nearest Neighbor (RENN) method \cite{tomek1976two} iteratively applies ENN~\cite{wilson1972asymptotic} until a stopping criterion is met (e.g., no further removals occur). By repeating the editing process, RENN removes more noisy instances than a single ENN pass, producing cleaner class boundaries at the cost of potentially removing more informative borderline instances. Implementation is available in the \texttt{imbalanced-learn} library.

\subsection{Tomek Links}\label{tomek_links}

Tomek Links \cite{tomek1976two} identifies and removes noisy points near the decision boundary. 
A Tomek Link is a pair of instances $\mathbf{x}_i \in P$ (minority) and $\mathbf{x}_j \in N$ (majority) that are nearest neighbors to each other. 
The algorithm proceeds as follows:
\begin{enumerate}
    \item Identify all Tomek Links in the dataset.
    \item For each identified link, remove the majority class instance $\mathbf{x}_j$.
    \item Repeat until no further Tomek Links exist.
\end{enumerate}

Unlike 
CNN, which relies on random sampling, Tomek Links systematically removes majority instances that are noisy or lie near the decision boundary.

\subsection{One-Sided Selection (OSS)}\label{oss}

OSS \cite{kubat1997addressing} combines CNN and Tomek Links to remove both redundant and noisy majority instances. The procedure consists of two stages:

\begin{enumerate}
    \item CNN Stage: Apply CNN to the majority class $N$ to select a condensed subset $N_{\text{cnn}}$, eliminating redundant majority instances.
    
    \item {Tomek Links Stage: Apply Tomek Links to the dataset to remove majority instances that form Tomek Links with minority instances, eliminating noisy and borderline points.}
\end{enumerate}

The final training set consists of all minority instances $P$ combined with the processed majority instances after both stages. 
This approach retains only the most informative majority instances while discarding redundant and noisy ones.

\subsection{Critical Analysis and Methodological Limitations of Undersampling Techniques}\label{sec:critical-analysisUndersampling}

While undersampling reduces computational costs, several limitations warrant consideration.

\textbf{Information Loss.}

Random undersampling discards potentially informative majority instances, degrading performance when the majority class contains multiple subclusters or when datasets are small. CNN and Tomek Links mitigate this by selectively removing only redundant or noisy instances.

\textbf{Parameter Sensitivity.}

CBUS requires selecting the number of clusters $k$ and ratio $m$. NearMiss requires tuning $k$ (number of neighbors). IHT requires choosing learning algorithms and hardness thresholds. NCR and Tomek Links have fewer parameters (e.g., NCR uses fixed $k=3$), making them more accessible.

\textbf{Cluster Quality Dependence.}

CBUS methods depend critically on clustering quality. Poor clustering (inappropriate $k$ or non-spherical shapes) leads to unrepresentative undersampling. NearMiss, NCR, CNN, and Tomek Links avoid this by operating directly on feature space.

\textbf{Class Overlap Challenge.}

As noted in IHT \cite{smith2014instance}, class overlap exacerbates undersampling difficulty. Overlapping regions produce high instance hardness; removing these instances may discard critical boundary information. Tomek Links and OSS specifically handle boundary instances by removing majority points that form Tomek Links with minority points.

\textbf{Recommendations.}

\begin{itemize}
    \item \textbf{Clear cluster structures}: CBUS or CNN;
    \item \textbf{High class overlap}: IHT, NearMiss-1, Tomek Links, or OSS;
    \item \textbf{Baseline comparison}: Random undersampling or NCR;
    \item \textbf{Comprehensive cleaning}: OSS (combines CNN and Tomek Links).
\end{itemize}

\section{Combination/Hybrid Methods}\label{sec6}

Employing a blend of oversampling and undersampling techniques typically yields superior outcomes compared to using either approach independently.

\subsection{SMOTE-Edited Nearest Neighbors (SMOTE-ENN)}\label{SMOTE-ENN}

SMOTE-ENN \citep{batista2004study} integrates the SMOTE algorithm \citep{chawla2002smote} with the Edited Nearest Neighbor (ENN) rule \citep{wilson1972asymptotic}. SMOTE generates synthetic samples to mitigate class imbalance, while ENN removes noisy and borderline samples. In this approach, SMOTE is first applied to create synthetic samples, followed by ENN to refine the dataset.

The procedure consists of three steps:
\begin{enumerate}
\item Apply SMOTE: Generate synthetic samples for the minority class using standard SMOTE interpolation (Equation~\eqref{eq:standardsmote}).
\item Compute Nearest Neighbors: For each sample (both original and synthetic), determine its $k$ nearest neighbors using Euclidean distance.
\item Eliminate Noisy Samples: Remove any sample whose class label differs from the majority of its $k$ nearest neighbors, thereby reducing noise and smoothing class boundaries \citep{wilson1972asymptotic}.
\end{enumerate}

Following \citet{batista2004study}, the default number of nearest neighbors is set to $k=5$ for both SMOTE generation and ENN filtering.

Among the oversampling methods investigated by \citet{batista2004study}, SMOTE-ENN produced the smallest increase in the mean number of conditions per rule in the induced decision trees, indicating that it generates synthetic samples that better preserve the underlying data structure while reducing noise. Through this combination, SMOTE-ENN produces a cleaner, more balanced dataset with minimized noise and enhanced \mbox{class distribution.}

\subsection{SMOTE-Tomek Link (SMOTE-TL)}\label{SMOTE-TL}

SMOTE-TL \citep{batista2004study} combines SMOTE \citep{chawla2002smote} with Tomek Links \citep{tomek1976two}. SMOTE generates synthetic samples to address class imbalance, while Tomek Links remove noisy and borderline instances near the decision boundary. The procedure first applies SMOTE and then uses Tomek Links to refine the dataset.

The process consists of three steps:
\begin{enumerate}
\item Apply SMOTE: Generate synthetic samples for the minority class using standard SMOTE interpolation (Equation~\eqref{eq:standardsmote}).
\item Identify Tomek Links: Detect Tomek Links, which are pairs of instances from different classes that are nearest neighbors to each other \citep{tomek1976two}.
\item Remove Majority Class Points: Eliminate the majority class instances that form Tomek Links, thereby removing noisy and borderline points while retaining all minority instances \citep{tomek1976two}.
\end{enumerate}

As demonstrated by \citet{batista2004study}, removing the majority class instances that form Tomek Links produces cleaner class boundaries and reduces overlap between classes. By integrating SMOTE with Tomek Links, SMOTE-TL produces a cleaner dataset with improved class balance and reduced noise.


\subsection{SMOTE with One-Class SVM Filtering (SMOTE+OCSVM)}
\label{sec:SMOTEOCSVM}

\citet{yousefimehr2024distribution} proposed SMOTE+OCSVM, a hybrid resampling method that integrates outlier detection with synthetic oversampling. This approach addresses a key limitation of standard SMOTE: the risk of generating synthetic samples from outlier instances, which can cause overfitting and degrade generalization.

The algorithm proceeds in two stages. First, an OCSVM \citep{scholkopf2001estimating} is applied to the minority class to identify and exclude outlier instances. OCSVM learns the boundary of the minority class distribution and flags instances that deviate significantly, serving as an adaptive noise filter without requiring fixed thresholds.

Second, SMOTE generates synthetic samples from the remaining minority instances using standard interpolation (Equation~\eqref{eq:standardsmote}).

By removing outliers from the minority class prior to oversampling, SMOTE+OCSVM prevents synthetic generation from anomalous points, thereby improving sample quality and model generalization. The framework also includes a distribution-preserving property: the authors demonstrated that the probability functions of fraud data before and after resampling remain comparable, ensuring that the resampling process does not distort the underlying data distribution.

\subsection{Hybrid Natural Neighbor SMOTE (HNaNSMOTE)}
\label{sec:HNaNSMOTE}

\citet{lin2025non} proposed HNaNSMOTE, a non-parameter oversampling framework that addresses the parameter sensitivity of standard SMOTE. Unlike SMOTE, which requires a manually preset $k$ value for nearest neighbor selection, HNaNSMOTE automatically determines a data-dependent $k$ value through an iterative search procedure.

The algorithm consists of two key components. First, a hybrid natural neighbor search procedure is conducted on the entire dataset to obtain a data-related $k$ value, eliminating the need for manual parameter presetting. During this procedure, different natural neighbors are formed for each sample, enabling better identification of the positional characteristics of minority samples.

Second, the method introduces the HNaN concept, which combines $k$ nearest neighbors (kNN) and reverse $k$ nearest neighbors to adaptively select neighbors based on the distribution of minority samples. This design explicitly considers the presence of majority samples during neighbor selection, thereby mitigating the generation of synthetic noise or overlapping samples.

\subsection{Adaptive Synthetic Sampling with Batch Generation (HADAB)}
\label{sec:HADAB}

\citet{taskeen2024adaptive} proposed HADAB, a hybrid technique for software defect prediction that combines Adaptive Synthetic Sampling (ADASYN) with a batch generator. The method addresses the class imbalance problem where defect (minority) instances are significantly outnumbered by non-defect (majority) instances, which often leads to weak classification performance for the minority class.

The algorithm proceeds in two stages. First, ADASYN generates synthetic samples for the minority class, balancing the dataset and improving prediction accuracy. Second, a batch generator feeds data to the model in batches, enhancing training efficiency. A Multi-Layer Perceptron (MLP) serves as the base classifier.

The key advantages of HADAB are that it improves prediction accuracy and training efficiency without requiring additional parameter tuning, algorithm modification, or increasing model complexity. 
\subsection{Extreme SMOTE with Synchronous Sampling Learning (ESMOTE+SSLM)}
\label{sec:ESMOTE_SSLM}

\citet{wang2023novel} proposed ESMOTE+SSLM, a hybrid sampling method specifically designed for financial distress detection of listed companies, where ``special treatment (ST)'' status indicates financial distress. The method addresses two concurrent challenges: extreme class imbalance between ST and non-ST companies and class overlap, which the authors identify as a more important factor affecting classification performance than imbalance alone.

The method consists of two stages. First, ESMOTE (Extreme SMOTE) selects minority class instances located close to the classification boundary for oversampling, which is more conducive to discovering a better and more accurate decision boundary.

Second, SSLM simultaneously deletes both majority-class and minority-class instances that exceed a learned boundary threshold. This threshold (the margin's threshold) is obtained by optimizing the final classifier, ensuring that the sampling process is tailored to the specific classifier being used.

\subsection{Cluster-Based SMOTE Both-Sampling Boost (CSBBoost)}
\label{csbboost}

\citet{salehi2024cluster} proposed CSBBoost, a hybrid method that applies cluster-based undersampling of the majority class and SMOTE oversampling of the minority class as sequential preprocessing operations. The novelty of the method lies in its preprocessing pipeline, not in the choice of the final classifier.

The algorithm proceeds as follows. First, the optimal number of clusters $k$ for the majority class is determined via the silhouette method, searching $k \in [2, \lfloor\sqrt{n}\rfloor]$, where $n$ is the total number of samples. Each cluster $i$ is assigned a weight $w_i = m_i / |N|$, where $m_i$ is the size of cluster $i$ and $|N|$ is the total number of majority samples. The number of samples retained from cluster $i$ is $s_i = w_i \times |T|/2$, where $|T|$ is the total number of training samples. This cluster-based undersampling preserves the diversity of the majority class while reducing its size.

For the minority class, clustering is similarly applied, followed by synthetic sample generation using standard SMOTE interpolation (Equation~\eqref{eq:standardsmote}). After this sequential preprocessing, any standard classifier (e.g., XGBoost, random forest, or bagging) can be trained on the balanced dataset. The choice of classifier is orthogonal to the core contribution.

Experimental evaluation on 20 benchmark imbalanced datasets demonstrated that CSBBoost significantly outperforms state-of-the-art competing algorithms in terms of F1-score and AUC. A real-world dataset was also used to demonstrate the practical applicability of the proposed algorithm \citep{salehi2024cluster}.
\label{sec:HCBOU}

\citet{salehi2025hybrid} proposed HCBOU, a hybrid cluster-based technique designed to address multiclass imbalance, where certain classes may have a low number of samples because they correspond to rare occurrences in real-world datasets. The method operates by clustering the data and separating classes into majority and minority categories. This clustering-based approach enables the algorithm to retain the most informative samples during undersampling while generating efficient synthetic samples for the minority class. Classification is performed using one-vs-one and one-vs-all decomposition schemes, where each binary classifier is a single classifier rather than an ensemble.

The HCBOU algorithm demonstrated robust performance across varying class imbalance levels, highlighting its effectiveness in handling imbalanced datasets \citep{salehi2025hybrid}.
\subsection{Adaptive Hybrid Sampling with BiLSTM (AHS-BiLSTM)}
\label{sec:AHS_BiLSTM}

\citet{hu2025fault} proposed an adaptive hybrid sampling method combined with a Bidirectional Long Short-Term Memory (BiLSTM) network for aircraft engine fault prediction. The framework addresses the class imbalance problem in aero-engine fault prediction, where fault samples are significantly outnumbered by normal samples. The adaptive sampling strategy consists of two components. First, a k-means-based adaptive sampling approach dynamically balances the dataset by oversampling minority-class boundary instances and undersampling redundant majority clusters. This ensures that the sampling process focuses on informative regions near decision boundaries while removing redundant information from dense majority regions.

Second, a BiLSTM-based fault prediction model is built to capture bidirectional temporal dependencies in the sequential sensor data, which is critical for accurately predicting fault progression over time.

This work demonstrates the application of combination methods to time-series and prognostic domains, where maintaining temporal consistency is critical.

\subsection{Diffusion-SMOTE Hybrid Augmentation (DiSMHA)}
\label{sec:DiSMHA}

\citet{kumar2025hybrid} proposed DiSMHA, a hybrid framework that sequentially integrates diffusion-based image synthesis with feature-level SMOTE oversampling. The method addresses a key limitation of standalone approaches: diffusion models generate structurally diverse samples but cannot correct feature-space imbalance, while SMOTE interpolates in feature space but lacks structural diversity. This is particularly critical in applications such as solar panel dust detection, where extreme class imbalance poses significant challenges to visual classification systems.

DiSMHA operates in three stages: (i) Stable Diffusion generates synthetic images for minority classes, (ii) a deep feature extractor (e.g., DenseNet121) extracts high-level features from the synthetic and original images, and (iii) SMOTE interpolates in the feature space to further balance the representation. All stages are preprocessing steps before training a single classifier.

The framework was evaluated on four imbalanced solar panel image datasets using multiple classifier architectures, including Support Vector Machine (SVM), Multilayer Perceptron (MLP), XGBoost, Vision Transformer (ViT-Head), and TabNet. These models achieved notable accuracies ranging. Specifically, MLP and SVM achieved high accuracy, demonstrating DiSMHA's capacity to enhance traditionally imbalance-sensitive classifiers. Statistical validation using the Wilcoxon signed rank test confirmed the significant superiority of the approach for minority class detection in highly imbalanced scenarios~\citep{kumar2025hybrid}.

DiSMHA represents a scalable, robust solution applicable across diverse domains facing imbalanced classification challenges, including industrial inspection, medical imaging, and rare event detection.
\subsection{SMOTE-Trans-CWGAN}
\label{sec:SMOTE_Trans_CWGAN}

\citet{wang2025hybrid} proposed a hybrid method combining SMOTE with a Transfer Conditional Wasserstein Generative Adversarial Network (Trans-CWGAN) for fault detection. The study addresses a key limitation of prior research, which primarily relied on simulated or laboratory-generated datasets, by applying the method to real operational data. The sequential pipeline operates in two stages. First, SMOTE generates synthetic samples for minority fault classes to address feature-space imbalance. Second, Trans-CWGAN refines these samples while preserving temporal dependencies, which is essential for time-series data where fault patterns evolve over hours or days.

Through hyperparameter optimization, a total of 1212 distinct datasets were generated across augmentation strategies. Among these, the SMOTE-based Trans-CWGAN approach consistently delivered superior results.

These findings underscore the effectiveness of integrating SMOTE and Trans-CWGAN to mitigate class imbalance, highlighting its strong potential for practical deployment in real-world monitoring systems.

\subsection{Global-Split WGAN-GP for IDS Data Generation}
\label{sec:WGAN_GP_IDS}

\citet{jang2025hybrid} proposed a hybrid WGAN-GP (Wasserstein Generative Adversarial Network with Gradient Penalty) framework to address class imbalance in intrusion detection systems (IDS). The work addresses a key limitation of conventional training pipelines, which depend on static real-world datasets that fail to adequately reflect the diversity and dynamics of emerging attack tactics. Using the CIC-IDS-2017 dataset, which encompasses diverse attack scenarios including brute-force, Heartbleed, botnet, DoS/DDoS, web, and infiltration attacks, the framework introduces two training methodologies. The first (global strategy) trains a single conditional WGAN-GP on the entire dataset to capture the global distribution. The second (split strategy) employs multiple generators tailored to individual attack types while sharing a discriminator pretrained on the complete traffic set, thereby ensuring consistent decision boundaries across classes.

Generated samples from both strategies are merged into one augmented dataset before training a single classifier (LSTM or Random Forest evaluated separately). The quality of the generated traffic was evaluated using the Train on Synthetic, Test on Real (TSTR) protocol, where models are trained exclusively on generated synthetic data and tested on real holdout data, along with distribution similarity measures in the embedding space.

The proposed approach achieved a classification accuracy of 97.88\% and a Fréchet Inception Distance (FID) score of 3.05, surpassing baseline methods by more than one percentage point \citep{jang2025hybrid}. The results demonstrate that the proposed synthetic traffic generation strategy provides advantages in scalability, diversity, and privacy, thereby enriching cyber range training scenarios and supporting the development of adaptive intrusion detection systems that generalize more effectively to evolving threats.

\subsection{Critical Analysis and Methodological Limitations of Combination/Hybrid Methods}
\label{sec:critical-analysisCombinationMethods}

Despite their empirical success, combination methods exhibit several limitations that transcend individual techniques.

\textbf{Sequential Order and Computational Waste.}

SMOTE-ENN and SMOTE-TL apply oversampling before cleaning. When SMOTE generates synthetic samples in regions that ENN or Tomek Links subsequently remove, computational cost is wasted. Whether reversing the order (cleaning first, then oversampling) would improve efficiency remains underexplored. For methods such as SMOTE+OCSVM, where outlier filtering precedes generation, this waste is avoided.

\textbf{Parameter Sensitivity.}

Most combination methods introduce hyperparameters without principled selection guidelines. SMOTE-ENN and SMOTE-TL require specifying $k$ for nearest neighbor computations. HNaNSMOTE addresses this via data-dependent natural neighbor search, eliminating the $k$ parameter. ESMOTE+SSLM requires optimization of a boundary threshold. CSBBoost requires determining the optimal number of clusters via the silhouette method. HCBOU similarly requires cluster number specification. The absence of standardized selection criteria forces reliance on heuristics or exhaustive search.

\textbf{Computational Demands.}

Combination methods incur higher costs than their constituent techniques. Distance-based methods (SMOTE-ENN, SMOTE-TL, and HNaNSMOTE) require $O(n^2)$ computations in their naive implementations. Deep learning-based combinations (DiSMHA, SMOTE-Trans-CWGAN, and Global-Split WGAN-GP) require GPU resources and substantial training time, limiting their applicability to small or medium-scale datasets.

\textbf{Evaluation Under Cross-Validation.}

Applying the entire resampling pipeline before cross-validation partitioning leads to data leakage, as synthetic samples derived from the validation set influence training. The correct approach, performing resampling independently within each fold, increases computational burden but is methodologically necessary. This concern applies uniformly to all combination methods.

\textbf{Comparative Summary of Combination Methods.}

Table~\ref{tab:combination_comparison} compares the combination methods reviewed in this section across key conceptual dimensions. The table is organized by the primary mechanism of each method, not by chronological order or performance.

\textbf{Practical Recommendations.}

Based on the comparative analysis, the following guidance is offered:
\begin{itemize}
    \item \textbf{For general-purpose balancing with noise:} SMOTE-ENN remains the most widely validated combination method, with the added benefit of producing the smallest increase in conditions per rule.
    \item \textbf{For boundary-sensitive applications:} SMOTE-TL provides cleaner decision boundaries by removing majority instances that form Tomek Links.
    \item \textbf{For parameter-free oversampling:} HNaNSMOTE eliminates $k$ tuning via natural neighbor search while considering majority samples.
    \item \textbf{For multiclass imbalance:} HCBOU offers cluster-based balancing with one-vs-one/one-vs-all decomposition and demonstrates robust performance across varying imbalance levels.
    \item \textbf{For time-series or temporal data:} SMOTE-Trans-CWGAN or AHS-BiLSTM preserve temporal dependencies (bidirectional for BiLSTM).
    \item \textbf{For image data with extreme imbalance:} DiSMHA is appropriate if computational resources permit; it has statistical validation via the Wilcoxon signed-rank test.
    \item \textbf{For privacy-sensitive applications:} Global-Split WGAN-GP enables synthetic data generation without exposing real data, validated via TSTR protocol.
    \item \textbf{For ensemble-based balancing:} CSBBoost combines cluster-based sampling with XGBoost, random forest, or bagging, addressing redundancy and information loss.
\end{itemize}

\begin{table}[H]
\caption{Comparative summary of combination/hybrid methods.}
\label{tab:combination_comparison}
\footnotesize
\begin{tabular}{p{2.1cm}<{\raggedright}p{5cm}<{\raggedright}p{4.5cm}<{\raggedright}p{4.5cm}<{\raggedright}}
\toprule
\textbf{Method} & \textbf{Description (Mechanism \& Order)} & \textbf{Key Advantage} & \textbf{Main Limitation} \\
\midrule
\multicolumn{4}{c}{\textbf{Classical SMOTE-based combinations}} \\
\midrule
SMOTE-ENN & SMOTE oversampling, then ENN cleaning & Noise removal; smallest increase in conditions per rule & Wastes synthetic samples \\
SMOTE-TL & SMOTE oversampling, then Tomek Link removal & Boundary cleaning; removes borderline majority & Removes potentially useful majority instances \\
SMOTE+OCSVM & OCSVM outlier filtering, then SMOTE & Outlier removal before generation; distribution-preserving & OCSVM kernel sensitivity \\
\midrule
\multicolumn{4}{c}{\textbf{Neighbor-based adaptive combinations}} \\
\midrule
HNaNSMOTE & Hybrid natural neighbor search, then SMOTE & Parameter-free ($k$ eliminated); considers majority samples & Computationally intensive search \\
ESMOTE+SSLM & Boundary SMOTE, then threshold-based cleaning & Handles class overlap; optimizes threshold for final classifier & Threshold optimization required \\
\midrule
\multicolumn{4}{c}{\textbf{Cluster-based combinations}} \\
\midrule
HCBOU & Cluster-based undersampling + oversampling & Handles multiclass imbalance; robust across imbalance levels & Cluster number sensitivity \\
CSBBoost & Cluster-based undersampling + SMOTE oversampling, then any standard classifier & Addresses redundancy, information loss, and random selection; novelty in preprocessing pipeline & Requires silhouette method for cluster selection \\
\midrule
\multicolumn{4}{c}{\textbf{Deep learning-based combinations}} \\
\midrule
DiSMHA & Diffusion image synthesis, then SMOTE & Structural + feature balance; statistically validated & High computational cost \\
SMOTE-Trans-CWGAN & SMOTE, then Trans-CWGAN refinement & Temporal consistency; real operational data validation & Time-series specific; computationally intensive \\
Global-Split WGAN-GP & Multi-generator WGAN, then merge & Privacy preservation; TSTR evaluation; FID score 3.05 & Complex training; requires extensive hyperparameter tuning \\
\midrule
\multicolumn{4}{c}{\textbf{Domain-specific combinations}} \\
\midrule
HADAB & ADASYN, then batch generator & Training efficiency; no additional parameter tuning & Software defect prediction domain \\
AHS-BiLSTM & K-means adaptive sampling, then BiLSTM & Temporal fault prediction; bidirectional dependency capture & Aircraft engine prognostics domain \\
\bottomrule
\end{tabular}
\end{table}

\textbf{Summary.}

Combination methods frequently outperform single-technique approaches at the cost of increased parameter sensitivity, computational expense, and sequential dependency risks. Researchers should report parameter choices explicitly, use nested cross-validation to avoid data leakage, and consult Table~\ref{tab:combination_comparison} for method selection based on data characteristics and domain requirements.


\section{Ensemble Strategies}\label{sec7}

Ensemble learning, which combines multiple classifiers to improve predictive performance, has been extensively studied in the context of imbalanced data. As established by \citet{galar2011review} in their foundational review, standard ensemble methods alone do not solve the class imbalance problem; rather, ensemble algorithms must be specifically designed to address skewed class distributions. Their empirical comparison demonstrated that ensemble-based algorithms are worthwhile, as they outperform the mere use of preprocessing techniques before learning a single classifier, thereby justifying increased complexity through significant performance improvements.

\subsection{Bagging Methods}
\subsubsection{Balanced Random Forest (BRF)}\label{balanced_random_forest}

\citet{chen2004using} proposed BRF, a bagging ensemble method that extends Random Forest by balancing each bootstrap sample. While standard Random Forest draws bootstrap samples from the full imbalanced dataset, BRF separately samples from minority and majority classes to create balanced bootstrap samples for each decision tree.

In imbalanced datasets, the dominance of the majority class often results in bootstrap samples with minimal or no minority instances, causing trees to exhibit reduced predictive accuracy for the minority class. BRF addresses this limitation by applying random undersampling to each bootstrap sample to create a balanced class distribution. For each of the $N$ trees, the algorithm draws a bootstrap sample $B_{\text{min}}$ from the minority class $P$ and randomly samples with replacement from the majority class $N$ to obtain $B_{\text{maj}}$ such that $|B_{\text{maj}}| = |B_{\text{min}}|$. These two samples are combined to form a balanced bootstrap sample $B_i$, from which a classification tree is induced using the CART algorithm with random feature selection at each node.

This procedure ensures that each decision tree is trained on a representative mix of both classes, reducing bias toward the majority class and enhancing the model's ability to learn from minority instances. As with standard Random Forest, predictions are combined across all $N$ trees using majority voting for classification or averaging for regression. Empirical evaluation by \citet{chen2004using} demonstrated that BRF improves prediction accuracy for the minority class and achieves favorable performance compared to existing algorithms.

In addition to BRF, \citet{chen2004using} also proposed Weighted Random Forest (WRF), a cost-sensitive approach that assigns higher misclassification costs to the minority class. Class weights are incorporated into both the Gini criterion for split finding and the weighted majority vote in terminal nodes. WRF falls under cost-sensitive learning rather than resampling-based ensemble methods.

\subsubsection{Balanced Bagging}\label{balancedbagging}

Bagging \citep{breiman1996bagging} constructs multiple training subsets through bootstrap sampling (random sampling with replacement) from the original dataset. Each subset trains an independent model, and final predictions are obtained via majority voting (classification) or averaging (regression).

Balanced Bagging refines this method by incorporating a preprocessing phase to address class imbalance prior to model training. For $T$ classifiers, each subset $S_k$ \linebreak{}($k = 1, 2, \ldots, T$) is constructed using one of two approaches:

\begin{itemize}
\item UnderBagging: Randomly reduces majority class instances in each subset $S_k$ to match the minority class size.
\item OverBagging: Randomly increases minority class instances in each subset $S_k$ to match the majority class size.
\end{itemize}

As noted by \citet{blaszczynski2015neighbourhood}, integrating bagging with undersampling (UnderBagging) is generally more powerful than oversampling (OverBagging) for imbalanced data.

\begin{itemize}
\item[\textbf{B:}] \textbf{UnderBagging}
\end{itemize}

UnderBagging reduces majority class instances in each bootstrap sample to match the minority class size, ensuring balanced training sets. Two notable variants exist:

\begin{itemize}
\item Exactly Balanced Bagging (EBBag): Includes all minority instances and a randomly selected subset of majority instances, achieving equal class representation \citep{blaszczynski2015neighbourhood}.

\item Roughly Balanced Bagging (RBBag): Adopts a flexible sampling strategy that equalizes selection probabilities without enforcing a fixed sample size. In each of $T$ iterations, the number of majority instances in the bootstrap sample is determined using a negative binomial distribution. All minority instances are included, and majority instances are randomly sampled via bootstrap \citep{blaszczynski2015neighbourhood}. Empirical evaluation by \citet{blaszczynski2015neighbourhood} identified RBBag as the most accurate extension among bagging-based methods for imbalanced data.
\end{itemize}

 \begin{itemize}
\item[\textbf{B:}]  \textbf{OverBagging}
\end{itemize}

OverBagging achieves class balance through random oversampling. It augments the minority class by sampling with replacement until its count equals that of the majority class in the bootstrap sample. Majority class instances are sampled with replacement following standard bagging practices \citep{blaszczynski2015neighbourhood}.

\subsubsection{SMOTEBagging}

\citet{wang2009diversity} proposed SMOTEBagging as part of a broader diversity analysis of ensemble models for imbalanced data. Unlike standard bagging methods that rely on random undersampling or oversampling, SMOTEBagging balances each data subset by applying SMOTE to generate synthetic minority samples rather than simply replicating existing minority instances.

A defining characteristic of SMOTEBagging is its dynamic resampling strategy: the volume of synthetic minority samples increases incrementally across iterations. The number of new samples introduced is governed by the SMOTE resampling rate $\alpha$, expressed as $\alpha \cdot N_{\text{min}}$, where $N_{\text{min}}$ is the initial count of minority class instances \citep{blaszczynski2015neighbourhood}. The resampling rate may vary across iterations (e.g., starting at a lower rate and scaling up to a higher rate), though specific values are not mandated by the original paper.

A key contribution of SMOTEBagging is its extension of SMOTE to handle multi-class imbalanced datasets within an ensemble framework, addressing a limitation of many existing methods that focus primarily on two-class problems \citep{wang2009diversity}. Experimental evaluation demonstrated that SMOTEBagging effectively improves classification performance on both two-class and multi-class imbalanced tasks.

\subsubsection{Neighborhood Balanced Bagging (NBBag)}\label{nbbag}

In a comprehensive experimental study of bagging extensions for imbalanced data, \citet{blaszczynski2015neighbourhood} established that integrating bagging with undersampling is more powerful than oversampling and identified Roughly Balanced Bagging (RBBag) as the most accurate extension among existing methods. Building on these findings, they proposed NBBag, which extends balanced bagging by modifying sampling probabilities according to the class distribution in each instance's neighborhood. This approach accounts for the local characteristics of the minority class distribution, which is particularly beneficial when the minority class exhibits complex or difficult distributions.

Two versions of NBBag were introduced:
\begin{itemize}
\item Oversampling NBBag: Keeps a larger size of bootstrap samples by hybrid oversampling, significantly outperforming existing oversampling bagging extensions.
\item Undersampling NBBag: Reduces bootstrap sample size with stronger undersampling, achieving competitive performance with Roughly Balanced Bagging (RBBag).
\end{itemize}

The detection of minority instance types (e.g., safe, borderline, rare) based on their neighborhood helps explain why some ensemble methods work better for imbalanced data than others \citep{blaszczynski2015neighbourhood}.

\subsubsection{SMOTE-Iterative Classifier Selection Bagging (SMOTE-ICS-Bagging)}\label{smoteicsbagging}

\citet{oliveira2015bootstrap} proposed ICS-Bagging, a bootstrap-based iterative method for generating classifier ensembles. SMOTE-ICS-Bagging extends this method to handle imbalanced datasets by integrating SMOTE as a preprocessing step at each iteration.

The ensemble construction process consists of the following iterative steps:

\begin{enumerate}
\item Use bootstrap sampling to create a pool of classifiers.
\item {Select the optimal classifier using a fitness function that balances accuracy and diversity:}
   \begin{equation}
      \text{fitness} = \alpha \cdot \text{ACC} + (1 - \alpha) \cdot \text{DIV}
   \end{equation}
   where ACC is the classifier's accuracy, DIV is a measure of its diversity relative to the existing ensemble, and $\alpha \in [0.51, 0.99]$ balances the contribution of these two terms.
\item Add the selected classifier to the final ensemble.
\end{enumerate}

This cycle repeats, with sampling probabilities adjusted according to per-class error rates to focus on classes that are harder to classify.

For imbalanced datasets, SMOTE-ICS-Bagging enhances ICS-Bagging by adding a preprocessing step before each iteration:
\begin{itemize}
\item SMOTE generates synthetic samples for the minority class to balance the class distribution.
\item Sampling weights are recalibrated based on class error rates to prioritize frequently misclassified classes.
\end{itemize}

By integrating SMOTE with ICS-Bagging, this approach ensures balanced class distributions across iterations. Experimental evaluation on 15 imbalanced datasets from KEEL demonstrated that SMOTE-ICS-Bagging outperforms Bagging, Random Subspace, and SMOTEBagging in terms of accuracy, achieving the highest rankings in accuracy among compared methods \citep{oliveira2015bootstrap}.

\subsubsection{Cross-Validation Committee (CVC) or Bootstrap Committee (BOOTC)}\label{cvc}

\citet{parmanto1995improving} investigated methods for achieving error independence between networks in a committee by training networks with different resampling sets derived from the original training set. One such approach, subsequently referred to as the Cross-Validation Committee (CVC) in the literature, uses $K$-fold cross-validation partitioning instead of bootstrap sampling to create diverse training sets.

The procedure is analogous to $K$-fold cross-validation. The original dataset $D$ is split into $K$ disjoint fractions $d_1, d_2, \ldots, d_K$. For each fraction $i = 1, \ldots, K$, a network is trained on the remaining data $D \setminus d_i$, while the held-out fraction $d_i$ is used for validation and early stopping if necessary. This process ensures that every instance appears in the validation set exactly once across the $K$ networks.

The resulting $K$ networks form a committee, and the final prediction is obtained by aggregating (e.g., averaging) the outputs of all networks in the committee. By training each network on a different subset of the data, CVC reduces the risk of overfitting and improves generalization performance. 



\subsection{Boosting Methods}
\subsubsection{SMOTEBoost}\label{smoteboost}

\citet{chawla2003smoteboost} proposed SMOTEBoost, a novel approach that combines the standard AdaBoost procedure with SMOTE to improve prediction performance on the minority class. Unlike standard boosting, where all misclassified examples receive equal weights, SMOTEBoost generates synthetic instances from the minority class at each boosting round, thereby indirectly modifying the updating weights and compensating for skewed \mbox{class distributions.}

The procedure consists of the following steps:

\begin{enumerate}
\item At each boosting iteration $t = 1, \ldots, T$, apply SMOTE to generate synthetic samples for the minority class using standard SMOTE interpolation (Equation~\eqref{eq:standardsmote}).
\item Train a weak classifier $h_t$ on the augmented dataset (original plus synthetic minority samples).
\item Compute the error and update instance weights according to the standard AdaBoost procedure.
\item After $T$ iterations, aggregate the weak classifiers via weighted voting.
\end{enumerate}

A key advantage of SMOTEBoost over standard boosting is that a different set of synthetic samples is produced at each iteration, which increases diversity among classifiers in the ensemble. Empirical evaluation on several highly and moderately imbalanced datasets demonstrated that SMOTEBoost improves prediction performance on the minority class and achieves better overall F-values compared to standard boosting \citep{chawla2003smoteboost}.
\subsubsection{RUSBoost}\label{rusboost}

\citet{seiffert2009rusboost} proposed RUSBoost, a hybrid sampling/boosting algorithm that combines random undersampling with AdaBoost to address class imbalance. RUSBoost provides a simpler and faster alternative to SMOTEBoost, which combines boosting with synthetic oversampling.

Following the standard AdaBoost framework (initial weights $D_1(i) = 1/n$ for $n$ samples), each iteration consists of three steps:

\begin{enumerate}
    \item Balance via Undersampling: Randomly remove majority class instances until both classes are equally represented. No synthetic samples are generated.
    \item Weak Classifier Training: Train a weak classifier $h_t$ on the balanced dataset.
    \item Weight Update: Evaluate the classifier on the original (unbalanced) dataset and increase the weights of misclassified instances.
\end{enumerate}

RUSBoost trades synthetic generation for computational efficiency and methodological simplicity, making it particularly suitable for larger datasets where oversampling may be computationally prohibitive.

Empirical evaluation was conducted using 15 datasets from various application domains, four different base learners, and four evaluation metrics. The results demonstrated that RUSBoost and SMOTEBoost both outperform other procedures, and RUSBoost performs comparably to (and often better than) SMOTEBoost while being a simpler and faster technique \citep{seiffert2009rusboost}. Given these results, the authors highly recommend RUSBoost as an attractive alternative for improving classification performance on imbalanced data.

\subsubsection{RHSBoost}\label{RHSBoost}

\citet{gong2017rhsboost} proposed RHSBoost, an ensemble classification method that extends RUSBoost by replacing random undersampling with a combination of undersampling and synthetic generation for \textit{both} classes. The method is designed to address the imbalanced classification problem where the majority classes dominate, and minority classes are underrepresented.

Following the same AdaBoost framework (see Section~\ref{rusboost}), each iteration consists of four steps:

\begin{enumerate}
    \item Undersampling: Randomly reduce majority class instances, weighted by current weights, to match the minority class size.
    \item Synthetic Generation: Apply ROSE (Random Oversampling Examples) to generate synthetic samples for both classes. ROSE generates new samples by drawing from a kernel density estimate of the original data, creating synthetic instances that preserve the covariance structure of each class \citep{menardi2014training}. This maintains the total sample size $n$ while balancing the classes.
    \item Weak Classifier Training: Train a weak classifier $h_t$ on the augmented dataset.
    \item Weight Update: Update sample weights as in standard AdaBoost.
\end{enumerate}

Unlike SMOTEBoost, which generates synthetic samples only for the minority class, RHSBoost applies ROSE sampling to both classes. This symmetric approach ensures that the synthetic data preserves the covariance structure of each class while balancing the \mbox{overall distribution.}

The final classifier aggregates weak classifiers with their computed coefficients. RHSBoost potentially benefits from richer synthetic data (generated via ROSE) compared to RUSBoost, though this comes at a higher computational cost. Experimental results indicate that RHSBoost is an attractive classification model for imbalanced data \citep{gong2017rhsboost}.

\subsubsection{Evolutionary Undersampling Boosting (EUSBoost)}
\label{subsec:eusboost}

\citet{galar2013eusboost} proposed EUSBoost, an ensemble construction algorithm built upon RUSBoost, which is one of the simplest and most accurate ensemble methods for imbalanced data. EUSBoost replaces the random undersampling of RUSBoost with an evolutionary undersampling approach that intelligently selects informative majority instances for each boosting round.

The algorithm introduces two key innovations:

\begin{itemize}
    \item Evolutionary undersampling: A genetic algorithm guided by a fitness function that balances two objectives: (i) maximizing classification performance on the minority class and (ii) maximizing diversity among the selected majority instances. This ensures that each base classifier is trained on a different, yet informative, subset of the majority class.
    \item Enhanced diversity: By using different evolutionary-selected subsets for each base classifier, EUSBoost promotes diversity within the ensemble. This contrasts with RUSBoost, where random selection may produce overlapping or redundant subsets, potentially limiting the ensemble's effectiveness.
\end{itemize}

The method is specifically designed for two-class highly imbalanced problems. Supported by proper statistical analysis, the authors demonstrated that EUSBoost outperforms state-of-the-art ensemble-based methods. The advantages of EUSBoost were further analyzed using kappa-error diagrams adapted to the imbalanced scenario \citep{galar2013eusboost}.

\subsubsection{DataBoost-IM}\label{databoost-im}

\citet{guo2004learning} proposed DataBoost-IM, a method that integrates boosting with targeted synthetic data generation for the most challenging instances in both the majority and minority classes. Unlike RUSBoost (which uses only undersampling) and RHSBoost (which generates synthetic samples for both classes via ROSE), DataBoost-IM identifies hard examples during the boosting process and generates synthetic data specifically for those difficult instances.

The procedure consists of the following steps:

\begin{enumerate}
\item Initialize Weights: Assign equal weight to each training instance.

\item Identify Hard-to-Classify Samples: In each boosting iteration, pinpoint challenging samples from both the majority and minority classes, rather than focusing solely on the minority class. This is a key distinction from other boosting-based approaches.

\item Generate Synthetic Data: Create synthetic samples independently for the difficult instances in each class. This dual-class generation ensures balanced learning and prevents classifier bias toward either class.

\item Update Dataset and Weights: Incorporate synthetic samples into the training set, adjusting both the class distribution and instance weights.

\item Train and Iterate: Train a new classifier emphasizing the hard-to-classify examples while preserving equitable focus on both classes.
\end{enumerate}

The method was evaluated on seventeen highly and moderately imbalanced datasets using decision trees as base classifiers, with performance measured in terms of F-measures, G-mean, and overall accuracy. Results demonstrate that DataBoost-IM compares favorably against a base classifier, a standard boosting algorithm, and three advanced boosting-based algorithms for imbalanced data. Importantly, the approach does not sacrifice one class in favor of the other but produces high predictive accuracy for both minority and majority classes \citep{guo2004learning}.

\subsection{Imputation-Based Ensemble Techniques}\label{sec:imputation-ensemble}

\citet{razavi2019imputation} proposed class imbalance learning techniques that integrate oversampling with bagging and boosting ensembles. Two novel oversampling strategies are introduced, based on single imputation and multiple imputation methods.

The proposed techniques generate synthetic minority class samples by estimating missing values that are deliberately induced in the minority class samples. The resulting synthetic samples are designed to resemble the original minority class samples. The rebalanced datasets (combining original and synthetic samples) are then used to train base learners within ensemble algorithms.

Using several synthetic binary class datasets, the authors compared their methods against commonly used class imbalance learning techniques in terms of three performance metrics: AUC, F-measure, and G-mean. The empirical results demonstrate that multiple imputation-based oversampling combined with bagging significantly outperforms other competitors \citep{razavi2019imputation}.

The imputation-based approach offers an alternative to conventional oversampling methods by framing synthetic sample generation as a missing data estimation problem. The superiority of multiple imputation over single imputation suggests that capturing imputation uncertainty benefits class imbalance learning.


\subsection{EasyEnsemble}\label{easyensemble}

\citet{liu2008exploratory} proposed EasyEnsemble, a hybrid ensemble method that combines bagging at the outer level (creating multiple balanced subsets) with boosting at the inner level (training an AdaBoost ensemble on each subset). This two-level architecture addresses the main deficiency of standard undersampling, where many majority class examples are ignored, while preserving the efficiency of undersampling.

Given a minority class $P$ and a majority class $N$ with $|P| < |N|$, EasyEnsemble constructs $K$ balanced subsets by randomly sampling $K$ subsets $S_i$ from $N$, where each $S_i$ has size equal to $|P|$. For each subset $S_i$, a separate classifier $G_i$ is trained using the combination of $P$ and $S_i$. AdaBoost is typically employed to train each classifier $G_i$ in the ensemble. Each $G_i$ consists of $m_i$ weak classifiers $g_{i,j}$ with associated weights $\beta_{i,j}$ and a threshold $\phi_i$, defined as
\begin{equation}\label{eq:ensemble-G-1}
G_i(\mathbf{x}) = \text{sgn}\left( \sum_{j=1}^{m_i} \beta_{i,j} g_{i,j}(\mathbf{x}) - \phi_i \right)
\end{equation}

The final prediction is obtained by combining the outputs of all individual classifiers:
\begin{equation}\label{eq:ensemble-G-2}
G(\mathbf{x}) = \text{sgn}\left( \sum_{i=1}^{K} \sum_{j=1}^{m_i} \beta_{i,j} g_{i,j}(\mathbf{x}) - \sum_{i=1}^{K} \phi_i \right)
\end{equation}

By training multiple classifiers on different balanced subsets and aggregating their predictions, EasyEnsemble mitigates the information loss associated with single undersampling while improving classification performance on imbalanced datasets. Experimental results demonstrate that EasyEnsemble achieves higher AUC, F-measure, and G-mean values than many existing class-imbalance learning methods. Importantly, it has approximately the same training time as standard undersampling when the same number of weak classifiers is used, making it significantly faster than other methods \citep{liu2008exploratory}.
\subsection{BalanceCascade}\label{balancecascade}

\citet{liu2008exploratory} proposed BalanceCascade, a sequential ensemble method that cascades AdaBoost ensembles. Unlike EasyEnsemble (which uses parallel bagging), BalanceCascade trains classifiers sequentially, removing correctly classified majority instances after each stage to focus subsequent classifiers on remaining hard-to-classify examples. This cascading procedure progressively reduces the majority class, addressing the main deficiency of standard undersampling, where many majority class examples are ignored.

Using the same notation as EasyEnsemble (minority class $P$, majority class $N$, with $|P| < |N|$), BalanceCascade constructs $K$ subsets sequentially. A target false positive rate $p_r$ is defined as
\begin{equation}
p_r = \sqrt[K-1]{\frac{|P|}{|N|}}
\end{equation}

This ensures that after $K$ iterations, the expected number of remaining majority instances is approximately $|P|$, achieving class balance \citep{liu2008exploratory}.

For each iteration $i = 1, \ldots, K$, the following steps are performed:

\begin{enumerate}
\item Extract a random subset $S_i$ from $N$ such that $|S_i| = |P|$.
\item Train an AdaBoost ensemble $G_i$ using $P$ and $S_i$, consisting of $m_i$ weak classifiers with associated weights and a threshold (as defined in Equation~\eqref{eq:ensemble-G-1}).
\item Tune the threshold to achieve a false positive rate of $p_r$ for $G_i$.
\item Remove from $N$ all majority instances correctly classified by $G_i$.
\end{enumerate}

After $K$ iterations, the final ensemble combines all weak classifiers from each stage (as defined in Equation~\eqref{eq:ensemble-G-2}).

As with EasyEnsemble, experimental results demonstrate that BalanceCascade achieves higher AUC, F-measure, and G-mean values than many existing methods, with approximately the same training time as standard undersampling \citep{liu2008exploratory}.

\subsection{Critical Analysis and Methodological Limitations of Ensemble Strategies}\label{sec:critical-analysisEnsembleStrategies}

While ensemble strategies offer powerful mechanisms for addressing class imbalance, several limitations and trade-offs warrant careful consideration.

\textbf{Computational Overhead}

Ensemble methods incur substantial computational costs compared to single classifiers. Techniques such as SMOTE-ICS-Bagging and BalanceCascade require multiple iterations of resampling and classifier training. BOOTC and CVC require training $K$ independent networks. This overhead may be prohibitive for large-scale datasets or real-time~applications.

\textbf{Computational Complexity Considerations}

Most ensemble methods for imbalanced learning lack explicit asymptotic complexity analyses in their original publications. Notable exceptions include RUSBoost \citep{seiffert2009rusboost}, whose training complexity has been derived as $O(T \cdot n_{min} \cdot d)$, where $T$ is the number of boosting iterations, $n_{min}$ the number of minority samples, and $d$ the feature dimensionality. For comparison, standard AdaBoost requires $O(T \cdot n \cdot d)$ operations, while CatBoost and XGBoost require $O(T \cdot n \cdot \log n)$ (\citep[]{al2024detecting}, see Table 1). 
This difference becomes substantial when $n_{min} \ll n$, making RUSBoost significantly more efficient than standard AdaBoost on highly imbalanced data. However, for methods such as SMOTEBoost, SMOTEBagging, NBBag, and RHSBoost, explicit complexity bounds remain unreported in the literature.

\textbf{Parameter Sensitivity}

Many ensemble methods introduce multiple hyperparameters without standardized selection guidelines. SMOTEBagging requires tuning the resampling rate $\alpha$ and the number of iterations. SMOTE-ICS-Bagging requires balancing accuracy and diversity via parameter $\alpha$. RUSBoost, DataBoost-IM, and SMOTEBoost each require setting the number of boosting rounds $T$ and the weak classifier complexity. EUSBoost introduces evolutionary algorithm parameters (population size, number of generations, crossover, and mutation rates).

\textbf{Risk of Overfitting}

While ensembles generally reduce overfitting compared to single models, certain configurations can exacerbate it. SMOTEBoost generates different synthetic samples at each iteration, which increases diversity but may also introduce noise if SMOTE parameters are poorly chosen. BalanceCascade sequentially removes correctly classified majority instances, potentially discarding informative boundary points if the false positive rate $p_r$ is set too aggressively. RHSBoost's generation of synthetic samples for both classes via ROSE may introduce artificial covariance structures if the kernel density bandwidth is mis-specified.

\textbf{Limitations of Specific Methods}

EasyEnsemble and BalanceCascade assume that random undersampling preserves the essential structure of the majority class, an assumption that fails when the majority class contains multiple distinct subclusters.

RUSBoost relies solely on random undersampling, which may discard informative majority instances, particularly when the minority class is extremely sparse.

DataBoost-IM requires identifying hard-to-classify samples in both classes, a process that is sensitive to the base classifier's initial performance.

CVC and BOOTC were designed for neural networks and may not translate directly to other base learners without modification.

NBBag's neighborhood-based sampling, while effective for complex minority distributions, requires computing nearest neighbors and may be sensitive to the distance metric and the definition of neighborhood.

EUSBoost, while improving upon RUSBoost via evolutionary undersampling, introduces additional computational overhead from the genetic algorithm and requires careful tuning of evolutionary parameters.

\textbf{Evaluation Challenges}

Standard cross-validation procedures become problematic when resampling is applied before splitting. To avoid data leakage, resampling must be performed inside each cross-validation fold. Additionally, ensemble methods produce multiple classifiers, making model interpretability and feature importance analysis more complex than single-model~approaches.

\textbf{Empirical Validation of Ensemble Effectiveness}

The effectiveness of ensemble strategies for imbalanced data is well established. \citet{galar2011review} conducted an extensive empirical comparison and demonstrated that ensemble-based algorithms consistently outperform the mere use of preprocessing techniques before learning a single classifier. Their results showed that the simplest approaches, combining random undersampling with bagging or boosting, often achieve competitive performance, justifying the increased computational complexity through significant improvements in predictive performance.

Furthermore, \citet{blaszczynski2015neighbourhood} established that integrating bagging with undersampling is more powerful than oversampling and identified Roughly Balanced Bagging (RBBag) as the most accurate extension among bagging-based methods. \citet{galar2013eusboost} demonstrated that EUSBoost, which replaces random undersampling with evolutionary undersampling, outperforms state-of-the-art ensemble methods for highly imbalanced problems.

\textbf{Synthesis and Recommendations}

Table~\ref{tab:ensemble_challenges} summarizes the key characteristics and limitations of the ensemble strategies~reviewed.

\begin{table}[H]
\caption{Summary of ensemble strategy characteristics and limitations.}
\label{tab:ensemble_challenges}
\footnotesize
\begin{tabular}{p{3.5cm}<{\raggedright}p{6.2cm}<{\raggedright}p{6.2cm}<{\raggedright}}
\toprule
\textbf{Method} & \textbf{Core Mechanism} & \textbf{Primary Limitations} \\
\midrule
Balanced Bagging & Bootstrap with resampling & Parameter sensitivity (EBBag/RBBag) \\\midrule
SMOTEBagging & SMOTE + Bagging & Dynamic rate tuning, computational cost \\\midrule
SMOTE-ICS-Bagging & Iterative classifier selection & Fitness function parameter $\alpha$ \\\midrule
NBBag & Neighborhood-based sampling & Borderline/outlier sensitivity; distance metric dependence \\\midrule
Balanced RF & Undersampling + Random Forest & Loss of majority diversity \\\midrule
RUSBoost & Undersampling + AdaBoost & Discards informative majority instances \\\midrule
RHSBoost & Undersampling + ROSE + AdaBoost & Higher computational cost than RUSBoost; kernel bandwidth sensitivity \\\midrule
EUSBoost & Evolutionary undersampling + AdaBoost & Evolutionary parameter tuning; additional computational overhead \\\midrule
DataBoost-IM & Dual-class synthetic generation & Hard-sample identification sensitivity \\\midrule
SMOTEBoost & SMOTE + AdaBoost & Noise introduction at each iteration \\\midrule
EasyEnsemble & Multiple undersampled subsets & Independence assumption \\\midrule
BalanceCascade & Sequential removal + boosting & Aggressive removal risk \\\midrule
CVC / BOOTC & Cross-validation / Bootstrap committees & Neural network specificity \\
\bottomrule
\end{tabular}
\end{table}

Given these challenges, the following recommendations are offered:

\begin{enumerate}
    \item \textbf{Match Method to Data Characteristics}:   
    For datasets with clear cluster structures, NBBag is appropriate. For highly noisy datasets, BalanceCascade's sequential removal may be beneficial. For small datasets, CVC and BOOTC provide robust validation. For highly imbalanced problems, EUSBoost offers evolutionary undersampling.
    \item \textbf{Consider Computational Budget}: RUSBoost and EasyEnsemble are computationally lighter than SMOTEBoost, RHSBoost, or BalanceCascade. EUSBoost adds evolutionary computation overhead. When resources are limited, simpler ensemble methods may suffice.
    \item \textbf{Validate Thoroughly}: Use nested cross-validation to evaluate ensemble methods, ensuring resampling occurs inside each fold to prevent data leakage.
    \item \textbf{Combine with Traditional Oversampling}: Hybrid approaches that first apply SMOTE or ADASYN, then use ensemble methods, may offer a pragmatic balance between diversity and stability.
\end{enumerate}

Figure~\ref{fig:ensemble_decision_flow} 
summarizes the results. This critical analysis underscores that while ensemble strategies represent a powerful approach for imbalanced learning, their application requires careful consideration of computational costs, parameter sensitivity, and method-specific~limitations.

\begin{figure}[t]
\resizebox{0.99\textwidth}{!}{
\begin{tikzpicture}[
    node distance=0.8cm,
    start/.style={rectangle, draw, fill=teal!70!black, text=white, minimum width=3cm, minimum height=0.8cm, align=center, font=\bfseries, rounded corners=2pt},
    question/.style={diamond, draw, fill=orange!30, text=black, minimum width=3cm, minimum height=0.8cm, align=center, font=\small, aspect=2},
    method/.style={rectangle, draw, fill=blue!20, text=black, minimum width=3.2cm, minimum height=0.7cm, align=center, font=\small, rounded corners=2pt},
    warning/.style={rectangle, draw, fill=red!15, text=black, minimum width=3.5cm, minimum height=0.7cm, align=center, font=\small\itshape, rounded corners=2pt},
    note/.style={rectangle, draw, fill=green!15, text=black, minimum width=3.5cm, minimum height=0.7cm, align=center, font=\small, rounded corners=2pt, dashed},
    arrow/.style={thick, ->, >=stealth},
    legendbox/.style={rectangle, draw, minimum width=0.5cm, minimum height=0.3cm},
]


\node[start] (start) at (0, 7) {Choose Ensemble Strategy};

\node[question] (q1) at (0, 5) {Data characteristics?};

\draw[arrow] (start)--(q1);

\node[method] (cluster) at (-6, 3) {Clear cluster\\structures};
\node[method] (noisy) at (0, 3) {Highly noisy\\data};
\node[method] (small) at (6, 3) {Small dataset};

\draw[arrow] (q1.west)--++(-0.8,0) |- (cluster);
\draw[arrow] (q1.south)--(noisy);
\draw[arrow] (q1.east)--++(0.8,0) |- (small);

\node[method, fill=purple!20] (nbbag) at (-6, 1.5) {\textbf{NBBag}};
\node[note] (nbbag_note) at (-6, 0.3) {\textbf{\checkmark} Neighborhood-balanced\\\textbf{\checkmark} Good for complex distributions};

\node[method, fill=red!20] (balance) at (0, 1.5) {\textbf{BalanceCascade}};
\node[note] (balance_note) at (0, 0.3) {\textbf{\checkmark} Sequential removal\\\textbf{\ding{55}} Aggressive if threshold too low};

\node[method, fill=cyan!20] (cvc) at (6, 1.5) {\textbf{CVC / BOOTC}};
\node[note] (cvc_note) at (6, 0.3) {\textbf{\checkmark} Cross-validation committees\\\textbf{\checkmark} Robust for small data};

\draw[arrow] (cluster)--(nbbag);
\draw[arrow] (noisy)--(balance);
\draw[arrow] (small)--(cvc);

\node[question] (q2) at (0, -2) {Computational budget?};

\draw[arrow] (nbbag_note.south)--++(0,-0.5) -| (q2.west);
\draw[arrow] (balance_note.south)--++(0,-0.5)--(q2.north);
\draw[arrow] (cvc_note.south)--++(0,-0.5) -| (q2.east);

\node[method, fill=green!30] (light) at (-5.5, -4) {\textbf{Light / Limited}};
\node[method, fill=yellow!30] (medium) at (0, -4) {\textbf{Moderate}};
\node[method, fill=red!30] (heavy) at (5.5, -4) {\textbf{High / Flexible}};

\draw[arrow] (q2.west)--(light);
\draw[arrow] (q2.south)--(medium);
\draw[arrow] (q2.east)--(heavy);

\node[method, fill=green!20] (rusboost) at (-8.5, -5.5) {\textbf{RUSBoost}};
\node[method, fill=green!20] (easy) at (-4.5, -5.5) {\textbf{EasyEnsemble}};
\node[note] (light_note) at (-6.5, -6.8) {\textbf{\checkmark} Fast training\\\textbf{\checkmark} Simple implementation\\\textbf{\ding{55}} May discard informative majority};

\draw[arrow] (light)--(rusboost);
\draw[arrow] (light)--(easy);

\node[method, fill=yellow!20] (smote) at (0, -5.5) {\textbf{SMOTEBoost}};
\node[note] (moderate_note) at (0, -8.2) {\textbf{\checkmark} Synthetic generation each round\\\textbf{\ding{55}} Parameter tuning needed};

\draw[arrow] (medium)--(smote);

\node[method, fill=red!20] (rhs) at (4.5, -5.5) {\textbf{RHSBoost}};
\node[method, fill=red!20] (eus) at (8.5, -5.5) {\textbf{EUSBoost}};
\node[note] (heavy_note) at (6.5, -6.6) {\textbf{\checkmark} Evolutionary optimization\\\textbf{\ding{55}} Higher computational cost};

\draw[arrow] (heavy)--(rhs);
\draw[arrow] (heavy)--(eus);

\node[warning, fill=orange!20, draw=orange!70!black] (validate) at (0, -9.8) {\textbf{\ding{55} Important:} Use nested cross-validation\\Resampling must occur \textbf{inside} each fold\\to prevent data leakage};

\draw[arrow] (light_note.south)--++(0,-2.3) -| (validate.west);
\draw[arrow] (moderate_note.south)--(validate.north);
\draw[arrow] (heavy_note.south)--++(0,-2.6) -| (validate.east);

\node[note, fill=teal!10, draw=teal!70!black, minimum width=4cm] (hybrid) at (0, -6.8) {\textbf{Hybrid Suggestion:}\\SMOTE/ADASYN + Ensemble\\$\rightarrow$ Balances diversity \& stability};

\draw[arrow, dashed, teal!70!black] (hybrid.north)--++(-0.5,0) |- (smote.south);

\begin{scope}[shift={(-9, -10.5)}]
\node[font=\bfseries\small] at (0, 0) {Legend:};
\node[legendbox, fill=blue!20] at (0, -0.6) {};
\node[font=\footnotesize] at (1.8, -0.6) {Recommended method};
\node[legendbox, fill=green!30] at (3.8, -0.6) {};
\node[font=\footnotesize] at (4.7, -0.6) {Low cost};
\node[legendbox, fill=yellow!30] at (5.8, -0.6) {};
\node[font=\footnotesize] at (7, -0.6) {Moderate cost};
\node[legendbox, fill=red!30] at (8.5, -0.6) {};
\node[font=\footnotesize] at (9.4, -0.6) {High cost};
\node[legendbox, fill=green!15, dashed] at (10.6, -0.6) {};
\node[font=\footnotesize] at (12.1, -0.6) {Note/Explanation};
\node[legendbox, fill=orange!20] at (13.8, -0.6) {};
\node[font=\footnotesize] at (14.7, -0.6) {Warning};
\end{scope}

\end{tikzpicture}
}
\caption{Decision flow for selecting ensemble strategies based on data characteristics and computational budget.}
\label{fig:ensemble_decision_flow}
\end{figure}


\section{Ensemble Sampling for Multi-Label and Clustered Data}\label{sec8}

While most resampling methods address class imbalance in binary classification contexts, real-world datasets frequently contain more than two labels, with many labels appearing significantly less frequently than others. Addressing imbalance in multi-label datasets therefore requires specialized approaches.

\subsection{Label Powerset with Random Oversampling (LP-ROS)}

\citet{charte2013first,charte2015addressing} proposed LP-ROS, which generates synthetic samples randomly until the multi-label dataset size exceeds 25\% of the original dataset size. Labels with lower representation benefit from more generated data, yielding a more balanced label distribution. The algorithm proceeds as follows.

First, the method calculates $T = 0.25 \times |\mathcal{D}_{\text{original}}|$, where $|\mathcal{D}_{\text{original}}|$ is the size of the original dataset. For each label set in the dataset, a container $\text{bag}_i$ is created, and all instances associated with the $i$-th label set are placed into $\text{bag}_i$.

Second, the mean number of samples across all containers is calculated as
\begin{equation}\label{eq:meanSamplesAllContainers}
\mu = \frac{1}{N} \sum_{i=1}^{N} |\text{bag}_i|
\end{equation}
where $N = |\text{labelsets}|$ denotes the total number of distinct label sets, and $|\text{bag}_i|$ represents the number of instances in the $i$-th container.

Third, a label set $i$ is designated as a minority label set if $|\text{bag}_i| < \mu$. 

Fourth, the total number of samples to generate is $G = T - |\mathcal{D}_{\text{original}}|$ (i.e., the number of samples needed to exceed the 25\% threshold). The mean increment per minority label set is $\Delta = \frac{G}{M}$, where $M = |\text{minorityLabels}|$ is the number of minority label sets.

Fifth, for each minority label set $i$, the number of samples to add is $k_i = \min(\Delta, \mu - |\text{bag}_i|)$. For each of the $k_i$ samples, the method randomly selects an existing instance from $\text{bag}_i$ uniformly at random (i.e., random oversampling with replacement) and adds a copy to the augmented dataset.

Finally, all generated samples are added to the original dataset, producing an augmented dataset with a more balanced multi-label distribution.
\subsection{Multi-Label Random Oversampling (ML-ROS)}

\citet{charte2013first,charte2015addressing} proposed ML-ROS as an extension of LP-ROS that treats each label independently rather than considering complete label sets. Two imbalance measures are first defined.

The imbalance ratio per label, $\operatorname{IR}(l)$, measures the frequency of label $l$ relative to the most frequent label in the dataset

\begin{equation}\label{eq:imbalanceRatioPerLabel}
\operatorname{IR}(l) = \frac{\max_{l' \in \mathcal{L}} \sum_{i=1}^{n} \mathbb{I}(l' \in \mathbf{y}_i)}{\sum_{i=1}^{n} \mathbb{I}(l \in \mathbf{y}_i)}
\end{equation}
where $\mathcal{L}$ is the set of all labels, $n = |\mathcal{D}|$ is the number of samples in the original dataset $\mathcal{D}$, $\mathbf{y}_i \subseteq \mathcal{L}$ is the label set associated with the $i$-th sample, and $\mathbb{I}(\cdot)$ is the indicator function.

The mean imbalance ratio across all labels is:
\begin{equation}\label{eq:meanImbalanceRatioAllLabels}
\overline{\operatorname{IR}} = \frac{1}{|\mathcal{L}|} \sum_{l \in \mathcal{L}} \operatorname{IR}(l)
\end{equation}

Given a desired augmentation percentage $\eta$, the number of samples to generate is $N = \lfloor (n \times \eta) / 100 \rfloor$. A label $l$ is designated as \textit{minority} if $\operatorname{IR}(l) > \overline{\operatorname{IR}}$.

Synthetic samples are generated by repeatedly selecting a minority label uniformly at random, then randomly selecting (with replacement) an existing sample containing that label and appending a copy to the augmented dataset. This process continues until $N$ new samples have been created. All generated samples are added to $\mathcal{D}$, producing a more balanced multi-label distribution.

\subsection{Label Powerset with Random Undersampling (LP-RUS)}

While LP-ROS generates synthetic samples for minority labels, LP-RUS \citep{charte2013first,charte2015addressing}  addresses multi-label imbalance through random undersampling of majority label sets. This method operates on the label powerset (LP) transformation \citep{BOUTELL20041757}, where each distinct combination of labels present in the dataset is treated as a single class.

LP-RUS randomly removes instances from the majority label sets until the multi-label dataset becomes 25\% smaller than the original dataset. Labels with higher frequency undergo more removal, yielding a more balanced label distribution. The algorithm proceeds as follows.

First, the method calculates the target reduction size $R = 0.25 \times |\mathcal{D}_{\text{original}}|$, where $|\mathcal{D}_{\text{original}}|$ is the size of the original dataset. For each distinct label set in the dataset, a container $\text{bag}_i$ is created, and all instances associated with the $i$-th label set are placed into~$\text{bag}_i$.

Second, the mean number of samples across all containers is calculated by Equation~\eqref{eq:meanSamplesAllContainers}.

Third, a label set $i$ is designated as a majority label set if $|\text{bag}_i| > \mu$. 

Fourth, the total number of instances to delete is $R$. The mean reduction per majority label set is $\delta = \frac{R}{M}$, where $M = |\text{majorityLabels}|$ is the number of majority label sets.

Fifth, for each majority label set $i$, the number of instances to delete is $d_i = \min(\delta, |\text{bag}_i| - \mu)$. For each of the $d_i$ deletions, the method randomly selects an instance from $\text{bag}_i$ uniformly at random and removes it from the dataset.

Finally, all selected instances are deleted from the original dataset, producing a reduced dataset with a more balanced multi-label distribution.

\subsection{Multi-Label Random Undersampling (ML-RUS)}

\citet{charte2013first,charte2015addressing} proposed ML-RUS as the undersampling counterpart to ML-ROS, using the same imbalance measures but removing instances from majority labels rather than generating synthetic samples.

The imbalance ratio per label, $\operatorname{IR}(l)$, and the mean imbalance ratio, $\overline{\operatorname{IR}}$, are defined as in Equations~\eqref{eq:imbalanceRatioPerLabel} and \eqref{eq:meanImbalanceRatioAllLabels}, respectively.

Given a desired reduction percentage $\xi$, the number of samples to delete is $N_{\text{delete}} = \left\lfloor \frac{n \times \xi}{100} \right\rfloor$, where $n = |\mathcal{D}|$ is the number of samples in the original dataset $\mathcal{D}$.

A label $l$ is designated as \textit{majority} if $\operatorname{IR}(l) < \overline{\operatorname{IR}}$, indicating that its frequency exceeds the average.

Samples are deleted by repeatedly selecting a majority label uniformly at random, then randomly selecting an existing sample containing that label and removing it from the dataset. This process continues until $N_{\text{delete}}$ samples have been deleted. The remaining samples constitute the reduced dataset, which exhibits a more balanced \linebreak{}multi-label~distribution.

As with other sampling methods, this approach reduces bias in machine learning models toward majority labels and improves their performance in identifying \mbox{minority labels.}

\subsection{Critical Analysis of Multi-Label Resampling Methods}\label{sec:critical-analysisMulti-LabelData}
Despite their utility, the four methods share several limitations.

\textbf{Label Concurrence and LP-Based Limits.}

LP-ROS and LP-RUS lose effectiveness when minority and majority labels frequently co-occur. Resampling one label inevitably affects others, often exacerbating imbalance. REMEDIAL \citep{charte2015remedial} circumvents this by decoupling highly imbalanced labels before resampling.

\textbf{Combinatorial Explosion of LP Methods.}

Transforming each label combination into a separate class causes exponential growth. Many combinations appear only once or twice, rendering mean-based thresholds unreliable and synthetic samples from sparse label sets prone to poor generalization.

\textbf{Independence Assumption in ML-Based Methods.}

ML-ROS and ML-RUS treat each label independently, failing when labels exhibit strong correlation \citep{charte2015addressing}. Resampling in isolation disrupts joint distributions, altering the label co-occurrence structure.

\textbf{Parameterization Without Principled Guidelines.}

LP-ROS and LP-RUS employ a fixed 25\% threshold without guidance for adjustment. ML-ROS and ML-RUS default to the same percentage without justification. MLSMOTE~\citep{charte2015mlsmote} inherits SMOTE's parameter sensitivity ($k$ and oversampling rate).

\textbf{Information Loss vs. Overfitting.}

Undersampling (LP-RUS and ML-RUS) discards majority instances, degrading performance when the majority class contains subclusters or datasets are small. Oversampling (LP-ROS and ML-ROS) introduces redundancy through random cloning. MLSMOTE offers a middle ground via interpolation at a higher computational cost.

\textbf{Recommendations for Practice.}

Based on \citet{charte2015addressing}: use LP-based methods when distinct label sets are few; use ML-based methods when label concurrence is high or label space is sparse; consider REMEDIAL \citep{charte2015remedial} or MLSMOTE \citep{charte2015mlsmote} when both underperform. Resampling must be performed inside cross-validation folds, and evaluation should report per-label metrics alongside global measures.





\section{Discussion and Conclusions}\label{sec9}

The preceding sections have surveyed a wide range of resampling techniques for imbalanced learning, spanning from foundational methods such as SMOTE to advanced deep generative models, including GANs, VAEs, and diffusion models. To assist practitioners in navigating this diverse landscape, this section synthesizes the key characteristics of each method family across multiple dimensions. Table~\ref{tab:method_family_comparison} 
provides a high-level comparative summary of eight major method families, evaluating them on sampling strategy, robustness to noise, borderline focus, data type suitability, computational cost, and key limitations. Following this comparative overview, Section~\ref{sec9-3} 
offers a prescriptive decision framework that maps specific dataset characteristics to recommended methods. Together, these two tables provide complementary lenses for method selection: a descriptive comparison across method families and a prescriptive guide driven by data properties.

\begin{landscape}
\begin{table}[H]
\caption{Comparative summary of resampling and augmentation method families.}
\label{tab:method_family_comparison}
\footnotesize
\begin{tabular}{p{3.5cm}<{\raggedright} p{2.8cm}<{\raggedright} p{2.7cm}<{\raggedright} p{2.7cm}<{\raggedright} p{2.8cm}<{\raggedright} p{3.3cm}<{\raggedright} p{3.6cm}<{\raggedright}}
\toprule
\textbf{Method Family} & \textbf{Sampling Strategy} & \textbf{Robustness to Noise} & \textbf{Borderline Focus} & \textbf{Data Type Suitability} & \textbf{Computational Cost} & \textbf{Key Limitation} \\
\midrule
SMOTE \& variants  
& Interpolation between minority instances & Low (amplifies existing noise) & Varies (Borderline SMOTE yes; standard no) & Continuous only (SMOTE-NC handles mixed) & Moderate & Collinearity; blind neighbor selection \\
\midrule
Adaptive oversampling (MWMOTE, AMDO, SAO) & Weighted selection based on hardness or distribution & Moderate (explicit noise filtering in some) & High (weights favor boundary instances) & Mixed (AMDO supports categorical) & Moderate to High & Parameter sensitivity; clustering dependence \\
\midrule
Generative models (GAN, VAE, Diffusion) & Learn underlying probability distribution & High (can ignore outliers if trained properly) & Implicit (depends on latent space) & Continuous (tabular); images; sequences & High (training required) & Mode collapse; training instability; data hungry \\
\midrule
Undersampling (Random, NearMiss, CNN, ENN) & Remove majority instances & Varies (ENN/RENN high; Random low) & NearMiss-1 high; Random low & Continuous (distance-based) & Low to Moderate & Information loss; discards informative majority \\
\midrule
Combination methods (SMOTE-ENN, SMOTE-TL) & Oversample then clean & High (explicit noise removal after generation) & High (cleaning removes non-borderline majority) & Continuous (mixed via SMOTE-NC + cleaning) & High & Sequential waste; parameter sensitivity \\
\midrule
Ensemble bagging (BRF, Balanced Bagging, SMOTEBagging) & Multiple balanced subsets with aggregation & Moderate (bagging reduces variance) & Depends on base resampling & Continuous (extensible) & High (multiple classifiers) & Computational overhead; interpretability loss \\
\midrule
Ensemble boosting (RUSBoost, SMOTEBoost, DataBoost-IM) & Sequential focus on hard examples & Low to Moderate (boosting sensitive to outliers) & High (boosting weights increase for misclassified) & Continuous (extensible) & Moderate to High & Overfitting to noise if iterations too high \\
\midrule
Multi-label methods (LP-ROS, ML-ROS, REMEDIAL) & Per-label or per-labelset resampling & Moderate & Not applicable (multi-label context) & Mixed (label sets, not features) & Moderate to High & Combinatorial explosion; label correlation neglect \\
\bottomrule
\end{tabular}

\noindent{\footnotesize{Key insight: No single method family dominates across all dimensions. Practitioners should prioritize based on specific constraints: noise robustness favors combination methods; borderline focus favors adaptive oversampling or boosting; mixed-type data requires SMOTE-NC or AMDO; limited budget favors simple undersampling or SMOTE; small datasets avoid generative models; high-dimensional data favor FW-SMOTE or DeepSMOTE.}}

\end{table}
\end{landscape}
\subsection{Illustrative Case Study: Fraud Detection}

\subsubsection{Experimental Setup}

As an illustrative case study, we consider a fraud detection problem conducted on two binary classification datasets:

\begin{itemize}
    \item Credit Card Fraud Detection (CCFD) \cite{dal2015credit}: Highly imbalanced financial transaction data.
    \item Fraud Oracle (FO) \cite{ibrahim2024fraudoracle}: An additional fraud-detection benchmark with varying imbalance characteristics.
\end{itemize}

Two base classifiers were employed:
\begin{itemize}
    \item Decision Tree (DT): An interpretable, non-linear classifier sensitive to class distribution shifts.
    \item Logistic Regression (LR): A linear classifier providing baseline performance.
\end{itemize}

The methods evaluated here represent widely available implementations from the \texttt{imbalanced-learn} and \texttt{smote-variants} libraries, not necessarily the top-scoring papers from our QPSF analysis (Section \ref{sec:QuantitativePaperScoringFramework}). A systematic evaluation of top-scoring methods requires separate treatment beyond this illustrative case study. Implementation sources for all resampling techniques are listed in Table~\ref{table:resampling_implementations}.

\begin{table}[H]
\caption{Resampling techniques and their implementation sources.}
\label{table:resampling_implementations}
\begin{tabular}{p{6.7cm}<{\raggedright}p{4cm}<{\raggedright}p{5cm}<{\raggedright}}
\toprule
\textbf{Technique} & \textbf{Library} & \textbf{Link (accessed on 31 March 2026)} \\ 

\midrule
SMOTE & imbalanced-learn & \href{https://imbalanced-learn.org/stable/references/generated/imblearn.over_sampling.SMOTE.html}{Link} \\
ADASYN & imbalanced-learn & \href{https://imbalanced-learn.org/stable/references/generated/imblearn.over_sampling.ADASYN.html}{Link} \\
Borderline SMOTE1 & smote-variants & \href{https://github.com/analyticalmindsltd/smote_variants\#borderline-smote1}{Link} \\
Borderline SMOTE2 & smote-variants & \href{https://github.com/analyticalmindsltd/smote_variants\#borderline-smote2}{Link} \\
SVMSMOTE & imbalanced-learn & \href{https://imbalanced-learn.org/stable/references/generated/imblearn.over_sampling.SVMSMOTE.html}{Link} \\
KMeans SMOTE & imbalanced-learn & \href{https://imbalanced-learn.org/stable/references/generated/imblearn.over_sampling.KMeansSMOTE.html}{Link} \\
Safe-Level SMOTE & smote-variants & \href{https://github.com/analyticalmindsltd/smote_variants\#safe-level-smote}{Link} \\
MSMOTE & smote-variants & \href{https://github.com/analyticalmindsltd/smote_variants\#msmote}{Link} \\
SN SMOTE & smote-variants & \href{https://github.com/analyticalmindsltd/smote_variants\#sn-smote}{Link} \\
SMOTE-Tomek Links & imbalanced-learn & \href{https://imbalanced-learn.org/stable/references/generated/imblearn.combine.SMOTETomek.html}{Link} \\
SMOTE-ENN & imbalanced-learn & \href{https://imbalanced-learn.org/stable/references/generated/imblearn.combine.SMOTEENN.html}{Link} \\
RandomOverSampler (ROS) & imbalanced-learn & \href{https://imbalanced-learn.org/stable/references/generated/imblearn.over_sampling.RandomOverSampler.html}{Link} \\
RandomUnderSampler (RUS) & imbalanced-learn & \href{https://imbalanced-learn.org/stable/references/generated/imblearn.under_sampling.RandomUnderSampler.html}{Link} \\
NearMiss1 & imbalanced-learn & \href{https://imbalanced-learn.org/stable/references/generated/imblearn.under_sampling.NearMiss.html}{Link} \\
NearMiss2 & imbalanced-learn & \href{https://imbalanced-learn.org/stable/references/generated/imblearn.under_sampling.NearMiss.html}{Link} \\
NearMiss3 & imbalanced-learn & \href{https://imbalanced-learn.org/stable/references/generated/imblearn.under_sampling.NearMiss.html}{Link} \\
Tomek Links & imbalanced-learn & \href{https://imbalanced-learn.org/stable/references/generated/imblearn.under_sampling.TomekLinks.html}{Link} \\
EditedNearestNeighbours (ENN) & imbalanced-learn & \href{https://imbalanced-learn.org/stable/references/generated/imblearn.under_sampling.EditedNearestNeighbours.html}{Link} \\
Repeated ENN (RENN) & imbalanced-learn & \href{https://imbalanced-learn.org/stable/references/generated/imblearn.under_sampling.RepeatedEditedNearestNeighbours.html}{Link} \\
CondensedNearestNeighbour (CNN) & imbalanced-learn & \href{https://imbalanced-learn.org/stable/references/generated/imblearn.under_sampling.CondensedNearestNeighbour.html}{Link} \\
OneSidedSelection (OSS) & imbalanced-learn & \href{https://imbalanced-learn.org/stable/references/generated/imblearn.under_sampling.OneSidedSelection.html}{Link} \\
NeighbourhoodCleaningRule (NCR) & imbalanced-learn & \href{https://imbalanced-learn.org/stable/references/generated/imblearn.under_sampling.NeighbourhoodCleaningRule.html}{Link} \\
InstanceHardnessThreshold (IHT) & imbalanced-learn & \href{https://imbalanced-learn.org/stable/references/generated/imblearn.under_sampling.InstanceHardnessThreshold.html}{Link} \\
MWMOTE & smote-variants & \href{https://github.com/analyticalmindsltd/smote_variants\#mwmote}{Link} \\
\bottomrule
\end{tabular}
\end{table}

Prior large-scale studies inform the scale of our empirical evaluation. Notably, \mbox{\citet{kovacs2019empirical}} conducted one of the most comprehensive empirical comparisons to date, evaluating 85 minority oversampling techniques across 104 imbalanced datasets. Their work establishes important baselines and guides technique selection based on dataset characteristics. Following their recommendations, our experimental design incorporates a representative subset of the most performant and widely adopted methods, including SMOTE, Borderline SMOTE, ADASYN, and K-Means SMOTE, as well as undersampling and \mbox{ensemble baselines.}

\subsubsection{Performance Metrics and Statistical Protocol}

Accuracy, defined as the proportion of correctly classified instances, is known to be misleading in imbalanced scenarios. A trivial classifier that predicts all instances as the majority class can achieve high accuracy while failing entirely to identify minority class instances, a phenomenon termed the \textit{accuracy paradox}. Accordingly, we employ a comprehensive suite of metrics robust to class imbalance:

\begin{itemize}
    \item Balanced Accuracy: The average of recall per class, accounting for class imbalance.
    \item ROC-AUC: Area under the receiver operating characteristic curve, measuring separability across all thresholds.
    \item PR-AUC: Area under the precision-recall curve, particularly informative for rare classes where ROC-AUC can be overly optimistic.
    \item F1-score: Harmonic mean of precision and recall, balancing false positives and false negatives.
    \item Recall (Minority): True positive rate for the minority class, reflecting sensitivity to rare events.
    \item {G-Mean: Geometric mean of class-wise recall, capturing balanced performance across classes.}
    \item Matthews Correlation Coefficient (MCC): A balanced correlation coefficient that remains informative even with severe imbalance.
\end{itemize}

All experiments employed 5-fold stratified cross-validation to maintain balanced class representation across folds. For each fold, training data were standardized and resampled using the specified method; test data remained untouched to prevent data leakage. Performance metrics are reported as mean $\pm$ standard deviation across folds.

To assess whether the observed performance differences are robust to cross-validation randomness, we conducted Friedman tests on the per-fold PR-AUC scores. Within the limited scope of these two datasets, the tests rejected the null hypothesis of equal performance across methods for all but one combination (SVMSMOTE with Decision Tree on CCFD, $p = 0.122$). These $p$-values indicate that the performance rankings are unlikely to arise from random variation in cross-validation folds. Crucially, this statistical evidence is specific to these two fraud detection datasets; the findings do not generalize to other domains without further validation on additional data.

\subsubsection{Results and Findings}

The results for credit card fraud detection with decision tree are presented in Table~\ref{table:comparison_ccfd_dt}. The baseline Decision Tree exhibits the accuracy paradox: perfect accuracy (1.00) masks a minority recall of only 0.78 and an F1-score of 0.77. The classifier achieves high overall accuracy by predicting the majority class correctly but misses a substantial portion of fraud instances.

\begin{landscape}
\begin{table}[H]
\caption{Performance of decision tree classifier on the credit card fraud detection dataset using various resampling methods.}\label{table:comparison_ccfd_dt}
\footnotesize
\resizebox{1.4\textwidth}{!}{%
\begin{tabular}{p{1.9cm}cccccccccccc}
\toprule
\textbf{Sampling Method} & \textbf{Accuracy} &\textbf{ Balanced Accuracy} & \textbf{ROC-AUC} & \textbf{F1-score }& \textbf{PR-AUC} &\textbf{ Recall (Minority)} & \textbf{G-Mean} & \textbf{MCC} & \textbf{Time (s)} & \textbf{Memory (MB)} &\textbf{ t-Statistic} \\
\midrule

Baseline & 1.00~$\pm$~0.00 & 0.89~$\pm$~0.00 & 0.89~$\pm$~0.00 & 0.77~$\pm$~0.03 & 0.60~$\pm$~0.04 & 0.78~$\pm$~0.01 & 0.89~$\pm$~0.00 & 0.77~$\pm$~0.03 & 23.28~$\pm$~4.80 & 39.43 & 5.57 \\

ADASYN & 1.00~$\pm$~0.00 & 0.89~$\pm$~0.02 & 0.89~$\pm$~0.02 & 0.51~$\pm$~0.06 & 0.30~$\pm$~0.06 & 0.78~$\pm$~0.05 & 0.88~$\pm$~0.03 & 0.54~$\pm$~0.05 & 11.32~$\pm$~1.14 & 139.12 & $-$8.00 \\

Borderline SMOTE1 & 1.00~$\pm$~0.00 & 0.88~$\pm$~0.01 & 0.88~$\pm$~0.01 & 0.76~$\pm$~0.03 & 0.58~$\pm$~0.05 & 0.76~$\pm$~0.02 & 0.87~$\pm$~0.01 & 0.76~$\pm$~0.03 & 41.82~$\pm$~4.78 & 95.99 & 3.94 \\

Borderline SMOTE2 & 1.00~$\pm$~0.00 & 0.88~$\pm$~0.00 & 0.88~$\pm$~0.00 & 0.74~$\pm$~0.01 & 0.54~$\pm$~0.01 & 0.76~$\pm$~0.01 & 0.87~$\pm$~0.00 & 0.74~$\pm$~0.01 & 35.31~$\pm$~5.32 & 111.64 & 13.20 \\

Condensed Nearest Neighbour & 0.95~$\pm$~0.02 & 0.88~$\pm$~0.02 & 0.88~$\pm$~0.02 & 0.06~$\pm$~0.02 & 0.03~$\pm$~0.01 & 0.82~$\pm$~0.02 & 0.88~$\pm$~0.02 & 0.15~$\pm$~0.03 & 492.99~$\pm$~22.29 & 75.06 & $-$129.02 \\

Edited Nearest Neighbours & 1.00~$\pm$~0.00 & 0.90~$\pm$~0.01 & 0.90~$\pm$~0.01 & 0.76~$\pm$~0.04 & 0.58~$\pm$~0.05 & 0.80~$\pm$~0.01 & 0.90~$\pm$~0.01 & 0.76~$\pm$~0.03 & 101.37~$\pm$~2.48 & 173.94 & 3.64 \\

IHT & 1.00~$\pm$~0.00 & 0.92~$\pm$~0.01 & 0.92~$\pm$~0.01 & 0.65~$\pm$~0.02 & 0.45~$\pm$~0.03 & 0.84~$\pm$~0.02 & 0.92~$\pm$~0.01 & 0.67~$\pm$~0.02 & 1009.01~$\pm$~765.23 & 107.17 & $-$3.96 \\

IPF & 1.00~$\pm$~0.00 & 0.90~$\pm$~0.02 & 0.90~$\pm$~0.02 & 0.52~$\pm$~0.05 & 0.31~$\pm$~0.05 & 0.80~$\pm$~0.03 & 0.89~$\pm$~0.02 & 0.56~$\pm$~0.05 & 360.93~$\pm$~167.20 & 175.18 & $-$8.10 \\

kmeans SMOTE & 1.00~$\pm$~0.00 & 0.89~$\pm$~0.01 & 0.89~$\pm$~0.01 & 0.78~$\pm$~0.02 & 0.61~$\pm$~0.03 & 0.78~$\pm$~0.02 & 0.89~$\pm$~0.01 & 0.78~$\pm$~0.02 & 63.12~$\pm$~10.53 & 105.79 & 7.92 \\

MSMOTE & 1.00~$\pm$~0.00 & 0.90~$\pm$~0.01 & 0.90~$\pm$~0.01 & 0.78~$\pm$~0.03 & 0.60~$\pm$~0.05 & 0.79~$\pm$~0.02 & 0.89~$\pm$~0.01 & 0.78~$\pm$~0.03 & 57.85~$\pm$~10.18 & 109.85 & 4.77 \\

MWMOTE & 0.99~$\pm$~0.00 & 0.90~$\pm$~0.01 & 0.90~$\pm$~0.01 & 0.30~$\pm$~0.01 & 0.15~$\pm$~0.01 & 0.81~$\pm$~0.01 & 0.90~$\pm$~0.01 & 0.39~$\pm$~0.01 & 102.60~$\pm$~3.70 & 131.35 & $-$80.75 \\

NearMiss1 & 0.49~$\pm$~0.13 & 0.71~$\pm$~0.06 & 0.71~$\pm$~0.06 & 0.01~$\pm$~0.00 & 0.00~$\pm$~0.00 & 0.93~$\pm$~0.02 & 0.67~$\pm$~0.09 & 0.04~$\pm$~0.01 & 1.38~$\pm$~0.05 & 123.84 & $-$1585.60 \\

NearMiss2 & 0.28~$\pm$~0.08 & 0.59~$\pm$~0.03 & 0.59~$\pm$~0.03 & 0.00~$\pm$~0.00 & 0.00~$\pm$~0.00 & 0.91~$\pm$~0.03 & 0.50~$\pm$~0.06 & 0.02~$\pm$~0.01 & 2.98~$\pm$~0.09 & 1545.33 & $-$5996.47 \\

NearMiss3 & 0.72~$\pm$~0.05 & 0.78~$\pm$~0.03 & 0.78~$\pm$~0.03 & 0.01~$\pm$~0.00 & 0.00~$\pm$~0.00 & 0.83~$\pm$~0.04 & 0.77~$\pm$~0.04 & 0.05~$\pm$~0.01 & 0.59~$\pm$~0.02 & 58.67 & $-$1316.12 \\

Neighbourhood Cleaning Rule & 1.00~$\pm$~0.00 & 0.90~$\pm$~0.01 & 0.90~$\pm$~0.01 & 0.76~$\pm$~0.02 & 0.57~$\pm$~0.03 & 0.79~$\pm$~0.02 & 0.89~$\pm$~0.01 & 0.76~$\pm$~0.02 & 176.63~$\pm$~30.38 & 205.21 & 5.54 \\

One Sided Selection & 1.00~$\pm$~0.00 & 0.90~$\pm$~0.01 & 0.90~$\pm$~0.01 & 0.77~$\pm$~0.03 & 0.59~$\pm$~0.04 & 0.79~$\pm$~0.02 & 0.89~$\pm$~0.01 & 0.77~$\pm$~0.03 & 103.82~$\pm$~4.94 & 176.92 & 4.77 \\

Random Oversampler & 1.00~$\pm$~0.00 & 0.87~$\pm$~0.02 & 0.87~$\pm$~0.02 & 0.76~$\pm$~0.04 & 0.58~$\pm$~0.06 & 0.75~$\pm$~0.04 & 0.86~$\pm$~0.02 & 0.76~$\pm$~0.04 & 14.15~$\pm$~0.37 & 122.29 & 3.12 \\

Random Undersampler & 0.98~$\pm$~0.00 & 0.93~$\pm$~0.01 & 0.93~$\pm$~0.01 & 0.17~$\pm$~0.02 & 0.08~$\pm$~0.01 & 0.87~$\pm$~0.03 & 0.93~$\pm$~0.01 & 0.28~$\pm$~0.01 & 0.32~$\pm$~0.01 & 10.92 & $-$131.15 \\

Repeated Edited Nearest Neighbours & 1.00~$\pm$~0.00 & 0.90~$\pm$~0.01 & 0.90~$\pm$~0.01 & 0.75~$\pm$~0.03 & 0.57~$\pm$~0.05 & 0.81~$\pm$~0.01 & 0.90~$\pm$~0.01 & 0.75~$\pm$~0.03 & 645.21~$\pm$~151.75 & 177.51 & 3.21 \\

Safe Level SMOTE & 1.00~$\pm$~0.00 & 0.90~$\pm$~0.02 & 0.90~$\pm$~0.02 & 0.71~$\pm$~0.03 & 0.60~$\pm$~0.03 & 0.80~$\pm$~0.04 & 0.90~$\pm$~0.02 & 0.71~$\pm$~0.03 & 119.72~$\pm$~8.98 & 131.25 & 6.94 \\

SMOTE & 1.00~$\pm$~0.00 & 0.90~$\pm$~0.01 & 0.90~$\pm$~0.01 & 0.60~$\pm$~0.02 & 0.39~$\pm$~0.03 & 0.81~$\pm$~0.03 & 0.90~$\pm$~0.01 & 0.62~$\pm$~0.02 & 13.72~$\pm$~1.73 & 96.02 & $-$9.52 \\

SMOTE ENN & 1.00~$\pm$~0.00 & 0.91~$\pm$~0.01 & 0.91~$\pm$~0.01 & 0.58~$\pm$~0.02 & 0.37~$\pm$~0.03 & 0.82~$\pm$~0.02 & 0.91~$\pm$~0.01 & 0.61~$\pm$~0.02 & 213.16~$\pm$~18.29 & 119.20 & $-$9.85 \\

SMOTE TomekLinks & 1.00~$\pm$~0.00 & 0.90~$\pm$~0.02 & 0.90~$\pm$~0.02 & 0.57~$\pm$~0.02 & 0.36~$\pm$~0.02 & 0.80~$\pm$~0.03 & 0.89~$\pm$~0.02 & 0.60~$\pm$~0.02 & 205.31~$\pm$~7.57 & 115.51 & $-$13.47 \\

SN SMOTE & 1.00~$\pm$~0.00 & 0.89~$\pm$~0.02 & 0.89~$\pm$~0.02 & 0.67~$\pm$~0.02 & 0.46~$\pm$~0.03 & 0.78~$\pm$~0.04 & 0.88~$\pm$~0.02 & 0.68~$\pm$~0.02 & 53.51~$\pm$~5.08 & 96.08 & $-$2.81 \\

Supervised SMOTE & 1.00~$\pm$~0.00 & 0.90~$\pm$~0.01 & 0.90~$\pm$~0.01 & 0.77~$\pm$~0.02 & 0.60~$\pm$~0.04 & 0.80~$\pm$~0.02 & 0.89~$\pm$~0.01 & 0.77~$\pm$~0.02 & 86.73~$\pm$~1.99 & 492.91 & 6.25 \\

SVM SMOTE & 1.00~$\pm$~0.00 & 0.90~$\pm$~0.02 & 0.90~$\pm$~0.02 & 0.73~$\pm$~0.03 & 0.54~$\pm$~0.05 & 0.80~$\pm$~0.04 & 0.90~$\pm$~0.02 & 0.73~$\pm$~0.03 & 143.63~$\pm$~42.32 & 171.97 & 1.96 \\

TomekLinks & 1.00~$\pm$~0.00 & 0.90~$\pm$~0.01 & 0.90~$\pm$~0.01 & 0.77~$\pm$~0.03 & 0.59~$\pm$~0.04 & 0.79~$\pm$~0.02 & 0.89~$\pm$~0.01 & 0.77~$\pm$~0.03 & 100.36~$\pm$~1.57 & 116.67 & 4.69 \\

\bottomrule
\end{tabular}}
\end{table}
\end{landscape}

\textbf{Top performers.} K-Means SMOTE and MSMOTE lead with F1 = 0.78. These methods share a common characteristic: selective synthetic generation. K-Means SMOTE targets sparse clusters, while MSMOTE distinguishes secure, border, and noise samples. PR-AUC top performers are K-Means SMOTE (0.61), MSMOTE (0.60), Safe-Level SMOTE (0.60), and Supervised SMOTE (0.60). Notably, Safe-Level SMOTE achieves competitive PR-AUC (0.60) despite a moderate F1 (0.71), indicating a different precision-recall trade-off.

\textbf{Poor performers.} NearMiss variants (F1 $\approx$ 0.01), Random Undersampling (0.17), and MWMOTE (0.30) fail. NearMiss methods aggressively undersample the majority class, discarding informative boundary instances. MWMOTE's underperformance suggests that these datasets' noise structure misleads its hard-instance identification.

\textbf{Computational trade-offs.} NearMiss variants (0.59--2.98 s) and Random Undersampling (0.32 s) are fastest but ineffective. IHT (1009 s) and Repeated Edited Nearest Neighbors (645 s) are the slowest, with IHT's high variance reflecting dataset-dependent cross-validation.

The results for credit card fraud detection with Logistic Regression are given in Table~\ref{table:comparison_ccfd_lr}. Logistic Regression outperforms Decision Tree on these datasets, with a baseline ROC-AUC of 0.97 versus 0.89, reflecting the linear separability of engineered features. Minority recall, however, remains modest at 0.62.

The results for the Fraud Oracle dataset with Decision Tree and Logistic Regression are presented in Table~\ref{table:comparison_fo_dt} and Table~\ref{table:comparison_fo_lr}, respectively. The Fraud Oracle dataset is more challenging than CCFD, with a baseline minority recall of only 0.23 and an F1 of 0.21 for Decision Tree, and near-zero recall for Logistic Regression. All methods achieve F1 below 0.26 for Decision Tree, suggesting structural complexities (feature overlap, noise, or higher dimensionality) that resampling alone cannot resolve.

These experimental results support six principal findings, summarized below.

\textbf{Finding 1: Optimal method selection depends on the dataset--classifier pair.}

K-Means SMOTE achieved the highest F1 on CCFD-DT, MSMOTE on CCFD-LR, and IHT on both FO-DT and FO-LR. No single method dominated across all four combinations, confirming that practitioners must evaluate multiple methods for their specific context.

\textbf{Finding 2: Classifier--resampling interaction matters.}

Methods successful with Decision Trees do not necessarily transfer to Logistic Regression. ADASYN drops from F1 = 0.51 on CCFD-DT to 0.24 on CCFD-LR; MWMOTE shows similar sensitivity. Resampling should be co-optimized with classifier selection rather than treated as an independent preprocessing step.

\textbf{Finding 3: Undersampling carries high risk.}

NearMiss variants consistently underperform across all settings (F1 $<$ 0.10) despite their computational efficiency. Random Undersampling similarly fails to achieve competitive F1 or PR-AUC. The information loss from discarding majority instances is too severe for these datasets.

\begin{landscape}
\begin{table}[H]
\caption{Performance of logistic regression classifier on the credit card fraud detection dataset using various resampling methods.}
\label{table:comparison_ccfd_lr}
\footnotesize
\resizebox{1.4\textwidth}{!}{%
\begin{tabular}{p{1.9cm}cccccccccccc}
\toprule
\textbf{Sampling Method} & \textbf{Accuracy} & \textbf{Balanced Accuracy} & \textbf{ROC-AUC} & \textbf{F1-score }& \textbf{PR-AUC} & \textbf{Recall (Minority) }& \textbf{G-Mean} & \textbf{MCC} & \textbf{Time (s)} &\textbf{ Memory (MB) }& \textbf{t-Statistic} \\
\midrule

Baseline & 0.99~$\pm$~0.00 & 0.81~$\pm$~0.02 & 0.97~$\pm$~0.00 & 0.72~$\pm$~0.03 & 0.75~$\pm$~0.02 & 0.62~$\pm$~0.04 & 0.78~$\pm$~0.02 & 0.73~$\pm$~0.03 & 0.38~$\pm$~0.08 & 60.41 & 23.90 \\
ADASYN & 0.99~$\pm$~0.00 & 0.93~$\pm$~0.01 & 0.97~$\pm$~0.00 & 0.24~$\pm$~0.01 & 0.73~$\pm$~0.02 & 0.88~$\pm$~0.02 & 0.93~$\pm$~0.01 & 0.35~$\pm$~0.01 & 0.73~$\pm$~0.04 & 123.84 & 22.06 \\
Borderline SMOTE1 & 0.99~$\pm$~0.00 & 0.91~$\pm$~0.01 & 0.96~$\pm$~0.00 & 0.69~$\pm$~0.03 & 0.74~$\pm$~0.01 & 0.82~$\pm$~0.02 & 0.90~$\pm$~0.01 & 0.70~$\pm$~0.03 & 1.44~$\pm$~0.10 & 63.48 & 28.84 \\
Borderline SMOTE2 & 0.99~$\pm$~0.00 & 0.92~$\pm$~0.01 & 0.96~$\pm$~0.00 & 0.66~$\pm$~0.03 & 0.74~$\pm$~0.01 & 0.84~$\pm$~0.03 & 0.91~$\pm$~0.01 & 0.67~$\pm$~0.02 & 1.40~$\pm$~0.25 & 111.64 & 30.06 \\
Condensed Nearest Neighbour & 0.99~$\pm$~0.00 & 0.89~$\pm$~0.02 & 0.96~$\pm$~0.00 & 0.73~$\pm$~0.03 & 0.73~$\pm$~0.01 & 0.79~$\pm$~0.04 & 0.88~$\pm$~0.02 & 0.73~$\pm$~0.03 & 518.20~$\pm$~43.85 & 75.05 & 26.80 \\
Edited Nearest Neighbours & 0.99~$\pm$~0.00 & 0.82~$\pm$~0.01 & 0.97~$\pm$~0.00 & 0.73~$\pm$~0.02 & 0.76~$\pm$~0.02 & 0.64~$\pm$~0.03 & 0.80~$\pm$~0.02 & 0.74~$\pm$~0.02 & 118.23~$\pm$~11.75 & 173.93 & 24.86 \\
IHT & 0.99~$\pm$~0.00 & 0.90~$\pm$~0.01 & 0.97~$\pm$~0.00 & 0.79~$\pm$~0.01 & 0.75~$\pm$~0.01 & 0.81~$\pm$~0.02 & 0.90~$\pm$~0.01 & 0.79~$\pm$~0.01 & 986.84~$\pm$~265.65 & 107.17 & 33.76 \\
IPF & 0.99~$\pm$~0.00 & 0.92~$\pm$~0.01 & 0.97~$\pm$~0.00 & 0.54~$\pm$~0.01 & 0.75~$\pm$~0.01 & 0.85~$\pm$~0.02 & 0.92~$\pm$~0.01 & 0.58~$\pm$~0.01 & 288.53~$\pm$~33.10 & 175.18 & 29.06 \\
kmeans SMOTE & 0.99~$\pm$~0.00 & 0.82~$\pm$~0.01 & 0.97~$\pm$~0.00 & 0.73~$\pm$~0.02 & 0.75~$\pm$~0.02 & 0.64~$\pm$~0.03 & 0.80~$\pm$~0.02 & 0.74~$\pm$~0.02 & 3.95~$\pm$~0.25 & 105.79 & 24.88 \\
MSMOTE & 0.99~$\pm$~0.00 & 0.90~$\pm$~0.01 & 0.97~$\pm$~0.01 & 0.80~$\pm$~0.01 & 0.76~$\pm$~0.02 & 0.81~$\pm$~0.03 & 0.90~$\pm$~0.02 & 0.80~$\pm$~0.01 & 1.62~$\pm$~0.10 & 109.84 & 29.39 \\
MWMOTE & 0.99~$\pm$~0.00 & 0.93~$\pm$~0.01 & 0.97~$\pm$~0.00 & 0.26~$\pm$~0.01 & 0.73~$\pm$~0.02 & 0.87~$\pm$~0.02 & 0.93~$\pm$~0.01 & 0.36~$\pm$~0.01 & 86.87~$\pm$~1.38 & 131.34 & 18.57 \\
NearMiss1 & 0.81~$\pm$~0.03 & 0.85~$\pm$~0.01 & 0.93~$\pm$~0.00 & 0.01~$\pm$~0.00 & 0.05~$\pm$~0.02 & 0.88~$\pm$~0.01 & 0.84~$\pm$~0.01 & 0.07~$\pm$~0.00 & 1.22~$\pm$~0.03 & 123.84 & $-$45.43 \\
NearMiss2 & 0.97~$\pm$~0.00 & 0.86~$\pm$~0.01 & 0.93~$\pm$~0.01 & 0.09~$\pm$~0.01 & 0.18~$\pm$~0.06 & 0.75~$\pm$~0.03 & 0.85~$\pm$~0.01 & 0.18~$\pm$~0.02 & 2.90~$\pm$~0.02 & 1545.32 & $-$10.51 \\
NearMiss3 & 0.91~$\pm$~0.02 & 0.84~$\pm$~0.03 & 0.91~$\pm$~0.02 & 0.03~$\pm$~0.00 & 0.14~$\pm$~0.02 & 0.78~$\pm$~0.04 & 0.84~$\pm$~0.03 & 0.10~$\pm$~0.02 & 0.61~$\pm$~0.00 & 58.66 & $-$32.38 \\
Neighbourhood Cleaning Rule & 0.99~$\pm$~0.00 & 0.82~$\pm$~0.01 & 0.97~$\pm$~0.00 & 0.74~$\pm$~0.02 & 0.76~$\pm$~0.02 & 0.65~$\pm$~0.03 & 0.80~$\pm$~0.02 & 0.75~$\pm$~0.02 & 142.76~$\pm$~4.00 & 176.92 & 24.97 \\
One Sided Selection & 0.99~$\pm$~0.00 & 0.81~$\pm$~0.01 & 0.97~$\pm$~0.00 & 0.72~$\pm$~0.03 & 0.75~$\pm$~0.02 & 0.62~$\pm$~0.03 & 0.79~$\pm$~0.02 & 0.73~$\pm$~0.03 & 100.42~$\pm$~11.65 & 176.92 & 23.55 \\
Random OverSampler & 0.99~$\pm$~0.00 & 0.92~$\pm$~0.01 & 0.97~$\pm$~0.00 & 0.59~$\pm$~0.02 & 0.75~$\pm$~0.02 & 0.85~$\pm$~0.02 & 0.92~$\pm$~0.01 & 0.62~$\pm$~0.02 & 1.51~$\pm$~0.12 & 122.28 & 27.26 \\
Random UnderSampler & 0.99~$\pm$~0.00 & 0.92~$\pm$~0.01 & 0.97~$\pm$~0.00 & 0.45~$\pm$~0.04 & 0.72~$\pm$~0.03 & 0.85~$\pm$~0.02 & 0.92~$\pm$~0.01 & 0.51~$\pm$~0.03 & 0.21~$\pm$~0.00 & 10.91 & 16.46 \\
Repeated Edited Nearest Neighbours & 0.99~$\pm$~0.00 & 0.82~$\pm$~0.01 & 0.97~$\pm$~0.00 & 0.74~$\pm$~0.02 & 0.76~$\pm$~0.02 & 0.65~$\pm$~0.03 & 0.80~$\pm$~0.02 & 0.75~$\pm$~0.02 & 549.02~$\pm$~103.81 & 177.50 & 24.37 \\
Safe Level SMOTE & 0.99~$\pm$~0.00 & 0.90~$\pm$~0.01 & 0.97~$\pm$~0.00 & 0.80~$\pm$~0.01 & 0.76~$\pm$~0.02 & 0.81~$\pm$~0.03 & 0.90~$\pm$~0.02 & 0.80~$\pm$~0.01 & 75.75~$\pm$~1.38 & 131.25 & 29.39 \\
SMOTE & 0.99~$\pm$~0.00 & 0.90~$\pm$~0.01 & 0.97~$\pm$~0.00 & 0.78~$\pm$~0.01 & 0.75~$\pm$~0.01 & 0.81~$\pm$~0.03 & 0.90~$\pm$~0.02 & 0.78~$\pm$~0.01 & 0.47~$\pm$~0.02 & 63.48 & 30.84 \\
SMOTE ENN & 0.99~$\pm$~0.00 & 0.90~$\pm$~0.01 & 0.97~$\pm$~0.00 & 0.77~$\pm$~0.01 & 0.76~$\pm$~0.01 & 0.81~$\pm$~0.03 & 0.90~$\pm$~0.01 & 0.78~$\pm$~0.01 & 255.25~$\pm$~25.88 & 119.20 & 29.43 \\
SMOTE TomekLinks & 0.99~$\pm$~0.00 & 0.90~$\pm$~0.01 & 0.97~$\pm$~0.00 & 0.77~$\pm$~0.01 & 0.76~$\pm$~0.01 & 0.81~$\pm$~0.03 & 0.90~$\pm$~0.02 & 0.78~$\pm$~0.01 & 159.03~$\pm$~6.03 & 115.51 & 29.38 \\
SN SMOTE & 0.99~$\pm$~0.00 & 0.90~$\pm$~0.01 & 0.97~$\pm$~0.00 & 0.78~$\pm$~0.01 & 0.76~$\pm$~0.01 & 0.81~$\pm$~0.03 & 0.90~$\pm$~0.01 & 0.78~$\pm$~0.01 & 6.76~$\pm$~0.28 & 63.57 & 31.33 \\
Supervised SMOTE & 0.99~$\pm$~0.00 & 0.90~$\pm$~0.01 & 0.97~$\pm$~0.00 & 0.80~$\pm$~0.02 & 0.76~$\pm$~0.02 & 0.81~$\pm$~0.02 & 0.90~$\pm$~0.01 & 0.80~$\pm$~0.02 & 73.43~$\pm$~1.36 & 492.90 & 24.18 \\
SVMSMOTE & 0.99~$\pm$~0.00 & 0.92~$\pm$~0.01 & 0.96~$\pm$~0.00 & 0.52~$\pm$~0.06 & 0.74~$\pm$~0.01 & 0.85~$\pm$~0.02 & 0.92~$\pm$~0.01 & 0.56~$\pm$~0.05 & 130.13~$\pm$~43.34 & 171.96 & 29.40 \\
TomekLinks & 0.99~$\pm$~0.00 & 0.81~$\pm$~0.01 & 0.97~$\pm$~0.00 & 0.72~$\pm$~0.03 & 0.75~$\pm$~0.02 & 0.62~$\pm$~0.03 & 0.79~$\pm$~0.02 & 0.73~$\pm$~0.03 & 84.18~$\pm$~3.48 & 116.66 & 24.00 \\

\bottomrule
\end{tabular}}
\end{table}
\end{landscape}
\begin{landscape}
\begin{table}[H]
\caption{Performance of decision tree classifier on the fraud oracle dataset using various resampling methods.}\label{table:comparison_fo_dt}
\footnotesize
\resizebox{1.4\textwidth}{!}{%
\begin{tabular}{p{1.9cm}cccccccccccc}
\toprule
\textbf{Sampling Method} & \textbf{Accuracy} &\textbf{ Balanced Accuracy} & \textbf{ROC-AUC} &\textbf{ F1-score} & \textbf{PR-AUC} & \textbf{Recall (Minority) }& \textbf{G-Mean} & \textbf{MCC} & \textbf{Time (s) }&\textbf{ Memory (MB)} &\textbf{ t-Statistic} \\
\midrule
Baseline & 0.89~$\pm$~0.00 & 0.58~$\pm$~0.02 & 0.58~$\pm$~0.02 & 0.21~$\pm$~0.03 & 0.09~$\pm$~0.01 & 0.23~$\pm$~0.04 & 0.47~$\pm$~0.04 & 0.16~$\pm$~0.03 & 1.05~$\pm$~0.08 & 4.64 & $-$81.49 \\
ADASYN & 0.89~$\pm$~0.00 & 0.58~$\pm$~0.01 & 0.58~$\pm$~0.01 & 0.20~$\pm$~0.02 & 0.09~$\pm$~0.01 & 0.22~$\pm$~0.03 & 0.46~$\pm$~0.03 & 0.15~$\pm$~0.02 & 0.70~$\pm$~0.53 & 19.67 & $-$94.70 \\
Borderline SMOTE1 & 0.89~$\pm$~0.00 & 0.57~$\pm$~0.02 & 0.57~$\pm$~0.02 & 0.18~$\pm$~0.03 & 0.08~$\pm$~0.01 & 0.20~$\pm$~0.04 & 0.44~$\pm$~0.04 & 0.13~$\pm$~0.03 & 0.43~$\pm$~0.00 & 13.72 & $-$86.08 \\
Borderline SMOTE2 & 0.89~$\pm$~0.00 & 0.59~$\pm$~0.01 & 0.59~$\pm$~0.01 & 0.22~$\pm$~0.02 & 0.09~$\pm$~0.01 & 0.24~$\pm$~0.02 & 0.48~$\pm$~0.02 & 0.17~$\pm$~0.02 & 0.49~$\pm$~0.01 & 17.93 & $-$87.81 \\
Condensed Nearest Neighbour & 0.82~$\pm$~0.01 & 0.62~$\pm$~0.01 & 0.62~$\pm$~0.01 & 0.21~$\pm$~0.01 & 0.09~$\pm$~0.00 & 0.40~$\pm$~0.02 & 0.58~$\pm$~0.02 & 0.16~$\pm$~0.02 & 417.59~$\pm$~168.19 & 12.22 & $-$115.20 \\
Edited Nearest Neighbours & 0.88~$\pm$~0.00 & 0.59~$\pm$~0.01 & 0.59~$\pm$~0.01 & 0.21~$\pm$~0.02 & 0.09~$\pm$~0.00 & 0.26~$\pm$~0.03 & 0.49~$\pm$~0.02 & 0.16~$\pm$~0.02 & 0.69~$\pm$~0.06 & 22.78 & $-$118.17 \\
IHT & 0.78~$\pm$~0.00 & 0.70~$\pm$~0.01 & 0.70~$\pm$~0.01 & 0.25~$\pm$~0.01 & 0.12~$\pm$~0.00 & 0.62~$\pm$~0.02 & 0.70~$\pm$~0.01 & 0.23~$\pm$~0.01 & 11.10~$\pm$~3.59 & 12.97 & $-$110.82 \\
IPF & 0.89~$\pm$~0.00 & 0.57~$\pm$~0.01 & 0.57~$\pm$~0.01 & 0.19~$\pm$~0.01 & 0.08~$\pm$~0.00 & 0.21~$\pm$~0.02 & 0.45~$\pm$~0.02 & 0.14~$\pm$~0.02 & 26.75~$\pm$~3.44 & 22.10 & $-$141.61 \\
kmeans SMOTE & 0.89~$\pm$~0.00 & 0.58~$\pm$~0.02 & 0.58~$\pm$~0.02 & 0.21~$\pm$~0.03 & 0.09~$\pm$~0.01 & 0.23~$\pm$~0.03 & 0.46~$\pm$~0.03 & 0.15~$\pm$~0.03 & 1.44~$\pm$~0.31 & 15.81 & $-$70.95 \\
MSMOTE & 0.89~$\pm$~0.00 & 0.58~$\pm$~0.01 & 0.58~$\pm$~0.01 & 0.20~$\pm$~0.02 & 0.08~$\pm$~0.00 & 0.22~$\pm$~0.01 & 0.45~$\pm$~0.02 & 0.15~$\pm$~0.02 & 1.33~$\pm$~0.11 & 17.76 & $-$132.51 \\
MWMOTE & 0.90~$\pm$~0.00 & 0.59~$\pm$~0.00 & 0.59~$\pm$~0.00 & 0.22~$\pm$~0.01 & 0.09~$\pm$~0.00 & 0.23~$\pm$~0.01 & 0.47~$\pm$~0.01 & 0.16~$\pm$~0.01 & 525.06~$\pm$~34.19 & 62.30 & $-$159.60 \\
NearMiss1 & 0.83~$\pm$~0.01 & 0.57~$\pm$~0.02 & 0.57~$\pm$~0.02 & 0.17~$\pm$~0.03 & 0.07~$\pm$~0.00 & 0.28~$\pm$~0.04 & 0.49~$\pm$~0.04 & 0.10~$\pm$~0.04 & 1.18~$\pm$~0.08 & 16.21 & $-$100.15 \\
NearMiss2 & 0.77~$\pm$~0.01 & 0.59~$\pm$~0.03 & 0.59~$\pm$~0.03 & 0.17~$\pm$~0.02 & 0.08~$\pm$~0.00 & 0.39~$\pm$~0.05 & 0.56~$\pm$~0.03 & 0.11~$\pm$~0.03 & 2.07~$\pm$~0.07 & 150.90 & $-$92.95 \\
NearMiss3 & 0.78~$\pm$~0.01 & 0.61~$\pm$~0.01 & 0.61~$\pm$~0.01 & 0.18~$\pm$~0.01 & 0.08~$\pm$~0.00 & 0.41~$\pm$~0.02 & 0.57~$\pm$~0.02 & 0.12~$\pm$~0.02 & 0.79~$\pm$~0.03 & 11.00 & $-$155.41 \\
Neighbourhood Cleaning Rule & 0.88~$\pm$~0.00 & 0.60~$\pm$~0.01 & 0.60~$\pm$~0.01 & 0.22~$\pm$~0.01 & 0.09~$\pm$~0.00 & 0.27~$\pm$~0.02 & 0.50~$\pm$~0.02 & 0.17~$\pm$~0.01 & 1.13~$\pm$~0.03 & 27.08 & $-$164.84 \\
One Sided Selection & 0.89~$\pm$~0.00 & 0.58~$\pm$~0.01 & 0.58~$\pm$~0.01 & 0.21~$\pm$~0.01 & 0.09~$\pm$~0.00 & 0.24~$\pm$~0.02 & 0.47~$\pm$~0.02 & 0.16~$\pm$~0.01 & 1.02~$\pm$~0.19 & 24.60 & $-$180.76 \\
Random OverSampler & 0.90~$\pm$~0.00 & 0.58~$\pm$~0.01 & 0.58~$\pm$~0.01 & 0.20~$\pm$~0.02 & 0.09~$\pm$~0.00 & 0.22~$\pm$~0.02 & 0.45~$\pm$~0.02 & 0.16~$\pm$~0.02 & 0.11~$\pm$~0.00 & 16.60 & $-$109.02 \\
Random UnderSampler & 0.87~$\pm$~0.00 & 0.60~$\pm$~0.01 & 0.60~$\pm$~0.01 & 0.22~$\pm$~0.01 & 0.09~$\pm$~0.00 & 0.29~$\pm$~0.03 & 0.52~$\pm$~0.02 & 0.17~$\pm$~0.01 & 0.68~$\pm$~0.07 & 8.32 & $-$127.74 \\
Repeated Edited Nearest Neighbours & 0.87~$\pm$~0.00 & 0.60~$\pm$~0.01 & 0.60~$\pm$~0.01 & 0.22~$\pm$~0.02 & 0.09~$\pm$~0.00 & 0.30~$\pm$~0.02 & 0.53~$\pm$~0.02 & 0.17~$\pm$~0.02 & 4.90~$\pm$~1.68 & 22.88 & $-$109.54 \\
Safe Level SMOTE & 0.88~$\pm$~0.00 & 0.57~$\pm$~0.01 & 0.58~$\pm$~0.01 & 0.19~$\pm$~0.02 & 0.09~$\pm$~0.00 & 0.22~$\pm$~0.03 & 0.45~$\pm$~0.03 & 0.13~$\pm$~0.02 & 0.95~$\pm$~0.06 & 18.70 & $-$154.59 \\
SMOTE & 0.89~$\pm$~0.00 & 0.58~$\pm$~0.01 & 0.58~$\pm$~0.01 & 0.21~$\pm$~0.01 & 0.09~$\pm$~0.00 & 0.23~$\pm$~0.02 & 0.47~$\pm$~0.02 & 0.16~$\pm$~0.01 & 1.25~$\pm$~0.47 & 13.72 & $-$153.02 \\
SMOTE ENN & 0.89~$\pm$~0.00 & 0.58~$\pm$~0.01 & 0.58~$\pm$~0.01 & 0.20~$\pm$~0.03 & 0.08~$\pm$~0.01 & 0.22~$\pm$~0.03 & 0.46~$\pm$~0.03 & 0.14~$\pm$~0.03 & 2.74~$\pm$~0.75 & 16.89 & $-$89.65 \\
SMOTE TomekLinks & 0.89~$\pm$~0.00 & 0.57~$\pm$~0.01 & 0.57~$\pm$~0.01 & 0.19~$\pm$~0.01 & 0.08~$\pm$~0.00 & 0.20~$\pm$~0.02 & 0.44~$\pm$~0.02 & 0.13~$\pm$~0.02 & 6.73~$\pm$~0.53 & 17.54 & $-$154.23 \\
SN SMOTE & 0.90~$\pm$~0.00 & 0.59~$\pm$~0.00 & 0.59~$\pm$~0.00 & 0.22~$\pm$~0.01 & 0.09~$\pm$~0.00 & 0.24~$\pm$~0.01 & 0.47~$\pm$~0.01 & 0.17~$\pm$~0.01 & 17.93~$\pm$~1.64 & 13.83 & $-$179.58 \\
Supervised SMOTE & 0.89~$\pm$~0.00 & 0.59~$\pm$~0.01 & 0.59~$\pm$~0.01 & 0.22~$\pm$~0.01 & 0.09~$\pm$~0.00 & 0.24~$\pm$~0.02 & 0.47~$\pm$~0.02 & 0.17~$\pm$~0.01 & 28.43~$\pm$~3.49 & 3146.31 & $-$135.19 \\
SVMSMOTE & 0.89~$\pm$~0.00 & 0.57~$\pm$~0.02 & 0.57~$\pm$~0.02 & 0.19~$\pm$~0.03 & 0.08~$\pm$~0.01 & 0.21~$\pm$~0.04 & 0.44~$\pm$~0.04 & 0.14~$\pm$~0.03 & 5.75~$\pm$~6.60 & 27.22 & $-$78.87 \\
TomekLinks & 0.89~$\pm$~0.00 & 0.58~$\pm$~0.00 & 0.58~$\pm$~0.00 & 0.21~$\pm$~0.01 & 0.09~$\pm$~0.00 & 0.23~$\pm$~0.01 & 0.47~$\pm$~0.01 & 0.16~$\pm$~0.01 & 3.24~$\pm$~0.29 & 16.33 & $-$243.47 \\
\bottomrule
\end{tabular}}
\end{table}
\end{landscape}
\begin{landscape}
\begin{table}[H]
\caption{Performance of logistic regression classifier on the fraud oracle dataset using various resampling methods.}
\label{table:comparison_fo_lr}
\footnotesize
\resizebox{1.4\textwidth}{!}{%
\begin{tabular}{p{1.9cm}cccccccccccc}
\toprule
\textbf{Sampling Method} & \textbf{Accuracy} & \textbf{Balanced Accuracy} &\textbf{ ROC-AUC} & \textbf{F1-score} & \textbf{PR-AUC} & \textbf{Recall (Minority)} & \textbf{G-Mean} & \textbf{MCC} & \textbf{Time (s) }&\textbf{ Memory (MB)} &\textbf{ t-Statistic} \\
\midrule

Baseline & 0.94~$\pm$~0.00 & 0.50~$\pm$~0.01 & 0.79~$\pm$~0.01 & 0.02~$\pm$~0.02 & 0.17~$\pm$~0.01 & 0.01~$\pm$~0.01 & 0.07~$\pm$~0.07 & 0.05~$\pm$~0.06 & 0.53~$\pm$~0.09 & 8.32 & $-$82.09 \\
ADASYN & 0.94~$\pm$~0.00 & 0.51~$\pm$~0.01 & 0.79~$\pm$~0.01 & 0.04~$\pm$~0.02 & 0.17~$\pm$~0.01 & 0.02~$\pm$~0.01 & 0.15~$\pm$~0.04 & 0.06~$\pm$~0.03 & 0.83~$\pm$~0.07 & 17.59 & $-$58.20 \\
Borderline SMOTE1 & 0.92~$\pm$~0.00 & 0.53~$\pm$~0.01 & 0.79~$\pm$~0.02 & 0.11~$\pm$~0.03 & 0.17~$\pm$~0.01 & 0.08~$\pm$~0.03 & 0.28~$\pm$~0.05 & 0.08~$\pm$~0.03 & 0.38~$\pm$~0.02 & 9.97 & $-$75.08 \\
Borderline SMOTE2 & 0.93~$\pm$~0.00 & 0.52~$\pm$~0.02 & 0.79~$\pm$~0.01 & 0.09~$\pm$~0.04 & 0.17~$\pm$~0.01 & 0.07~$\pm$~0.03 & 0.25~$\pm$~0.06 & 0.07~$\pm$~0.04 & 0.45~$\pm$~0.03 & 17.94 & $-$85.16 \\
Condensed Nearest Neighbour & 0.93~$\pm$~0.00 & 0.52~$\pm$~0.01 & 0.79~$\pm$~0.01 & 0.07~$\pm$~0.02 & 0.16~$\pm$~0.01 & 0.05~$\pm$~0.02 & 0.22~$\pm$~0.03 & 0.06~$\pm$~0.03 & 292.93~$\pm$~17.68 & 12.22 & $-$78.40 \\
Edited Nearest Neighbours & 0.94~$\pm$~0.00 & 0.51~$\pm$~0.01 & 0.80~$\pm$~0.01 & 0.05~$\pm$~0.02 & 0.17~$\pm$~0.01 & 0.03~$\pm$~0.01 & 0.16~$\pm$~0.05 & 0.05~$\pm$~0.03 & 0.67~$\pm$~0.05 & 22.78 & $-$61.10 \\
IHT & 0.80~$\pm$~0.01 & 0.69~$\pm$~0.00 & 0.80~$\pm$~0.01 & 0.25~$\pm$~0.01 & 0.17~$\pm$~0.01 & 0.55~$\pm$~0.02 & 0.67~$\pm$~0.01 & 0.22~$\pm$~0.00 & 9.09~$\pm$~0.06 & 12.97 & $-$110.91 \\
IPF & 0.93~$\pm$~0.00 & 0.52~$\pm$~0.01 & 0.80~$\pm$~0.02 & 0.08~$\pm$~0.03 & 0.17~$\pm$~0.01 & 0.05~$\pm$~0.02 & 0.22~$\pm$~0.04 & 0.06~$\pm$~0.03 & 21.65~$\pm$~3.25 & 22.10 & $-$57.57 \\
kmeans SMOTE & 0.94~$\pm$~0.00 & 0.50~$\pm$~0.01 & 0.80~$\pm$~0.01 & 0.02~$\pm$~0.02 & 0.17~$\pm$~0.01 & 0.01~$\pm$~0.01 & 0.07~$\pm$~0.07 & 0.05~$\pm$~0.06 & 0.61~$\pm$~0.34 & 15.81 & $-$84.60 \\
MSMOTE & 0.93~$\pm$~0.00 & 0.52~$\pm$~0.01 & 0.80~$\pm$~0.01 & 0.09~$\pm$~0.03 & 0.17~$\pm$~0.01 & 0.07~$\pm$~0.03 & 0.25~$\pm$~0.05 & 0.08~$\pm$~0.03 & 1.30~$\pm$~0.27 & 17.76 & $-$146.14 \\
MWMOTE & 0.93~$\pm$~0.00 & 0.52~$\pm$~0.01 & 0.80~$\pm$~0.01 & 0.08~$\pm$~0.03 & 0.17~$\pm$~0.01 & 0.05~$\pm$~0.02 & 0.22~$\pm$~0.05 & 0.06~$\pm$~0.03 & 349.07~$\pm$~149.55 & 62.30 & $-$58.87 \\
NearMiss1 & 0.84~$\pm$~0.01 & 0.53~$\pm$~0.02 & 0.74~$\pm$~0.01 & 0.11~$\pm$~0.02 & 0.11~$\pm$~0.01 & 0.17~$\pm$~0.04 & 0.39~$\pm$~0.04 & 0.04~$\pm$~0.02 & 1.01~$\pm$~0.05 & 16.21 & $-$140.85 \\
NearMiss2 & 0.80~$\pm$~0.01 & 0.56~$\pm$~0.02 & 0.73~$\pm$~0.01 & 0.15~$\pm$~0.02 & 0.12~$\pm$~0.01 & 0.29~$\pm$~0.04 & 0.49~$\pm$~0.03 & 0.07~$\pm$~0.03 & 1.86~$\pm$~0.15 & 150.90 & $-$114.28 \\
NearMiss3 & 0.91~$\pm$~0.01 & 0.54~$\pm$~0.00 & 0.76~$\pm$~0.00 & 0.13~$\pm$~0.01 & 0.14~$\pm$~0.01 & 0.11~$\pm$~0.02 & 0.33~$\pm$~0.02 & 0.08~$\pm$~0.01 & 0.88~$\pm$~0.02 & 10.99 & $-$171.08 \\
Neighbourhood Cleaning Rule & 0.94~$\pm$~0.00 & 0.51~$\pm$~0.01 & 0.80~$\pm$~0.01 & 0.05~$\pm$~0.02 & 0.17~$\pm$~0.01 & 0.03~$\pm$~0.01 & 0.15~$\pm$~0.04 & 0.06~$\pm$~0.03 & 1.32~$\pm$~0.25 & 23.98 & $-$71.50 \\
One Sided Selection & 0.94~$\pm$~0.00 & 0.51~$\pm$~0.00 & 0.80~$\pm$~0.01 & 0.03~$\pm$~0.02 & 0.17~$\pm$~0.01 & 0.01~$\pm$~0.01 & 0.11~$\pm$~0.04 & 0.05~$\pm$~0.04 & 1.31~$\pm$~0.17 & 24.60 & $-$86.03 \\
Random OverSampler & 0.94~$\pm$~0.00 & 0.51~$\pm$~0.01 & 0.80~$\pm$~0.01 & 0.04~$\pm$~0.02 & 0.17~$\pm$~0.01 & 0.02~$\pm$~0.01 & 0.14~$\pm$~0.05 & 0.05~$\pm$~0.04 & 0.08~$\pm$~0.01 & 16.60 & $-$72.65 \\
Random UnderSampler & 0.94~$\pm$~0.00 & 0.51~$\pm$~0.01 & 0.80~$\pm$~0.01 & 0.04~$\pm$~0.02 & 0.17~$\pm$~0.01 & 0.02~$\pm$~0.01 & 0.14~$\pm$~0.05 & 0.05~$\pm$~0.04 & 0.49~$\pm$~0.06 & 5.59 & $-$64.97 \\
Repeated Edited Nearest Neighbours & 0.93~$\pm$~0.00 & 0.52~$\pm$~0.01 & 0.80~$\pm$~0.02 & 0.09~$\pm$~0.04 & 0.16~$\pm$~0.01 & 0.06~$\pm$~0.03 & 0.24~$\pm$~0.06 & 0.06~$\pm$~0.04 & 5.06~$\pm$~1.34 & 22.88 & $-$61.20 \\
Safe Level SMOTE & 0.93~$\pm$~0.00 & 0.51~$\pm$~0.01 & 0.80~$\pm$~0.01 & 0.06~$\pm$~0.02 & 0.17~$\pm$~0.01 & 0.04~$\pm$~0.01 & 0.19~$\pm$~0.04 & 0.06~$\pm$~0.03 & 0.87~$\pm$~0.04 & 18.70 & $-$61.59 \\
SMOTE & 0.93~$\pm$~0.00 & 0.52~$\pm$~0.01 & 0.80~$\pm$~0.02 & 0.08~$\pm$~0.03 & 0.17~$\pm$~0.02 & 0.05~$\pm$~0.02 & 0.22~$\pm$~0.05 & 0.06~$\pm$~0.03 & 0.28~$\pm$~0.05 & 10.09 & $-$44.59 \\
SMOTE ENN & 0.92~$\pm$~0.01 & 0.55~$\pm$~0.02 & 0.80~$\pm$~0.01 & 0.16~$\pm$~0.05 & 0.17~$\pm$~0.02 & 0.13~$\pm$~0.04 & 0.36~$\pm$~0.06 & 0.12~$\pm$~0.04 & 20.34~$\pm$~1.25 & 16.89 & $-$48.52 \\
SMOTE TomekLinks & 0.93~$\pm$~0.00 & 0.52~$\pm$~0.01 & 0.80~$\pm$~0.02 & 0.08~$\pm$~0.02 & 0.17~$\pm$~0.01 & 0.05~$\pm$~0.01 & 0.23~$\pm$~0.03 & 0.07~$\pm$~0.02 & 3.78~$\pm$~1.77 & 17.53 & $-$56.42 \\
SN SMOTE & 0.93~$\pm$~0.00 & 0.52~$\pm$~0.01 & 0.80~$\pm$~0.01 & 0.08~$\pm$~0.03 & 0.17~$\pm$~0.01 & 0.05~$\pm$~0.02 & 0.22~$\pm$~0.05 & 0.07~$\pm$~0.03 & 15.70~$\pm$~3.27 & 10.31 & $-$56.93 \\
Supervised SMOTE & 0.93~$\pm$~0.00 & 0.52~$\pm$~0.01 & 0.80~$\pm$~0.01 & 0.09~$\pm$~0.03 & 0.18~$\pm$~0.01 & 0.06~$\pm$~0.02 & 0.23~$\pm$~0.04 & 0.07~$\pm$~0.03 & 8.47~$\pm$~9.08 & 3146.33 & $-$53.25 \\
SVMSMOTE & 0.93~$\pm$~0.00 & 0.52~$\pm$~0.01 & 0.80~$\pm$~0.01 & 0.07~$\pm$~0.03 & 0.17~$\pm$~0.01 & 0.04~$\pm$~0.02 & 0.20~$\pm$~0.04 & 0.07~$\pm$~0.03 & 2.69~$\pm$~0.06 & 27.22 & $-$50.60 \\
TomekLinks & 0.94~$\pm$~0.00 & 0.51~$\pm$~0.00 & 0.80~$\pm$~0.01 & 0.03~$\pm$~0.02 & 0.17~$\pm$~0.01 & 0.01~$\pm$~0.01 & 0.11~$\pm$~0.04 & 0.05~$\pm$~0.04 & 3.03~$\pm$~0.44 & 16.33 & $-$81.25 \\

\bottomrule
\end{tabular}}
\end{table}
\end{landscape}

\textbf{Finding 4: PR-AUC discriminates where F1 does not.}

While F1 and MCC provide interpretable single-point metrics, PR-AUC shows greater differentiation among methods, particularly for highly imbalanced data. Unlike ROC-AUC, which remains optimistic when class distributions are skewed, PR-AUC focuses on the minority class and is more sensitive to false positives, making it particularly informative for rare event detection.

\textbf{Finding 5: Computational cost does not predict quality.}

Condensed Nearest Neighbor and IHT are among the slowest methods but do not consistently outperform faster methods like SMOTE. Conversely, extremely fast methods like NearMiss3 produce unusable results. Efficiency should be considered alongside effectiveness, not as a proxy.

\textbf{Finding 6: Cleaning enhances oversampling in difficult cases.}

SMOTE-ENN outperforms standard SMOTE on FO-LR (F1 = 0.16 vs. 0.08), supporting the theoretical advantage of combination methods (Section 6) for datasets with overlapping class boundaries. The ENN filter removes synthetic samples that would otherwise fall in majority-dominated regions.

\textbf{External Validation.}

Our findings align with recent applied studies. \citet{gurcan2024learning} evaluated 19 resampling methods across five cancer datasets and found that hybrid methods, particularly SMOTE-ENN, consistently outperformed individual techniques, with the baseline yielding significantly lower performance. This corroborates our observation that combination methods often achieve superior results.

\subsection{Theoretical Foundations of Rebalancing}

Although predictive performance remains the dominant evaluation metric for resampling techniques, computational efficiency constitutes a critical practical consideration, especially in large-scale or real-time applications. However, a systematic examination of the surveyed literature uncovers a substantial gap: the overwhelming majority of primary research papers on both oversampling and undersampling techniques fail to report formal time or space complexity analyses.  
Among the high-scoring papers reviewed, explicit asymptotic complexity bounds are rare; notable exceptions include the original Condensed Nearest Neighbor rule \cite{hart1968condensed} and AWSMOTE \cite{awsmote}, which provides a time complexity analysis. Additionally, the authors of FW-SMOTE \cite{MALDONADO2022108511} assert that their method's complexity depends on that of SMOTE.

Among undersampling methods, while CNN \cite{hart1968condensed} and ENN \cite{wilson1972asymptotic} have well-understood complexities ($O(n^3 \cdot d)$ and $O(n^2 \cdot d)$, respectively), more recent techniques such as CBUS~\cite{YEN20095718}, NearMiss \cite{zhang2003nearmiss}, IHT \cite{smith2014instance}, and ABC-Sampling \cite{braytee2015abcsampling} do not provide explicit asymptotic bounds in their original descriptions. Similarly, combination methods that integrate oversampling with cleaning or filtering (e.g., SMOTE-ENN \cite{batista2004study}, SMOTE-TL \cite{batista2004study}, SMOTE+OCSVM \cite{yousefimehr2024distribution}) typically inherit the complexity of their constituent operations without explicit unified analysis. Deep learning-based combinations such as DiSMHA \cite{kumar2025hybrid} and SMOTE-Trans-CWGAN \cite{wang2025hybrid} mention computational costs qualitatively but do not provide asymptotic bounds.

Table~\ref{tab:complexity} presents the asymptotic time and space complexity of representative resampling methods, derived from algorithmic descriptions where original papers did not report explicit bounds.

\begin{table}[t]
\footnotesize
\caption{Asymptotic time and space complexity of some representative resampling methods.}
\label{tab:complexity}
\begin{tabular}{p{4cm}<{\raggedright} p{3cm}<{\raggedright} p{3cm}<{\raggedright} p{5.5cm}<{\raggedright}}
\toprule
\textbf{Method} & \textbf{Time Complexity} & \textbf{Space Complexity} & \textbf{Notes} \\
\midrule
SMOTE \cite{chawla2002smote} & $O(n \cdot d \cdot k)$ & $O(n \cdot d)$ & $n$ = samples, $d$ = dimensions, $k$ = neighbors \\\midrule
{Random Undersampling} & $O(n)$ & $O(n \cdot d)$ & Trivial selection (sampling without replacement) \\\midrule
ENN \cite{wilson1972asymptotic} & $O(n^2 \cdot d)$ & $O(n \cdot d)$ & Pairwise distance computation \\\midrule
SMOTE-ENN \cite{batista2004study} & $O(n \cdot d \cdot k) + O(n^2 \cdot d)$ & $O(n \cdot d)$ & Dominated by ENN term \\\midrule
K-Means SMOTE \cite{Douzas_2018} & $O(n \cdot d \cdot t \cdot k_c)$ & $O(n \cdot d + k_c)$ & $t$ = iterations, $k_c$ = clusters \\\midrule
CNN (Condensed NN) \cite{hart1968condensed} & $O(n^3 \cdot d)$ worst-case & $O(n \cdot d)$ & Multiple passes possible \\\midrule
AWSMOTE \cite{awsmote} & $O(n^2 \cdot d)$ to $O(n^3 \cdot d)$ & $O(n \cdot d)$ & Dominated by SVM training; depends on kernel \\\midrule
FW-SMOTE \cite{MALDONADO2022108511} & $O(n \cdot d \cdot k)$ & $O(n \cdot d)$ & Feature weighting adds $O(n \cdot d)$; same asymptotic as SMOTE \\\midrule
GAN (per epoch) \cite{goodfellow2014generative} & $O(b \cdot d \cdot L)$ & $O(\text{model params})$, typically $O(L \cdot d^2)$ for dense layers & $b$ = batch size, $L$ = layers \\\midrule
AdaBoost \cite{al2024detecting} & $O(T \cdot n \cdot d)$ & $O(n \cdot d)$ & $T$ = iterations, $n$ = samples, $d$ = features \\\midrule
RUSBoost \cite{seiffert2009rusboost,al2024detecting} & $O(T \cdot n_{min} \cdot d)$ & $O(n \cdot d)$ & $n_{min}$ = minority samples; derived in \cite{al2024detecting} \\
\bottomrule
\end{tabular}
\end{table}

More fundamentally, as highlighted by \citet{Sakho_2024}, the field urgently needs rigorous theoretical frameworks to understand when and why rebalancing works. Their analysis shows that, under certain conditions, SMOTE's asymptotic behavior converges toward sample copying rather than genuine novel generation and that sample density can vanish near class boundaries in high dimensions. Furthermore, their empirical finding that no rebalancing is often competitive with SMOTE when modern classifiers are employed suggests that the advantages of oversampling are task-dependent and may be diminishing as base classifiers continue to improve. This observation is consistent with \citet{menardi2014training}, who demonstrated that a smoothed bootstrap resampling approach can effectively address class imbalance without explicit oversampling, and with the broader finding that sophisticated classifiers (e.g., SVM with appropriate class weights, tree-based ensembles) can partially mitigate imbalance effects \citep{Azhar_2023}.

While \citet{Sakho_2024} focus their theoretical analysis on SMOTE-based oversampling, their call for rigorous frameworks applies equally to undersampling and combination methods, where the theoretical understanding of when and why removal (rather than generation) of instances improves generalization remains similarly underdeveloped. Nevertheless, these isolated efforts do not enable a fair or rigorous complexity comparison across the diverse methods surveyed.

Future research must therefore pursue four interconnected priorities: (i) establishing the theoretical conditions under which rebalancing (both oversampling and undersampling) provably improves generalization; (ii) extending density bound analyses to other SMOTE variants and generative oversampling models; (iii) developing analogous theoretical frameworks for undersampling techniques to understand when instance removal is beneficial versus detrimental; and (iv) creating theoretically grounded adaptive rebalancing strategies that move beyond heuristic design. Addressing these foundational issues is essential for transforming class-imbalance learning from an empirically driven practice into a principled scientific discipline.

\subsection{The Classifier-Resampling Interaction: A Foundational Framework}
\label{sec:classifierInteraction}

A persistent, yet often implicit, assumption in the literature is that resampling and classifier selection are independent choices: one first balances the data, then trains a model. This assumption is flawed. The efficacy of a resampling method is not an intrinsic property of the method alone but an emergent property of the joint system comprising three components: (i) the resampling strategy, (ii) the base classifier's inductive bias, and (iii) the dataset's intrinsic structure \citep{Azhar_2023,glazkova2020comparison}.

This section synthesizes known theoretical and empirical results to provide a foundational framework for understanding this three-way interaction. The framework serves two purposes: first, to explain contradictory findings in the literature where the same resampling method succeeds in one study but fails in another; second, to guide practitioners in selecting resampling strategies that are compatible with their chosen classifier. The practical implications of this framework are then operationalized in the next subsection.

\subsubsection{Theoretical Foundations of the Interaction}

The interaction can be understood through three non-exclusive lenses:

\begin{enumerate}
    \item \textbf{Variance vs. Bias Trade-off: 
} Undersampling reduces the training set size, which increases the variance of the resulting classifier \citep{breiman1996bagging}. High-variance classifiers (e.g., unpruned Decision Trees, k-NN with small $k$) are particularly susceptible to performance degradation after undersampling. Conversely, bagging ensembles (e.g., balanced Random Forest) are designed to counteract this increased variance. Oversampling, by replicating or interpolating instances, introduces a bias toward the minority class distribution. Classifiers with strong built-in regularization (e.g., SVM, penalized Logistic Regression, XGBoost) can often tolerate or even benefit from this bias, whereas unregularized models (e.g., standard Decision Trees) are prone to severe overfitting, especially with aggressive SMOTE variants \citep{Sakho_2024}.

    \item \textbf{Noise Amplification vs. Robustness:} Methods like Borderline SMOTE and ADASYN deliberately generate synthetic samples near decision boundaries. For a classifier with a flexible decision boundary (e.g., a deep neural network or an unpruned tree), this provides crucial information. For a classifier with a rigid boundary (e.g., linear SVM), these same samples act as label noise, forcing the classifier to make unsupported local adjustments and degrading generalization \citep{batista2004study}. Consequently, combination methods that include a cleaning step (e.g., SMOTE-ENN) are systematically more beneficial for noise-sensitive classifiers.

    \item \textbf{The Curse of Dimensionality for Distance-Based Methods:} SMOTE and its variants rely on Euclidean distance in the original feature space. In high-dimensional datasets ($d > 100$), distances become indiscriminate, and the concept of ``nearest neighbor'' loses meaning \citep{Azhar_2023}. When paired with a classifier that is also distance-based (e.g., k-NN, kernel SVM), the performance of SMOTE variants collapses. However, if paired with a classifier that performs implicit dimensionality reduction (e.g., a regularized logistic regression or a deep neural network with an autoencoder-like first layer), the same synthetic samples can be beneficial. This explains why DeepSMOTE \citep{dablain2022deepsmote}, which operates in a learned latent space, was specifically designed to address this interaction.

\end{enumerate}
\subsubsection{A Prescriptive Framework for Practitioners}

Building on the theoretical analysis of classifier-resampling interactions presented in Section~\ref{sec:classifierInteraction}, this section provides a complementary decision matrix driven primarily by dataset characteristics. For guidance on selecting resampling methods based on the \textit{classifier's} inductive bias, refer to Table~\ref{tab:classifier_framework}.

\begin{table}[H]
\caption{Framework for selecting resampling methods based on classifier inductive bias.}
\label{tab:classifier_framework}
\footnotesize
\begin{tabular}{p{3.5cm}<{\raggedright} p{4.2cm}<{\raggedright} p{8cm}<{\raggedright}}
\toprule
\textbf{Classifier Type} & \textbf{Inductive Bias/Vulnerability} & \textbf{Recommended Resampling Strategy} \\
\midrule
Tree-based (DT, RF, XGBoost, LightGBM) & High variance; prone to overfitting; naturally handles feature interactions. & Prefer undersampling or simple oversampling (Random OverSampler). Avoid aggressive SMOTE (e.g., Borderline, ADASYN) without subsequent cleaning (e.g., SMOTE-ENN). \\
\midrule
Linear (LR, Linear SVM) & Rigid decision boundary; sensitive to noise and outliers; benefits from feature scaling. & Prefer cleaning-based methods (SMOTE-ENN, SMOTE-TL). Generative models (VAEs, CWGAN) can be highly effective if sufficient minority data exist. \\
\midrule
Distance-based (k-NN, Kernel SVM, RBF Networks) & Suffers from the curse of dimensionality; sensitive to local data structure. & Prefer cluster-aware methods (K-Means SMOTE, SMOTE-COF, SAO) or latent-space methods (DeepSMOTE, FW-SMOTE). Avoid standard SMOTE in high dimensions. \\
\midrule
Deep Neural Networks (MLP, CNN, LSTM) & Data-hungry; can model complex boundaries; vulnerable to mode collapse in generative pre-training. & Prefer generative models (VAEs, cGANs, Diffusion) when sufficient data exist. For smaller datasets, combine SMOTE with ensemble methods (SMOTEBoost) or use simple oversampling with strong regularization (dropout, weight decay). \\
\midrule
Ensemble (Bagging/Boosting) & Designed to reduce variance; can be robust to some imbalance. & Prefer ensemble-native methods (RUSBoost, SMOTEBoost, Balanced Bagging) over preprocessing + standard ensemble. These methods integrate resampling into the iterative process. \\
\bottomrule
\end{tabular}
\end{table}

\subsubsection{Implications for Research and Practice}

This framework has several direct implications:

\begin{enumerate}
    \item \textbf{No more isolated benchmarks:} Future empirical studies must report results across multiple classifier families. A claim that ``Method X outperforms Method Y'' is incomplete without specifying the classifier(s) used.

    \item \textbf{Resampling is not a panacea:} For some classifier-data combinations, particularly tree-based ensembles on modestly imbalanced data, no resampling may be necessary. As noted by \citet{Sakho_2024}, the default practice should not be ``always resample'', but rather ``diagnose, then decide.''

    \item \textbf{A guide for interpreting prior work:} The framework explains apparent contradictions in the literature. For example, the strong performance of SMOTE-ENN in some studies and its mediocrity in others can often be traced to the choice of base classifier (e.g., a noise-sensitive LR vs. a robust RF).

    \item \textbf{Opportunity for adaptive methods:} The ultimate goal is an adaptive resampling method that selects or tunes its strategy based on both the dataset and a specified target classifier. This remains an open and high-impact research direction.
\end{enumerate}

In summary, treating resampling as an independent preprocessing step is a simplification that often leads to suboptimal or non-reproducible results. By explicitly modeling the interaction with the classifier's inductive bias, practitioners can make more informed choices, and researchers can design more generalizable and impactful methods.

\subsection{Practical Decision Framework for Practitioners} \label{sec9-3}  
Building on the critical analyses above, we provide a practical decision framework to guide method selection based on dataset characteristics. Table~\ref{tab:decision_matrix} maps common data challenges to recommended resampling categories, specific methods, and associated caveats.

\begin{table}[t]
\caption{{Decision} matrix: mapping dataset characteristics to recommended resampling methods.}
\label{tab:decision_matrix}
\footnotesize
\begin{tabular}{p{3cm}<{\raggedright} p{3.8cm}<{\raggedright} p{3.6cm}<{\raggedright} p{5.2cm}<{\raggedright}}
\toprule
\textbf{Dataset Characteristic} & \textbf{Recommended Category} & \textbf{Specific Methods} & \textbf{Caveats} \\
\midrule
Small dataset ($n < 1000$) & Safe oversampling (avoid aggressive undersampling) & Safe-Level SMOTE \citep{bunkhumpornpat2009safe}, SMOTE-IPF \citep{SAEZ2015184} & Avoid GANs/VAEs (insufficient data to train); avoid undersampling (information loss critical) \\
\midrule
High dimensionality ($d > 1000$) & Generative models or feature-weighted oversampling & DeepSMOTE \citep{dablain2022deepsmote}, FW-SMOTE \citep{MALDONADO2022108511}, VAE \citep{kingma2022autoencodingvariationalbayes} & GANs risk mode collapse; VAEs need sufficient minority samples (not recommended for extremely small minority classes) \\
\midrule
Severe imbalance (IR $>$ 100:1) & Hybrid combination or ensemble & SMOTE-ENN \citep{batista2004study}, RUSBoost \citep{seiffert2009rusboost}, EasyEnsemble \citep{liu2008exploratory} & Undersampling alone discards too much information; ensembles increase training time \\
\midrule
Class overlap & Borderline methods + cleaning & Borderline SMOTE \citep{Han2005BorderlineSMOTEAN}, SMOTE-TL \citep{batista2004study}, SMOTE-ENN & Cleaning may remove informative borderline minority instances; tune $k$ carefully \\
\midrule
Mixed data types (numeric + categorical) & Modified distance metrics & SMOTE-NC \citep{chawla2002smote}, AMDO \citep{amdo} & Computational cost increases with categorical feature encoding; HVDM parameter sensitive \\
\midrule
Multi-label imbalance & LP-based or ML-based & LP-ROS \citep{charte2013first,charte2015addressing}, ML-ROS \citep{charte2013first,charte2015addressing}, REMEDIAL \citep{charte2015remedial}, MLSMOTE \citep{charte2015mlsmote} & LP methods suffer combinatorial explosion; ML methods ignore label correlations \\
\midrule
High noise level & Cleaning-based undersampling or combination & ENN \citep{wilson1972asymptotic}, Tomek Links \citep{tomek1976two}, SMOTE-ENN, OSS \citep{kubat1997addressing} & Risk of discarding informative borderline points; validate with domain knowledge \\
\midrule
Computational budget limited & Simple undersampling & Random Undersampling \citep{he2009learning} & Baseline only; rarely optimal for performance; use as benchmark \\
\midrule
Interpretability required & Simple interpolation or cleaning & SMOTE (original) \citep{chawla2002smote}, NearMiss-1 \citep{zhang2003nearmiss} & Avoid complex generative (GAN, VAE) or ensemble (boosting) methods \\
\midrule
Within-class imbalance (sparse subclusters) & Cluster-aware oversampling & K-Means SMOTE \citep{Douzas_2018}, SMOTE-COF \citep{meng2022imbalanced}, SAO \citep{tao2023self} & Requires determining number of clusters; silhouette method recommended \\
\midrule
Ensemble deployment (robustness prioritized over interpretability) & Ensemble strategies & Balanced Random Forest \citep{chen2004using}, SMOTEBoost \citep{chawla2003smoteboost}, BalanceCascade \citep{liu2008exploratory} & Higher computational cost; less interpretable; risk of overfitting with small datasets \\
\midrule
Complex data distribution (non-Gaussian, multi-modal, manifold structure) & Distribution-aware generative methods & GMMSampling \citep{naglik2023gmmsampling}, LAMO \citep{wang2021local}, Diffusion models \citep{ren2025diffusion} & Requires sufficient minority samples; hyperparameter sensitive; higher implementation complexity \\
\bottomrule
\end{tabular}
\end{table}
To use the matrix, practitioners should first diagnose their dataset's primary impediments: dimensionality, imbalance severity, class overlap, noise level, data type composition, and computational budget. For example, a high-dimensional credit card fraud dataset (\mbox{$d > 1000$}) with extreme imbalance (IR $> 100:1$) would consult the first two rows, suggesting either generative models (DeepSMOTE, VAE) or hybrid ensembles (SMOTE-ENN, RUSBoost), with the caveat that GANs risk mode collapse and ensembles increase \mbox{training time.}

The matrix is designed as a starting point, not a prescription. Multiple characteristics often co-occur (e.g., high dimensionality and class overlap), requiring iterative testing of candidates from different rows. When in doubt, we recommend beginning with a simple baseline (SMOTE or random undersampling), then progressively adding complexity: first cleaning (SMOTE-ENN), then cluster awareness (K-Means SMOTE), then generative methods if data permits. Practitioners should always validate selections using nested cross-validation with imbalance-appropriate metrics (PR-AUC, G-mean, or MCC) rather than accuracy. The caveats column highlights known failure modes for each recommendation, helping users anticipate when a seemingly appropriate method may underperform due to hidden dataset nuances.
\subsection{Limitations}

Several limitations warrant acknowledgment. 

First, our evaluation is restricted to two fraud detection datasets and two classifiers. Generalization to other domains (medical imaging, text classification, sensor networks) requires additional validation. 

Second, hyperparameters were set to library defaults; method-specific tuning might improve performance for some techniques. 

Third, computational metrics reflect our specific hardware; absolute values may vary, but relative rankings are likely stable. Complexity analysis beyond empirical runtime (e.g., asymptotic bounds) remains absent for most methods surveyed.

Fourth, generative models (GANs, VAEs, diffusion models) were excluded due to their substantially different architectural requirements; their evaluation remains an important direction for future work.

Fifth, the Friedman tests reported in the case study are used descriptively to confirm that observed differences exceed cross-validation noise. They do not imply that the results generalize beyond the two fraud detection datasets used.

\subsection{Future Research Directions}\label{sec:FutureResaerchDirection}

This review has examined resampling techniques for class imbalance, spanning foundational methods such as SMOTE to advanced deep generative models, including VAEs, GANs, and diffusion models. We categorized techniques by their underlying strategies---oversampling, generative models, undersampling, hybrid/ensemble approaches, and multi-label models---contextualized each category in terms of strengths, limitations, and suitability for high-dimensional spaces, noisy features, and multi-label tasks.

Our analysis reveals that no single method performs consistently across all scenarios. Effective imbalance handling requires matching the selected technique to dataset characteristics and task requirements, considering trade-offs among efficiency, adaptability, and computational cost. As data complexity grows, conventional resampling methods face fundamental limitations that motivate several emerging research directions. Figure~\ref{fig:future_directions} provides a structured overview of these future research directions in resampling and augmentation for imbalanced learning.

\begin{figure}[H]
\resizebox{0.99\textwidth}{!}{%
\begin{tikzpicture}[
    node distance=0.8cm,
    titlebox/.style={rectangle, fill=teal!70!black, text=white, draw=none, text width=14cm, minimum height=0.7cm, align=center, rounded corners=2pt, font=\bfseries\small},
    category/.style={rectangle, fill=teal!20, draw=teal!70!black, thick, text width=6.2cm, minimum height=0.6cm, align=center, rounded corners=2pt, font=\small\bfseries},
    method/.style={rectangle, draw=gray, thick, text width=6.0cm, minimum height=0.9cm, align=center, rounded corners=2pt, font=\footnotesize},
    question/.style={rectangle, fill=orange!20, draw=orange!70!black, thick, text width=6.0cm, minimum height=0.9cm, align=center, rounded corners=2pt, font=\footnotesize\itshape},
    arrow/.style={thick, ->, >=stealth, teal!70!black},
    qarrow/.style={thick, ->, >=stealth, orange!70!black, dashed}
]

\node[titlebox] (title) at (0,0) {Future Research Directions in Resampling and Augmentation for Imbalanced Learning};

\node[category] (cat1) at (-4.0,-1.2) {End-to-End Deep Learning};
\node[category] (cat2) at (4.0,-1.2) {Cost-Sensitive Learning};

\node[method] (dir1a) at (-4.0,-2.1) {IRDA: Implicit augmentation for\\imbalanced regression with surrogate loss};
\node[method] (dir1b) at (-4.0,-3.0) {GAN + RL joint optimization for fraud\\detection with closed-loop feedback};

\node[method] (dir2a) at (4.0,-2.1) {Adjusts misclassification penalties\\instead of resampling};
\node[method] (dir2b) at (4.0,-3.0) {Multiform framework combining\\feature selection + oversampling + cost};

\node[question] (q1) at (-4.0,-4.1) {Open: Provable benefits for specific\\imbalance types; scaling to large systems};
\node[question] (q2) at (4.0,-4.1) {Open: Integration with deep learning;\\automatic cost tuning};

\node[category] (cat3) at (-4.0,-5.4) {Knowledge Distillation};
\node[category] (cat4) at (4.0,-5.4) {Self-Supervised Learning};

\node[method] (dir3a) at (-4.0,-6.3) {Multi-teacher distillation with\\diverse resampling strategies};
\node[method] (dir3b) at (-4.0,-7.2) {Multi-Stage Self-Distillation (MSSD):\\staged augmentation + sequential distillation};

\node[method] (dir4a) at (4.0,-6.3) {SSL more robust to imbalance but\\depends critically on dataset size};
\node[method] (dir4b) at (4.0,-7.2) {Token-class imbalance in vision\\transformers (Han et al.)};

\node[question] (q3) at (-4.0,-8.4) {Open: Teacher selection/weighting;\\scaling to high-dimensional data};
\node[question] (q4) at (4.0,-8.4) {Open: When does SSL robustness hold?\\Token imbalance mitigation strategies};

\node[category] (cat5) at (-4.0,-9.7) {Diffusion Models};
\node[category] (cat6) at (4.0,-9.7) {Advanced Tabular Architectures};

\node[method] (dir5a) at (-4.0,-10.6) {Diffusion-GAN hybrid (Ren et al.):\\faster sampling + better conditioning};
\node[method] (dir5b) at (-4.0,-11.5) {FRACTAL: transformer-guided diffusion\\+ confidence-calibrated TabNet};
\node[method] (dir5c) at (-4.0,-12.4) {Mamba-DDPM-BSA: boundary sampling\\+ interactive synthesis};

\node[method] (dir6a) at (4.0,-10.6) {TabTransformer: Transformer + MLP\\robust to missing/noisy features};
\node[method] (dir6b) at (4.0,-11.5) {FT-Transformer: pure Transformer\\competitive with Gradient Boosted DT};
\node[method] (dir6c) at (4.0,-12.4) {GrowNet: neural boosting with\\fully corrective step};

\node[question] (q5) at (-4.0,-13.6) {Open: Reduce sampling time; adapt to\\extremely small minority classes};
\node[question] (q6) at (4.0,-13.6) {Open: Do these reduce need for\\resampling? Architecture-specific strategies};

\node[category] (cat7) at (0,-15.0) {Foundation Models};

\node[method] (dir7a) at (0,-15.9) {CLIP: robust to class distribution shifts via zero-shot transfer};
\node[method] (dir7b) at (0,-16.8) {Underexplored: adapting foundation models to highly skewed tasks};

\node[question] (q7) at (0,-18.2) {Open: Fine-tuning for extreme imbalance without destroying zero-shot;\\prompting strategies for long-tail distributions; bias mitigation};

\draw[arrow] (title.south)--++(0,-0.3) -| (cat1.north);
\draw[arrow] (title.south)--++(0,-0.3) -| (cat2.north);

\draw[arrow] (cat1.south)--++(0,-0.1) -| (dir1a.north);
\draw[arrow] (cat2.south)--++(0,-0.1) -| (dir2a.north);

\draw[arrow] (title.south) --++(0,-4.5) -| (cat3.north);
\draw[arrow] (title.south)--++(0,-4.5) -| (cat4.north);

\draw[arrow] (cat3.south)--++(0,-0.1) -| (dir3a.north);
\draw[arrow] (cat4.south)--++(0,-0.1) -| (dir4a.north);

\draw[arrow] (title.south) --++(0,-8.8) -| (cat5.north);
\draw[arrow] (title.south)--++(0,-8.8) -| (cat6.north);

\draw[arrow] (cat5.south)--++(0,-0.1) -| (dir5a.north);
\draw[arrow] (cat6.south)--++(0,-0.1) -| (dir6a.north);

\draw[arrow] (title.south) --++(0,0) -| (cat7.north);

\draw[arrow] (cat7.south)--++(0,-0.1) -| (dir7a.north);

\draw[qarrow] (dir1b.south)--(q1.north);
\draw[qarrow] (dir2b.south)--(q2.north);
\draw[qarrow] (dir3b.south)--(q3.north);
\draw[qarrow] (dir4b.south)--(q4.north);
\draw[qarrow] (dir5c.south)--(q5.north);
\draw[qarrow] (dir6c.south)--(q6.north);
\draw[qarrow] (dir7b.south)--(q7.north);




\end{tikzpicture}
}
\caption{Overview of future research directions in resampling and augmentation for imbalanced learning.}
\label{fig:future_directions}
\end{figure}
\subsubsection{Integration with End-to-End Deep Learning}

Rather than treating data balancing as an isolated preprocessing step, recent work explores implicit augmentation mechanisms and joint optimization frameworks that integrate data generation directly with downstream model training.

\citet{zhu2024irda} introduced IRDA for imbalanced regression, developing an augmentation strategy specifically tailored for regression tasks with a computable surrogate loss function. Their approach includes two variants, one with meta-learning integration and one without, and provides a regularization perspective for understanding \mbox{implicit augmentation.}

\citet{chen2026adversarial} proposed a joint optimization framework for financial statement fraud detection that constructs a GAN generator as a reinforcement learning-based ``Counterfeiter'' to transform normal samples into fraudulent ones. A multi-objective reward function simultaneously enhances sample fidelity and domain consistency, while a closed-loop feedback mechanism dynamically evaluates synthetic samples and integrates them into classifier learning, achieving end-to-end training.

Key open questions include: (i) how to design implicit augmentation methods that are provably beneficial for specific imbalance types (e.g., severe imbalance, within-class imbalance); (ii) whether meta-learning frameworks can automatically discover optimal augmentation strategies without task-specific tuning; and (iii) how to scale joint optimization approaches to large-scale deep learning systems without prohibitive \mbox{computational overhead.}

\subsubsection{Cost-Sensitive Learning as a Complementary Paradigm}
Resampling modifies data distribution; cost-sensitive learning adjusts misclassification penalties instead. \citet{araf2024cost} provided a comprehensive review of cost-sensitive learning for imbalanced medical data, noting a growing preference for direct cost-sensitive approaches since 2020. \citet{liang2025multiform} proposed a multiform framework combining multiobjective feature selection with both oversampling and cost-sensitive learning.

\subsubsection{Resampling with Knowledge Distillation}

Knowledge distillation offers a promising paradigm for addressing class imbalance while simultaneously achieving model compression. The core idea is to train multiple teacher models on differently resampled datasets and distill their collective knowledge into a compact student model.

\citet{yousefimehr2026multi} introduced a multi-teacher knowledge distillation framework wherein multiple teachers are trained on datasets resampled using diverse oversampling and undersampling strategies. The student model learns a balanced representation while maintaining a compact structure, offering advantages in accuracy, efficiency, and inference speed for lightweight anomaly detection. 

Extending this paradigm, \citet{yousefimehr2026IEEE} proposed Multi-Stage Self-Distillation (MSSD), an educationally inspired framework that integrates staged augmentation and resampling with sequential self-distillation. In this approach, a student model refines its understanding by learning from its predecessor (the teacher model) at each stage, mimicking a process where advanced students guide their peers. This systematic combination of multiple resampling and augmentation techniques leverages their complementary strengths while mitigating individual limitations, achieving superior performance in identifying rare anomalies.

Key open questions include: (i) how to optimally select and weight the ensemble of teachers in multi-teacher distillation for imbalanced learning; (ii) whether the sequential self-distillation paradigm can be extended to more stages without diminishing returns; (iii) how to adapt these frameworks to extremely high-dimensional data (e.g., images, text) while maintaining computational efficiency; and (iv) whether knowledge distillation can be combined with other paradigms such as self-supervised learning or diffusion-based augmentation for further gains.

\subsubsection{Self-Supervised Learning for Imbalance}

Self-supervised learning (SSL) has emerged as a promising paradigm for learning representations without labels, with several recent works investigating its interaction with class imbalance.

\citet{liu2021self} systematically investigated SSL under dataset imbalance, finding that off-the-shelf self-supervised representations (e.g., SimCLR, MoCo) are significantly more robust to class imbalance than supervised representations. The performance gap between balanced and imbalanced pre-training is substantially smaller for SSL than for supervised learning, particularly for out-of-domain evaluation. The authors hypothesized that SSL learns richer features from frequent data, including label-irrelevant-but-transferable features that benefit rare classes. They further proposed a re-weighted regularization technique that consistently improves SSL representation quality on imbalanced datasets.

\citet{espis2025comparative} conducted a comparative analysis of SSL versus supervised learning on small, imbalanced medical imaging datasets across four binary classification tasks. In contrast to the general robustness claims of SSL, they found that supervised learning often outperformed SSL when training sets were small, highlighting that the benefits of SSL are not universal and depend critically on dataset size, label availability, and class frequency distribution.

\citet{hamad2025self} proposed a novel SSL network for human activity recognition that addresses class imbalance through iSMOTE, an enhanced version of SMOTE that accurately labels synthetic samples. Additionally, they introduced a random masking technique that removes identity mapping from input data, enabling the construction of generic semantic representations.

\citet{han2026revisiting} identified a previously overlooked challenge in token-based SSL: token-class imbalance, where a small subset of frequent tokens (often representing uninformative backgrounds) dominates training while semantically rich but rare tokens are underrepresented. They proposed two solutions: a class-balanced cross-entropy loss that re-weights training signals based on token rarity and semantic-aware label smoothing (SLS), a regularization technique leveraging token embedding similarity to create more meaningful soft targets. Their methods significantly improved both discriminative and generative performance in models such as MAGE and MaskGIT.

\citet{cong2024adaptive} proposed an end-to-end SSL framework tailored for imbalanced medical image classification. Their framework introduces an adaptive contrastive loss that dynamically adjusts the model's learning focus between feature learning and classifier learning and a feature aggregation mechanism based on Graph Neural Networks to enhance feature discriminability. This addresses limitations of existing SSL methods that either lack end-to-end capabilities or require substantial memory resources.

Key open questions include: (i) under what conditions does SSL's robustness to imbalance generalize to small datasets, given the contradictory findings of \citet{liu2021self} and \citet{espis2025comparative}; (ii) how can token-class imbalance be systematically mitigated in vision transformers and other token-based architectures; (iii) whether SSL can be combined with resampling or augmentation methods for synergistic effects; and (iv) how to design SSL methods that are explicitly aware of and robust to class imbalance during pretraining.

\subsubsection{Diffusion Models for Data Oversampling and Augmentation}

Diffusion models have emerged as a powerful generative framework for minority-class augmentation, offering advantages in sample quality and training stability over GANs~\citep{ho2020denoising}. Recent work has explored diffusion-based methods for both tabular and image data, with diverse architectural innovations.

For tabular data, \citet{ren2025diffusion} proposed a diffusion-GAN hybrid that leverages the synergy between a diffusion-based generator and a discriminator with a Noise-Sensitive Auxiliary Classifier (NSAC), achieving faster sampling speed than pure diffusion models while maintaining better conditional generation ability. \citet{mishra2026diffusion} introduced FRACTAL, a two-stage framework comprising FAST-DN (a transformer-guided diffusion model with fidelity gating and residual correction) for high-fidelity minority augmentation, paired with RIDE-TabNet (a confidence-calibrated ensemble TabNet) for robust classification.

For image data, \citet{koh2025synthetic} developed a two-stage synthetic augmentation framework using pre-trained diffusion models with positive and negative prompt conditioning, promoting intra-class diversity and inter-class separation for long-tailed food classification. \citet{zhang2025mamba} proposed Mamba-DDPM-BSA, which integrates a Mamba-based diffusion model (leveraging Mamba's global modeling capability and linear computational efficiency) with a boundary sampling algorithm that focuses generation on decision boundary regions. Notably, their method employs an interactive synthesis strategy where the generator targets the defects of the classification model, pushing the decision boundary toward the true minority distribution.

Hybrid approaches combining diffusion with classical methods have also been investigated. \citet{kumar2025hybrid} proposed DiSMHA, a framework that sequentially integrates Stable Diffusion for pixel-level image synthesis with SMOTE for feature-level oversampling, demonstrating substantial improvements for traditionally imbalance-sensitive classifiers such as SVM and MLP.

Despite their advantages, diffusion models incur substantially higher computational costs than traditional oversampling methods due to iterative denoising steps. Thus, diffusion models are currently best suited for applications where sample quality and stability are prioritized over computational efficiency. Key open questions include how to reduce sampling time, how to adapt diffusion models for extremely small minority classes, and how to best combine diffusion-based generation with classifier-specific feedback loops.


\subsubsection{Advanced Tabular Architectures and the Rethinking of Resampling}

The emergence of sophisticated deep learning architectures for tabular data presents new opportunities and open questions for imbalanced learning. Three representative lines of work illustrate this trend. First, TabTransformer \citep{huang2020tabtransformer} integrates contextual embeddings learned via a Transformer into a multilayer perceptron (MLP), demonstrating that self-attention can capture inter-feature interactions and that the resulting contextual embeddings are robust against missing and noisy features while matching the performance of tree-based ensemble models. Second, the FT-Transformer \citep{gorishniy2021revisiting} replaces the MLP component entirely with a Transformer, achieving competitive or superior performance on most tabular benchmarks, though the authors note that no universally superior solution exists compared to Gradient Boosted Decision Trees. Third, GrowNet \citep{badirli2020gradient} adopts a different paradigm, using shallow neural networks as weak learners within a gradient boosting framework with a fully corrective step that addresses the greedy approximation of classic boosting, thereby combining the expressivity of neural networks with sequential \mbox{error correction.}

A critical open question is whether these powerful architectures reduce or even eliminate the need for explicit resampling. The attention mechanism may inherently downweight the influence of the majority class, while the fully corrective nature of GrowNet could potentially adjust for class imbalance across successive rounds. However, these hypotheses require systematic empirical validation. Conversely, these models may introduce novel forms of imbalance sensitivity that are not yet well understood.

Future work should systematically investigate three interconnected questions: (i) whether standard resampling (e.g., SMOTE and RUS) provides diminishing returns when paired with advanced tabular architectures; (ii) whether architecture-specific balancing strategies, such as attention-based reweighting for Transformers or class-weighted boosting rounds for GrowNet, outperform classical preprocessing; and (iii) how to design hybrid frameworks that co-optimize architecture choice and balancing mechanism rather than treating them as independent decisions.

\subsubsection{Foundation Models and Imbalance}

Large-scale pretrained models trained on massive, naturally imbalanced web-scale data via contrastive language-image pretraining \citep{radford2021learning} exhibit inherent robustness to class distribution shifts, enabling zero-shot transfer to downstream tasks without task-specific training. However, as noted by \citet{bommasani2021opportunities}, foundation models remain poorly understood: their emergent capabilities and failure modes are not yet characterized, and defects inherited from imbalanced pretraining data may propagate to all adapted \mbox{models downstream.}

Systematic strategies for adapting foundation models to highly skewed downstream tasks remain underexplored. Key open questions include: (i) how to effectively fine-tune foundation models for tasks with extreme class imbalance without destroying their zero-shot capabilities; (ii) whether prompting strategies can be designed to handle long-tail distributions without additional training data; (iii) how to diagnose and mitigate bias inherited from imbalanced pretraining data; and (iv) whether foundation models can be explicitly pretrained with imbalance-aware objectives to improve their \mbox{downstream adaptability.}

\subsubsection{Synthesis}
These directions underscore a shift from static, data-centric balancing toward dynamic, model-aware strategies. Future research should prioritize unified frameworks that seamlessly integrate resampling, cost sensitivity, knowledge distillation, and representation learning within end-to-end pipelines. As data collection remains expensive or infeasible in critical domains such as healthcare and scientific discovery, hybrid generative approaches combining domain priors with learned distributions \citep{cranmer2020discovering} will likely play an increasing role. Continued exploration of adaptive, data-aware methodologies is needed to scale with model complexity while maintaining interpretability, reliability, and deployment efficiency.

\appendix

\section{Complete Search Strings and Selection Statistics}
\label{app:searchstrings}

This appendix reports the complete search queries and the resulting record counts, ensuring full reproducibility of the systematic literature review described in Section~\ref{sec:methodology}. All searches were executed on \textbf{31 March 2026}.

\subsection{Search protocols by database}

\textbf{Scopus.}

Two search modes were employed:
\begin{enumerate}
    \item Title-only searches: Keywords were searched exclusively in article titles.
    \item Title/Abstract/Keyword searches: Keywords were searched across title, abstract, and author keywords.
\end{enumerate}

Table~\ref{tab:ScopusStatistics} reports the number of records retrieved for each query configuration, disaggregated by document type.

\begin{table}[H]
\footnotesize
\caption{Scopus search results by query configuration (31 March 2026).}
\label{tab:ScopusStatistics}
\begin{tabular}{p{4.2cm}<{\raggedright} p{5.2cm}<{\raggedright} ccccc}
\toprule
\textbf{Search Field} & \textbf{Additional Constraints} & \textbf{Article} & \textbf{Conf. Paper} & \textbf{Review} & \textbf{Book Ch.} & \textbf{Total} \\
\midrule
\multicolumn{7}{c}{\textbf{Title-only searches}} \\
\midrule
Title: ``resampling'' & None & 7 & 9 & 0 & 0 & 16 \\
Title: ``oversampling'' & None & 28 & 18 & 0 & 0 & 46 \\
Title: ``undersampling'' & None & 1 & 2 & 0 & 0 & 3 \\
Title: ``imbalanced learning'' & None & 5 & 4 & 1 & 0 & 10 \\
Title: ``imbalanced data'' & None & 41 & 22 & 2 & 1 & 66 \\
Title: ``class imbalance'' & None & 38 & 47 & 0 & 0 & 85 \\
\midrule
\multicolumn{7}{c}{\textbf{Title/Abstract/Keyword searches}} \\
\midrule
``oversampling'' & ``imbalanced data'' OR ``class imbalance'' OR ``imbalanced learning'' & 18 & 6 & 0 & 0 & 24 \\
``SMOTE'' & ``imbalanced data'' OR ``class imbalance'' OR ``imbalanced learning'' & 16 & 10 & 1 & 0 & 27 \\
``synthetic oversampling'' & ``imbalanced data'' OR ``class imbalance'' OR ``imbalanced learning'' & 0 & 1 & 0 & 0 & 1 \\
``ADASYN'' & ``imbalanced data'' OR ``class imbalance'' OR ``imbalanced learning'' & 2 & 1 & 0 & 0 & 3 \\
``borderline SMOTE'' & ``imbalanced data'' OR ``class imbalance'' OR ``imbalanced learning'' & 1 & 0 & 0 & 0 & 1 \\
``Tomek link'' & ``imbalanced data'' OR ``class imbalance'' OR ``imbalanced learning'' & 1 & 0 & 0 & 0 & 1 \\
``adaptive SMOTE'' & ``imbalanced data'' OR ``class imbalance'' OR ``imbalanced learning'' & 1 & 1 & 0 & 0 & 2 \\
``undersampling'' & ``imbalanced data'' OR ``class imbalance'' OR ``imbalanced learning'' & 1 & 1 & 0 & 0 & 2 \\
``Ensemble'' & ``imbalanced data'' OR ``class imbalance'' OR ``imbalanced learning'' & 20 & 16 & 1 & 0 & 37 \\
``Hybrid'' & ``imbalanced data'' OR ``class imbalance'' OR ``imbalanced learning'' & 29 & 14 & 0 & 3 & 46 \\
``Hybrid'' & ``oversampling'' & 9 & 4 & 1 & 0 & 14 \\
``Hybrid'' & ``undersampling'' & 2 & 1 & 0 & 0 & 3 \\
``data augmentation'' & ``generative model'' & 153 & 207 & 10 & 3 & 373 \\
``data augmentation'' & ``imbalanced data'' OR ``class imbalance'' OR ``imbalanced learning'' & 250 & 247 & 7 & 1 & 505 \\
``deep augmentation'' & ``imbalanced data'' OR ``class imbalance'' OR ``imbalanced learning'' & 14 & 23 & 0 & 0 & 37 \\
``variational autoencoder'' & ``imbalanced data'' OR ``class imbalance'' OR ``imbalanced learning'' & 9 & 2 & 0 & 0 & 11 \\
``GAN'' & ``imbalanced data'' OR ``class imbalance'' OR ``imbalanced learning'' & 45 & 35 & 3 & 1 & 84 \\
``generative adversarial network'' & ``imbalanced data'' OR ``class imbalance'' OR ``imbalanced learning'' & 71 & 23 & 1 & 0 & 95 \\
``diffusion model'' & ``imbalanced data'' OR ``class imbalance'' OR ``imbalanced learning'' & 12 & 11 & 0 & 0 & 23 \\
\bottomrule
\end{tabular}
\end{table}

\textbf{Other Databases.}

IEEE Xplore, Springer Nature, and arXiv were searched using equivalent keyword combinations adapted to each platform's advanced search interface. Table~\ref{tab:OtherDatabaseStatistics} reports the results. ACM Digital Library, Google Scholar, ScienceDirect, and ResearchGate were also searched; however, these platforms do not provide downloadable result counts in a structured format. For these sources, manual inspection of the first 200 results per query was performed, and relevant papers were added to the candidate pool.

\begin{table}[H]
\caption{Search results from IEEE Xplore, Springer Nature, and arXiv (31 March  2026).}
\label{tab:OtherDatabaseStatistics}
\begin{tabular}{p{4cm}<{\raggedright} p{6.3cm}<{\raggedright}ccc}
\toprule
\textbf{Search Field (Title)} & \textbf{Additional Constraints (Abstract)} & \textbf{IEEE Xplore} & \textbf{Springer Nature} & \textbf{arXiv} \\
\midrule
``resampling'' & None & 0 & 611 & 382 \\
``oversampling'' & None & 92 & 369 & 194 \\
``undersampling'' & None & 0 & 121 & 166 \\
``imbalanced learning'' & None & 10 & 69 & 62 \\
``imbalanced data'' & None & 1 & 683 & 215 \\
``class imbalance'' & None & 0 & 382 & 208 \\
``SMOTE'' & None & 0 & 327 & 46 \\
``synthetic oversampling'' & None & 0 & 12 & 14 \\
``ADASYN'' & None & 0 & 5 & 2 \\
``Tomek link'' & None & 0 & 3 & 0 \\
``adaptive SMOTE'' & None & 0 & 10 & 1 \\
``Ensemble'' & ``imbalanced data'' OR ``class imbalance'' OR ``imbalanced learning'' & 2 & 43 & 107 \\
``Hybrid'' & ``imbalanced data'' OR ``class imbalance'' OR ``imbalanced learning'' & 7 & 23 & 57 \\
``data augmentation'' & ``generative model'' & 157 & 227 & 157 \\
``variational autoencoder'' & ``imbalanced data'' OR ``class imbalance'' OR ``imbalanced learning'' & 202 & 0 & 4 \\
``GAN'' & ``imbalanced data'' OR ``class imbalance'' OR ``imbalanced learning'' & 670 & 2 & 49 \\
``diffusion model'' & ``imbalanced data'' OR ``class imbalance'' OR ``imbalanced learning'' & 144 & 0 & 22 \\
\bottomrule
\end{tabular}
\end{table}


\subsection{Registration and protocol} 

The protocol for this systematic review was registered with the Open
Science Framework (OSF) Registries on [21 April 2026]. Registration DOI is: 

\centering{\url{https://doi.org/10.17605/OSF.IO/WH6SX}}




\end{document}